\documentclass[11pt]{book}

\usepackage[T1]{fontenc}
\usepackage{hyperref}
\usepackage{microtype}
\usepackage{lettrine}
\usepackage{xltabular}

\usepackage{standalone}
\usepackage{pgfplots}
\pgfplotsset{compat=1.17}
\usepgfplotslibrary{external}
\usepgfplotslibrary{groupplots}
\usepgfplotslibrary{fillbetween}
\usetikzlibrary{fadings}

\usepackage[commands,environments,enumerate,citations,notes,a4paper]{AVT}

\bibliography{Terenin-A-2022-PhD-Thesis}

\title{Gaussian Processes and Statistical Decision-making in Non-Euclidean Spaces}
\author{Alexander Terenin}
\date{February 2022}

\begin{document}

\begin{titlepage}
\maketitlehooka
\centering
\huge
\null
\vfill
\thetitle
\par
\vfill
\LARGE
\theauthor
\par
\large
Department of Mathematics
\par
Imperial College London
\par
\vfill
\null
\vfill
a dissertation submitted for the degree of
\par
Doctor of Philosophy
\par
\strut
\par
\thedate
\par
\vfill
\null
\maketitlehookd
\end{titlepage}

\chapter*{Declaration}

This dissertation and all research contained in it are a product of my original work, except where indicated otherwise by explicit statement or reference.
All ideas, quotations, and data originating from the works of others, published or otherwise, are fully acknowledged according to standard practices of the academic discipline.

\chapter*{Copyright}

The copyright of this thesis rests with the author. Unless otherwise indicated, its contents are licensed under a Creative Commons Attribution 4.0 International License (CC BY).

Under this license, you may copy and redistribute the material in any medium or format for both commercial and non-commercial purposes.
You may also create and distribute modified versions of the work. 
This is on the condition that you credit the author.

When reusing or sharing this work, ensure you make the license terms clear to others by naming the license and linking to the license text. 
Where a work has been adapted, you should indicate that the work has been changed and describe those changes.

Please seek permission from the copyright holder for uses of this work that are not included in this license or permitted under UK Copyright Law.

\chapter*{Acknowledgments}

This thesis is dedicated to the several hundred Twitter users who thought what I was working on was interesting enough to give it a read, or a comment.
Thanks to you, my experiment in writing a thesis in an open-source manner visible to the public was a resounding success.
Getting the opportunity to write this thesis has been a dream come true, and I am delighted to share the experience with you.

The original acknowledgments I had planned to write during most of my doctoral work would have started on a far less positive note.
The sheer volume of kind feedback I've received in the weeks during which I was writing this thesis, almost all of it from people I've never met, has convinced me to leave the past behind, and not write about certain people, actions, and events that do not deserve to be remembered.
Instead, I choose to thank those due to whom I had the chance.

Firstly, I am profoundly grateful to Marc Deisenroth for taking me as a student, and giving me a \emph{third} chance to complete a Ph.D., and to Seth Flaxman for taking me as a co-supervisee in order to make this possible.
Most people who do not succeed the first time, whatever the reason may be, don't even get a second chance, much less a third one.
You believed in me in the darkest moments of my academic career, and anything that I ever accomplish in my career will only have been possible because of that belief.

Secondly, I am also profoundly grateful to David Draper for believing in me, and taking me as a student when nobody was willing to talk to me or take my ideas seriously.
It is thanks to you that I have a master's degree, and though I was not able to finish a doctorate at the time under your guidance, I hope this thesis shows that you did the right thing in supporting me, even when it became clear I had no academic future at the place I was at, and later in helping me return to scientific work.

Thirdly, I am grateful to Eric Xing  for hiring me when my career was otherwise in the midst of full-blown chaos.
The time I spent at Petuum convinced me that the machine learning community was the right intellectual home for me, and prompted me to look beyond the narrow scope of ideas I had previously been around.
Though I was not truly ready to be an independent researcher at that time, and therefore accomplished little, I am grateful for the opportunity and all that I learned from it.

I thank my thesis committee members, Michael Bronstein and Stefano Ermon, for examining me and for their comments during the viva.
Your careful reading helped significantly improve the quality of this thesis.
I am particularly grateful for Michael's comment that "[t]he introductory chapters, with enough effort, could probably be transformed into a book." 
These are among the kindest words my work has ever received.

I also thank my large and ever-growing set of scientific collaborators.
You all give me confidence that, together, we can solve all of the hardest and most important problems!
I am constantly learning from each and every one of you, and none of the work in this thesis would have been possible without your support, ideas, and opportunity to discuss.

I am particularly grateful to Viacheslav Borovitskiy for both the opportunity to collaborate on all of the work presented in this thesis, and to reconnect with my heritage by constantly practicing my Russian.
Together, we were able to write substantially stronger work than any of us, or at least certainly I, would have alone---the awards we've won together speak for themselves.
A large part of why I have gotten better at practicing mathematics over the years---a skill of fundamental importance---is thanks to working with you.

I thank James T. Wilson, Steindór Sæmundsson, Samuel Cohen, and So Takao, for the opportunity to do research and write together. 
I am grateful to Yicheng Luo, Sanket Kamthe, Hugh Salimbeni, Yasemin Bekiroğlu, Sicelukwanda Zwane, K. S. Sesh Kumar, and others in the group who have spoken to me at length about ideas during my time.
I feel fortunate to to have been your labmate, and have learned an immense amount from you over the years.

I thank Andreas Hochlehnert, Kai Biegun, and Lucas Cosier for trusting in me as a researcher enough to allow me to supervise your work. 
Watching your ideas develop and grow has been a beautiful experience, and I look forward to seeing you shine in the years to come.

I am grateful to Michael Hutchinson, James-Michael Leahy, Peter Mostowsky, Iskander Azangulov, and others who I have collaborated with on papers, or spoken to at length about ideas.
I am particularly grateful to Brandon Amos both for the opportunity to collaborate, and for suggesting I speak to Marc Deisenroth in the weeks leading up to me becoming his student---without this, my path would have been very different.
I thank Nick Sharp for teaching me how to make the three-dimensional figures in this thesis.

I am grateful to Imperial College, and the Department of Mathematics in particular, for both the opportunity in funding my studies, and for support when things did not go as planned.
I am particularly thankful to Henrik Jensen: thank you for treating me with kindness in difficult times.
I am grateful to Pierre Degond, Darryl Holm, and Chris Hallsworth for your ideas in teaching me mathematics. 
I joined Imperial because I wanted to get better at mathematics, and it is thanks largely to you that I feel I succeeded at this.

I am thankful to Måns Magnusson, Leif Jonsson, and Shawfeng Dong for collaborating with me in my early days as a researcher, and helping me learn and get my bearings together.
In particular, I thank Peter Drake and Raya Feldman for first setting me on this path.
I am grateful to Matthew Johnson and Chris De Sa for the counterexamples sent to me in the early times, which convinced me that proper mathematics was of utmost importance and that I had to improve at it in order trust myself to say things that are true.

I am grateful to my friends at Carnegie Mellon University, including Dominic Chen, Aurick Qiao, Willie Neiswanger, Kumail Jaffer, Ziv Scully, Sarah Allen Scully, Sol Boucher, Guillaume Didier, Stefan Muller, Priya Donti, Gabriele Farina, Noam Brown, Ben Blum, Evan Cavallo, and others who convinced me to apply to a PhD program by showing me that many that of the people spending their life studying ideas were just like me.
I am grateful to Yuanran Zhu for showing me the same, but in a different place and time.

I am grateful to friends who have supported me over the years, including to Jon Frost for almost a decade of friendship, as well as Paula Siauchó Unriza, Victor Espinosa, Sam Aragon, and others who know me well and have been there for me over the years.
Even as time had gone on, whether we last spoke days or years ago, I have always found nothing about our friendship to have changed whenever we have had the chance to speak again.

Finally, I want to thank my family, including my mom, Irina Terenina. 
Without your sacrifices, I would not have spent my youth in the United States, or had almost any of the opportunities I have had. 
I am grateful to my grandmother, Olga Novikova, for your wisdom and for keeping our family together.
I wish you could have seen me graduate and write this thesis.

I am grateful to my dad, Vadim Terenin: your authoritative presentation style has at times blinded me to your ideas. 
I am grateful to my stepmom, Rheesa Eddings, for both your deeply insightful way of understanding the world, and for your humanity, diplomacy, kindness, helpful ideas, and support over the years.

My gratitude to family includes those who are family by virtue of their friendship and support for my entire adult life, including Jerry Hirsch, John Halper, Lois Morera, Christian Morera, and Peter Morera.
Thank you for continuing to stay in touch, for being the extended family I never otherwise had, and for being with me as my journey has unfolded.

Though this list of acknowledgments has ballooned beyond what might otherwise be expected in a dissertation, I nonetheless feel compelled to write it in full by virtue of my path being what it was.
It is also without a doubt incomplete, so I am grateful to anyone whose name should have been here but has been accidentally omitted, including whoever suggested I limit all paragraphs written to seven lines or less.
To conclude, I offer you, the reader, my exceeding gratitude for taking the time to read my work and ideas.

\chapter*{Abstract}

Bayesian learning using Gaussian processes provides a foundational framework for making decisions in a manner that balances what is known with what could be learned by gathering data. 
In this dissertation, we develop techniques for broadening the applicability of Gaussian processes.
This is done in two ways.

Firstly, we develop pathwise conditioning techniques for Gaussian processes, which allow one to express posterior random functions as prior random functions plus a dependent update term.
We introduce a wide class of efficient approximations built from this viewpoint, which can be randomly sampled once in advance, and evaluated at arbitrary locations without any subsequent stochasticity.
This key property improves efficiency and makes it simpler to deploy Gaussian process models in decision-making settings.

Secondly, we develop a collection of Gaussian process models over non-Euclidean spaces, including Riemannian manifolds and graphs.
We derive fully constructive expressions for the covariance kernels of scalar-valued Gaussian processes on Riemannian manifolds and graphs.
Building on these ideas, we describe a formalism for defining vector-valued Gaussian processes on Riemannian manifolds. 
The introduced techniques allow all of these models to be trained using standard computational methods.

In total, these contributions make Gaussian processes easier to work with and allow them to be used within a wider class of domains in an effective and principled manner.
This, in turn, makes it possible to potentially apply Gaussian processes to novel decision-making settings.

\tableofcontents

\listoffigures

\chapter*{Principal Notation}
\chaptermark{Principal Notation}

\section*{General}

\newif\ifnodotfill
\begin{xltabular}{\textwidth}{@{} p{1.625cm} X<{\ifnodotfill\else\unskip\kern-1pc\leaders\hbox{\makebox[1pc][r]{\makebox[0pc]{.}}}\hfill\kern0pc\fi} @{} >{\raggedleft\arraybackslash}p{0.625cm} @{}}
$(\Omega,\c{F},\P)$ & Probability space & \labelcpageref{ntn:prob} \\
$\E(\.)$ & Expectation of a real-valued random variable & \labelcpageref{ntn:expectation-cov} \\
$\Cov(\.,\.)$ & Covariance between two real-valued random variables & \labelcpageref{ntn:expectation-cov} \\
$y \~ \pi$ & Random variable $y$ with distribution $\pi$ & \labelcpageref{ntn:rv} \\
$\c{M}_1(X)$ & Space of probability measures over $X$ & \labelcpageref{ntn:space-of-measures} \\
$D_{\f{KL}}(\.||\.)$ & Kullback--Leibler divergence between probability measures & \labelcpageref{ntn:kl-div} \\
$W_{p,d}(\.,\.)$ & Wasserstein distance of $p$th order over distance $d$ & \labelcpageref{ntn:wass-dist} \\
\global\nodotfilltrue
$\N$ & Natural numbers, not including zero \\ 
$\Z$ & Integers \\
$\R$ & Real numbers \\ 
$\C$ & Complex numbers \\
$\R^d$ & Euclidean space of dimension $d$ \\ 
$\R^X$ & Space of functions from $X$ to $\R$ \\
$\v{a}$ & Euclidean vector (bold italic letters) \\ 
$\m{A}$ & Matrix (bold upface letters) \\
\global\nodotfillfalse
$\oplus$ & Direct sum of vector spaces & \labelcpageref{ntn:direct-sum} \\
$V^*$ & Dual space of a topological vector space & \labelcpageref{ntn:dual-space} \\
$\c{A}^*$ & Adjoint of a bounded linear operator & \labelcpageref{ntn:bounded-operators} \\
$\norm{\.}$ & Norm of a vector in a Banach space & \labelcpageref{ntn:banach-hilbert} \\
$\innerprod{\.}{\.}$ & Inner product between vectors in a Hilbert space & \labelcpageref{ntn:banach-hilbert} \\
$\dualprod{\.}{\.}$ & Duality pairing between topological vector spaces & \labelcpageref{ntn:dual-pair} \\
$L(V;W)$ & Banach space of bounded linear operators & \labelcpageref{ntn:bounded-operators} \\
$C^0(X;\R)$ & Banach space of real-valued continuous functions & \labelcpageref{ntn:conts-fns} \\
$L^p(X;\R)$ & Lebesgue space & \labelcpageref{ntn:leb-space} \\
\end{xltabular}

\section*{Bayesian learning}

\begin{xltabular}{\textwidth}{@{} p{1.625cm} X<{\unskip\kern-1pc\leaders\hbox{\makebox[1pc][r]{\makebox[0pc]{.}}}\hfill\kern0pc} @{} >{\raggedleft\arraybackslash}p{0.625cm} @{}}
$(\Theta,\mathit\Theta)$ & Measurable space representing quantity of interest & \labelcpageref{ntn:meas} \\ 
$(Y,\c{Y})$ & Measurable space representing the data & \labelcpageref{ntn:meas} \\
$A_{(\.)}$ & Measurable set & \labelcpageref{ntn:meas} \\
$\theta, y$ & Random variables & \labelcpageref{ntn:rv} \\ 
$\pi_{(\.)}$ & Probability measure & \labelcpageref{ntn:prob} \\ 
$f_{(\.)}$ & Probability density & \labelcpageref{ntn:density} \\
$(\.\given\.)$ & Jointly measurable stochastic process & \labelcpageref{ntn:jmsp} \\
$\pi_{(\. \given \.)}$ & Probability kernel & \labelcpageref{ntn:prob-ker} \\ 
$f_{(\.\given\.)}$ & Conditional probability density & \labelcpageref{ntn:density} \\
\end{xltabular}

\section*{Markov decision processes}

\begin{xltabular}{\textwidth}{@{} p{1.625cm} X<{\unskip\kern-1pc\leaders\hbox{\makebox[1pc][r]{\makebox[0pc]{.}}}\hfill\kern0pc} @{} >{\raggedleft\arraybackslash}p{0.625cm} @{}}
$S$ & State space & \labelcpageref{ntn:mdp} \\ 
$A$ & Action space & \labelcpageref{ntn:mdp} \\ 
$r(\.\given\.,\.)$ & Reward kernel & \labelcpageref{ntn:mdp} \\ 
$p(\.\given\.,\.)$ & Transition kernel & \labelcpageref{ntn:mdp} \\ 
$\pi$ & Policy & \labelcpageref{ntn:mdp-policy} \\ 
$V^{(\pi)}$ & Value function with respect to policy $\pi$ & \labelcpageref{ntn:mdp-value} \\
$R^{(\pi)}$ & Regret of policy $\pi$ & \labelcpageref{ntn:mdp-regret} \\
\end{xltabular}

\section*{Multi-armed bandits}

\begin{xltabular}{\textwidth}{@{} p{1.625cm} X<{\unskip\kern-1pc\leaders\hbox{\makebox[1pc][r]{\makebox[0pc]{.}}}\hfill\kern0pc} @{} >{\raggedleft\arraybackslash}p{0.625cm} @{}}
$X$ & Space of arms & \labelcpageref{ntn:mab} \\
$K$ & Number of arms & \labelcpageref{ntn:mab-number-arms} \\ 
$T$ & Maximum time & \labelcpageref{ntn:mab-regret} \\
$y$ & Observed stochastic rewards & \labelcpageref{ntn:mab} \\
$f(\.)$ & Expected rewards & \labelcpageref{ntn:mab} \\
$\eps$ & Observation noise & \labelcpageref{ntn:mab} \\ 
$R$ & Cumulative stochastic regret of a bandit algorithm & \labelcpageref{ntn:mab-regret} \\ 
$\Delta(\.)$ & Regret of a specific arm & \labelcpageref{ntn:mab-regret-arms} \\ 
$n_t$ & Cumulative number of arm pulls up to time $t$ & \labelcpageref{ntn:mab-regret-arms} \\
$f^+_t(\.)$ & Upper confidence bound at time $t$ & \labelcpageref{ntn:mab-ucb} \\
$\mu_t(\.)$ & Empirical mean rewards at time $t$ & \labelcpageref{ntn:mab-means} \\
$\sigma_t(\.)$ & Width of confidence bounds at time $t$ & \labelcpageref{ntn:mab-ucb} \\
$c_t$ & Scaling constant for confidence bound width at time $t$ & \labelcpageref{ntn:mab-ucb} \\
\end{xltabular}

\section*{Gaussian processes}

\begin{xltabular}{\textwidth}{@{} p{1.625cm} X<{\unskip\kern-1pc\leaders\hbox{\makebox[1pc][r]{\makebox[0pc]{.}}}\hfill\kern0pc} @{} >{\raggedleft\arraybackslash}p{0.625cm} @{}}
$\f{N}(\mu,\sigma^2)$ & Gaussian distribution with mean $\mu$ and variance $\sigma^2$ & \labelcpageref{ntn:norm-dist} \\
$\f{N}(\v\mu,\m\Sigma)$ & Multivariate Gaussian with mean $\v\mu$ and covariance $\m\Sigma$ & \labelcpageref{ntn:mvn-dist} \\ 
$k(x,x')$ & Positive semi-definite kernel & \labelcpageref{ntn:kernel} \\ 
$\m{K}_{\v{x}\v{x}'}$ & Kernel matrix with entries $k(x_i, x'_j)$ & \labelcpageref{ntn:kernel} \\
$\f{GP}(\mu,k)$ & Gaussian process with mean $\mu$ and covariance kernel $k$ & \labelcpageref{ntn:gp-dist} \\ 
$f$ & Prior Gaussian process & \labelcpageref{ntn:gp-model} \\ 
$\v\eps$ & Likelihood noise & \labelcpageref{ntn:gp-model} \\
$\m\Sigma$ & Covariance of likelihood noise & \labelcpageref{ntn:gp-model} \\
$(\v{x},\v\gamma)$ & Data used to condition the model $y_i = f(x_i) + \eps_i$  & \labelcpageref{ntn:gp-model} \\
$f\given\v{y}$ & Posterior Gaussian process conditioned on $\v{y} = \v\gamma$ & \labelcpageref{ntn:gp-model} \\ 
$n$ & Number of training data points & \labelcpageref{ntn:gp-complexity} \\ 
$n_*$ & Number of evaluation locations & \labelcpageref{ntn:gp-complexity} \\
$\tilde{f}$ & Approximate prior defined via a finite basis expansion & \labelcpageref{ntn:gp-approx-prior} \\
$\phi_i(\.)$ & Basis function used by approximate prior & \labelcpageref{ntn:gp-approx-prior} \\
$w_i$ & Random weight of basis function $\phi_i$ & \labelcpageref{ntn:gp-approx-prior} \\
$\ell$ & Number of basis functions used in approximate prior & \labelcpageref{ntn:gp-approx-prior} \\
$k(x_j,\.)$ & Canonical basis function & \labelcpageref{ntn:canonical-basis-fns} \\ 
$v_j$ & Random weight of canonical basis function $k(x_j,\.)$ & \labelcpageref{ntn:canonical-basis-fns} \\
$H_k$ & Reproducing kernel Hilbert space with kernel $k$ & \labelcpageref{ntn:rkhs} \\ 
$\v\phi$ & Finite-dimensional approximate feature map & \labelcpageref{ntn:approx-feature-map} \\ 
$m$ & Number of inducing points in sparse approximation & \labelcpageref{ntn:gp-num-inducing} \\

\end{xltabular}

\section*{Manifolds and graphs}

\begin{xltabular}{\textwidth}{@{} p{1.625cm} X<{\unskip\kern-1pc\leaders\hbox{\makebox[1pc][r]{\makebox[0pc]{.}}}\hfill\kern0pc} @{} >{\raggedleft\arraybackslash}p{0.625cm} @{}}
$X$ & Manifold & \labelcpageref{ntn:manifold} \\ 
$TX$ & Tangent bundle & \labelcpageref{ntn:tangent-bdl} \\ 
$T^*X$ & Cotangent bundle & \labelcpageref{ntn:cotangent-bdl} \\ 
$C^\infty(X)$ & Space of smooth real-valued functions & \labelcpageref{ntn:smooth-fns} \\
$\Gamma(TX)$ & Space of smooth sections & \labelcpageref{ntn:smooth-sections} \\
$d$ & Dimension of $X$ & \labelcpageref{ntn:manifold} \\ 
$f_*$ & Pushforward of a function $f$ & \labelcpageref{ntn:vector-pushforward} \\ 
$g(\.,\.)$ & Metric tensor of a Riemannian manifold & \labelcpageref{ntn:metric-tensor} \\
$\lap_g$ & Laplace--Beltrami operator induced by $g$ & \labelcpageref{ntn:laplace-beltrami} \\ 
$(\lambda_n, f_n)$ & Laplace--Beltrami eigenpair & \labelcpageref{ntn:laplace-beltrami-eigenpairs} \\ 
$\Phi(\lap_g)$ & Operator defined via functional calculus & \labelcpageref{ntn:functional-calculus} \\ 
$\c{W}_g$ & Gaussian white noise induced by $g$ & \labelcpageref{ntn:riemannian-white-noise} \\
$\m\lap$ & Graph Laplacian & \labelcpageref{ntn:graph-laplacian} \\ 
$\bc{W}$ & Standard Gaussian & \labelcpageref{ntn:graph-white-noise} \\
$\Phi(\m\lap)$ & Matrix defined via functional calculus & \labelcpageref{ntn:matrix-functional-calculus} \\ 
$F$ & Frame over $X$ & \labelcpageref{ntn:frame} \\ 
$\m{P}_F(\.)$ & Projection map induced by $F$ & \labelcpageref{ntn:frame-projection} \\ 
\end{xltabular}

\chapter{Introduction}
\label{ch:intro}

\lettrine{L}{earning} from experience in order to change behavior is one of the defining abilities of biological systems, which differentiates them from other kinds of systems found in the world.
Replicating the processes biological systems use to learn and adapt is a fundamental goal of science and technology.
To this end, the development of mathematical formalisms rich enough to capture the notion of learning is one of the crowning achievements of statistics, machine learning, and artificial intelligence.

One such formalism is the \emph{Bayesian} view of learning.
The idea behind Bayesian learning is to represent the information known about the quantity of interest using probability.
To do so, the relationship between the quantity of interest and the data is formalized as a joint probability distribution.
This gives rise to a conditional probability distribution describing what was learned about the quantity of interest by observing the data.

Bayesian learning fits naturally within a theory of \emph{decision}, which describes how an abstract decision system should select actions in pursuit of a goal.
This is done by learning how different actions affect pursuit of the goal, and selecting optimal actions consistent with what was learned.
By virtue of being probabilistic, such decision systems assess and propagate uncertainty, enabling them to balance what is already known with what could be learned by taking actions---a concept known as the \emph{explore-exploit tradeoff}.

The performance of a decision system can be evaluated by examining how quickly its decisions improve and become optimal.
A decision system's \emph{regret} is the reduction in its quality of decisions by virtue of not knowing the quantity of interest in advance.
In most non-trivial settings, one can show that some regret is inevitable: a decision-making system must make some degree of mistakes in order to learn.
A decision system is considered \emph{optimal} if its regret is within a constant factor of the best possible regret.

Decision systems with optimal or close-to-optimal regret require less data in order to solve their respective tasks, and are called \emph{data-efficient}.
Data-efficiency is a key concern in practical settings, where data-collection takes time and can be expensive.
By virtue of resolving explore-exploit tradeoffs in a manner amenable to regret analysis, the Bayesian formalism gives broad tools for constructing data-efficient decision systems.

The key limitation of the Bayesian approach is that it often leads to computational problems which are intractable.
Conditional distributions generally contain more information than actually needed to make optimal decisions, yet calculating them is largely unavoidable.
Probabilistic decision systems are thus most attractive in settings where their strengths---including data-efficiency, solid technical foundations, and amenability to analysis---can shine, while computational costs are kept under control.

In my view, \emph{Gaussian processes} are one such setting: they are powerful enough to model wide classes of unknown quantities of interest, yet their computational costs are generally polynomial.
Better yet, Gaussian-process-based decision systems have demonstrated excellent performance in real-world scientific applications.
Studying Gaussian processes is therefore a promising avenue towards improved understanding of Bayesian learning and Bayesian decision-making in pursuit of artificial intelligence.

The goals of this dissertation are twofold: (i) to make Gaussian processes easier to work with when used within larger decision systems, and (ii) to expand the set of settings where Gaussian processes can used, enabling construction of decision systems for applications not previously considered.
Contributions toward (i) include path-wise conditioning techniques studied in \Cref{ch:pathwise}, and contributions toward (ii) include non-Euclidean Gaussian processes studied in \Cref{ch:noneuclidean}.
Following these, \Cref{ch:discussion} concludes.

To pursue these goals, it is critically important that all of the concepts described in the preceding paragraphs be made into rigorous mathematics, so that the ideas described in the sequel ultimately reduce to definitions and implications, and not metaphor or opinion.
Together, we therefore begin by defining the key mathematical notions needed.

\subsection*{Contributions}

The work presented in this thesis is published as a series of papers.
My contributions to the individual works are primarily on the theoretical and methodological side, and are described below.

In \Cref{ch:pathwise}, we present pathwise conditioning techniques: this work is published as \textcite{wilson20,wilson21}.
My contributions include (i) development of the random-function-based formalism for describing pathwise conditioning of Gaussian processes, (ii) error analysis of basis-function-approximation-based pathwise sampling, (iii) re-interpretation of inducing point methods, and (iv) review of prior sampling methods.
All of these were developed jointly with the other authors. 

In \Cref{ch:noneuclidean}, we present Matérn Gaussian processes in non-Euclidean settings: this work is published as \textcite{borovitskiy20,borovitskiy21,jacquier21,hutchinson21}.
My contributions include (i) developing the differential-geometric and stochastic-partial-differential-equation-based formalisms for defining these processes, and (ii) describing computational techniques for working with these models in practice.
All of these ideas were developed jointly with the other authors.

\section{Bayesian learning}

The first concept we develop in depth along our path towards a mathematically precise understanding of statistical decision-making is \emph{Bayesian learning}---a mathematical formalism for reasoning about unknown quantities of interest on the basis of data.
Bayesian learning is a \emph{probabilistic} theory: a model specifies how the quantity of interest and the data depend on one another within a probability distribution.
Learning entails calculating how the distribution of the quantity of interest changes upon observing the data.

One of the key strengths of Bayesian theory is that it applies in wide generality, owing to the substantial scope of probability theory. 
In particular, one can study learning of quantities of interest that are function-valued using observed data consisting of pointwise function evaluations.
This, however, requires a non-elementary treatment, as one cannot rely solely on probability densities to define what conditional distributions are.
We therefore begin by recalling the necessary mathematical formalism.

For an overview of Bayesian methods from a model-building perspective, see \textcite{gelman14}, and for an overview from a mathematical perspective see \textcite{ghosal17,gine15}.

\subsection{Review of probability theory}
We adopt the language of measure-theoretic probability, which we now describe.
To ease presentation, we state the definitions together with useful ways of thinking about them.
We use \textcite{kallenberg06} as our standard reference, along with certain results from \textcite{bogachev07a,bogachev07b}.

\label{ntn:meas}
We say that \emphmarginnote{measurable space} is a pair $(Y,\c{Y})$ consisting of a set $Y$ and a $\sigma$-algebra $\c{Y}$ over $Y$.
A \emph{$\sigma$-algebra} is a set of subsets of $Y$ containing the space itself which is closed under countable unions, intersections, complements, and therefore set-theoretic monotone limits.
These can be reinterpreted as Boolean logical operations, so $\c{Y}$ can be thought of as the set of all true-false questions one can ask about elements of the set $Y$. 
These questions, then, are closed under \emph{and/or/not} operations and monotone limits thereof.

\parmarginnote{Product of measurable spaces}
Given two measurable spaces $(Y,\c{Y})$ and $(\Theta,\mathit\Theta)$, if we form the Cartesian product $Y \x \Theta$, then we can define the \emph{product $\sigma$-algebra} $\c{Y}\ox\mathit\Theta$ as the smallest $\sigma$-algebra containing all sets of the form $A_y \x A_\theta$ with $A_y \in\c{Y}$ and $A_\theta\in\mathit\Theta$.
\marginnote{Measurable subspace}
For a measurable subset $Y' \subseteq Y$, we can define $\c{Y}' = \{A_y \^ Y' : A_y \in\c{Y}\}$, which is also a $\sigma$-algebra, thereby making $(Y',\c{Y}')$ into a measurable space. 
We call $\c{Y}'$ the \emph{subset $\sigma$-algebra}.

If $Y$ is a topological space, the \emphmarginnote{Borel $\sigma$-algebra} $\c{B}(Y)$ is defined as the smallest $\sigma$-algebra containing the open sets in the topology.
Topological spaces, in turn, are sets equipped with additional structure enabling them to admit notions such as locality and convergence---we review these spaces in additional detail in \Cref{ch:noneuclidean}.

\parmarginnote{Measurable function}
A map $f : Y \-> Y'$ between measurable spaces $(Y,\c{Y})$ and $(Y',\c{Y}')$ is said to be \emph{measurable} if its preimage defines a map $f^{-1} : \c{Y}' \-> \c{Y}$ between the respective $\sigma$-algebras.
This intuitively means that true-false questions for the space $Y'$ can be asked and answered relative to those in $Y$.
On product spaces, a map $f : Y \-> \Theta \x \Theta'$ is measurable if its components are measurable, but a map $f : Y \x Y' \-> \Theta$ \emph{need not} automatically be measurable if $f(\.,y') : Y \-> \Theta$ and $f(y,\.) : Y' \-> \Theta$ are measurable for all $y$ and $y'$.

\label{ntn:prob}
A \emphmarginnote{probability measure} is a non-negative countably additive map $\pi_y : \c{Y} \-> \R$ satisfying $\pi_Y(Y) = 1$. 
This can be thought of as a map that takes a true/false question, and assigns a number indicating how close to true or false its answer is according to the measure---in this view, probability measures describe uncertainty.
A \emphmarginnote{probability space} $(\Omega,\c{F},\P)$ consists of a measurable space and probability measure.
The set $\Omega$ can be thought of as a space of abstract random numbers, with a random number generator described by $\P$.

\label{ntn:rv}
Given a probability space, we say that a \emphmarginnote{random variable} is a measurable map $y : \Omega \-> Y$.
A random variable, then, maps random numbers $\omega\in\Omega$ into the space $Y$.
The \emph{distribution}\marginnote{Distribution} of a random variable is defined as the \emph{pushforward measure} $\pi_y = y_* \P$ where $y_*$ is defined as $(y_* \P)(A_y) = \P(y^{-1}(A_y))$ for all $A_y\in\c{Y}$.
The probability of an event $A_y$, then, is determined by measuring the probability of random numbers under which $A_y$ occurs---measurability guarantees this is possible.

\parmarginnote{Notation for random variables}
In this work, we will generally \emph{not} adopt the standard convention of suppressing $\omega$-arguments of random variables from our notation, and will write such arguments explicitly in cases that other function arguments are also used.
Though this makes expressions slightly denser, it also avoids ambiguity and presents the mathematics more precisely.

\label{ntn:expectation-cov}
The \emphmarginnote{expectation} $\E(y)$ of a real-valued random variable $y$, if it exists, is defined as its integral with respect to $\P$. 
This can be thought of as the average value of $y$ under the random numbers generated from $\P$, with averaging performed via the algebraic structure of the reals.
This notion extends to vector-valued random variables.
The \emphmarginnote{covariance} of two real-valued random variables, if it exists, is defined as $\Cov(y,y') = \E(yy') - \E(y)\E(y')$.
From this, we define the covariance of a finite-dimensional random vector componentwise.

\parmarginnote{Equality of random variables}
There are multiple senses in which we can say random variables are equal.
We say that $y = y'$ \emph{surely} if they are equal as mathematical functions.
This notion is essentially never used, because it is exceedingly strong.
We say that $y = y'$ \emph{almost surely} if $\P(y \neq y') = 0$, in which case $y$ and $y'$ are equal as functions, except possibly on a set of probability zero.
We say that $y = y'$ \emph{in distribution} if $y_* \P = y'_* \P$, meaning their distributions are equal.

\parmarginnote{Existence of a random variable with given distribution}
For a given probability space $(\Omega,\c{F},\P)$, a random variable $y : \Omega \-> Y$ with distribution $\pi_y$ need not exist.
However, if it does not, there always exists another probability space $(\Omega',\c{F}',\P')$ and a measurable map $i : \Omega' \-> \Omega$ such that $\P = i_* \P'$, and for which a $\pi_y$-distributed random variable $y : \Omega' \-> Y$ \emph{does} exist.
We thus implicitly assume all probability spaces are large enough to ensure all random variables with prescribed distributions exist.

\label{ntn:prob-ker}
\parmarginnote{Probability kernel}
We will be interested in probability measures extended to allow them to be parameterized by other quantities.
A \emph{probability kernel} is defined to be a map $\pi_{y\given\theta} : \c{Y} \x \Theta \-> \R$ satisfying two conditions: (i) the map $\pi_{y\given\theta}(\.,\vartheta) : \c{Y} \-> \R$ is a probability measure for all $\vartheta\in\Theta$, and (ii) the map $\pi_{y\given\theta}(A_y,\.) : \Theta \-> \R$ is measurable for all $A_y\in\c{Y}$. 
In particular, this means that a probability kernel can be integrated against other measures.

\label{ntn:jmsp}
\parmarginnote{Conditional distribution}
Random variables can also be extended to parameterize them by other quantities.
A $\c{F}\ox \mathit\Theta$-measurable map $y\given\theta : \Omega \x \Theta \-> Y$ is called a \emph{jointly measurable stochastic process}---we largely eschew this terminology to emphasize the Bayesian formalism.
In particular, unlike most presentations of stochastic processes, we \emph{never} think of $\Theta$ as representing time.
Define the \emph{conditional distribution} of $y\given\theta$ to be $(y\given\theta)_*\P$, with the pushforward taken in the first argument---by \Cref{lem:rcrv-rvm-equiv}, this is a probability kernel.

\subsection{Bayes' Rule for probability measures}

The key idea behind Bayesian learning is to formalize \emph{learning} using the concept of \emph{conditional probability}.
This entails (i) building a model quantifying the relationship between the \emph{quantity of interest} $\theta$, and the \emph{data} $y$, and (ii) quantifying what is learned using Bayes' Rule.
We now begin to make these concepts precise in the language of measure-theoretic probability.

\begin{definition}[Bayesian model]
Let $Y$ and $\Theta$ be sets.
A \emph{Bayesian model} is a probability measure defined on $\Theta \x Y$.
\end{definition}

A model can be constructed in a number of different ways.
The most common technique is to specify two components: (i) a probability distribution describing what is known about the quantity of interest external to the data, and (ii) how the data relates to the quantity of interest.
These are called the \emph{prior distribution} and \emph{likelihood}, respectively.
For example, a simple Gaussian model is given by
\[
y_i \given \theta &\~[N](\theta, 1)
&
\theta &\~[N](0,1)
.
\]
In this model, $\theta$ is an unknown scalar assigned a standard normal prior, while $y_i\given\theta$ is assigned a Gaussian likelihood centered at $\theta$.
Our first goal is to make this notation into proper mathematics.

\begin{proposition}
A Bayesian model can be constructed in the following ways.
\1 Using measures: integrate the following two components.
\1 \emph{Prior}: a probability measure $\pi_\theta : \mathit\Theta \-> \R$.
\2 \emph{Likelihood}: a probability kernel $\pi_{y\given\theta} : \c{Y}\x\Theta \-> \R$.
\0 
\2 Using random variables: compose the following two components.
\1 \emph{Prior}: a random variable $\theta:\Omega\->\Theta$.
\2 \emph{Likelihood}: a jointly measurable stochastic process $y\given\theta: \Omega\x\Theta\->Y$.
\0 
\0 
Moreover, if one takes the pushforward of the composition of the obtained random variables with respect to $\P\x\P$, then the two constructions coincide.
\end{proposition}

\begin{proof}
For the former, define 
\[
\pi_{\theta,y}(A_\theta\x A_y) = \int_{A_\theta} \pi_{y\given\theta}(A_y\given\theta) \d\pi_\theta(\theta)
\]
which extends to the full product $\sigma$-algebra by \Cref{lem:cyl-prod}, giving the desired probability measure.
For the latter, define the random variable
\[
\gamma : \Omega\x\Omega &\-> \Theta \x Y
&
\gamma : (\omega,\omega') &\|> (\theta(\omega), (y\given\theta)(\omega',\theta(\omega)))
\]
and take $\pi_{\theta,y} = \gamma_* (\P\x\P)$. 
We now prove these expressions coincide.
Write 
\[
\pi_{\theta,y}(A_\theta\x A_y) &= \int_{A_\theta} \pi_{y\given\theta}(A_y\given\theta) \d\pi_\theta(\theta)
\\
&= \int_\Omega\1_{\theta(\omega)\in A_\theta} \pi_{y\given\theta}(A_y\given\theta(\omega)) \d\bb{P}(\omega)
\\
&= \int_\Omega\1_{\theta(\omega)\in A_\theta} \int_\Omega \1_{(y\given\theta)(\omega',\theta(\omega)) \in A_y} \d\bb{P}(\omega') \d\bb{P}(\omega)
\\
&= \int_\Omega\int_\Omega \1_{(\theta(\omega),(y\given\theta)(\omega',\theta(\omega))) \in A_\theta \x A_y} \d\bb{P}(\omega') \d\bb{P}(\omega)
\\
&= (\P\x\P)(\gamma \in A_\theta \x A_y)
\]
which follows using Tonelli's Theorem and \Cref{lem:cyl-prod}.
\end{proof}

These two components should be interpreted as follows: $\theta$ describes what is known about the quantity of interest external to the data, and $y\given\theta$ describes how the data $y$ relates to the quantity of interest.
More specifically, the likelihood describes how $y$ would be distributed if $\theta$ was known and equal to the conditioned value.

The condition that the pushforward of the composition of the prior and likelihood needs to be taken with respect to the product measure $\P\x\P$ should be interpreted as a kind of \emph{non-duplicate dependence} condition which ensures that $y\given\theta$ depends on $\theta$ only through its second argument, and not through the abstract random numbers $\omega$.

Given a Bayesian model, we can formalize the notion of what is \emph{learned} about $\theta$ from observing $y$ as its conditional probability distribution---note that this is meant in a \emph{distributional} sense, not a random-variable-based sense.
The formulation we use requires topological assumptions: we say that a topological space is \emph{Polish} if it is homeomorphic to a complete separable metric space.
Note that any two uncountable Polish spaces are Borel-isomorphic: see \textcite[Chapter 1]{villani08}, for discussion.
The key result is given as follows.

\begin{figure}
\begin{subfigure}{0.55\textwidth}
\includegraphics{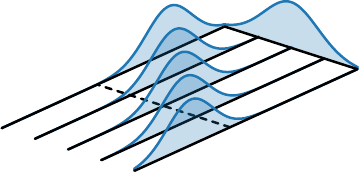}
\end{subfigure}
\begin{subfigure}{0.44\textwidth}
\tikzset{external/export next=false}
\begin{tikzpicture}
\begin{axis}[axis lines={none}, height={7.5cm}, axis equal, view={{310}{22.5}}, xmin={-0.6}, xmax={1.75}, ymin={-1.75}, ymax={1.75}, zmin={0}, zmax={1}]
\node at (0,0.03,0.31) {\includegraphics[scale=0.25]{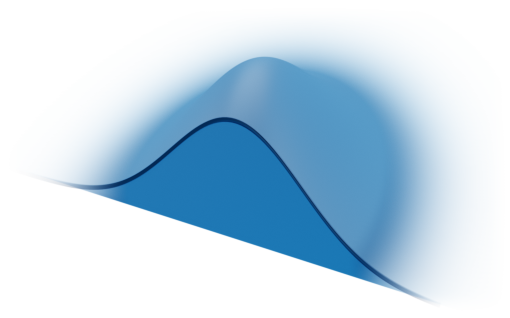}};
\addplot3[no markers, domain={-1.75:1.75}, smooth, samples={32}, samples y={0}, thick, color={black}, opacity={0}] ({-0.5},{x},{0.9 * exp(-x^2) * exp(-(x + 0.5)^2)});
\addplot3[no markers, very thick, color={black}, line cap={round}, dashed] coordinates {(-0.5,-1.75,0) (-0.5,1.75,0)};
\end{axis}
\end{tikzpicture}
\end{subfigure}
\caption[Bayesian models]{Here, we illustrate a Bayesian model. For all possible values of the quantity of interest, the likelihood describes the respective distribution of the data, shown on the left. This is combined with the prior, also shown on the left, to form a joint distribution, shown on the right. From this, the posterior distribution, also shown on the right, is obtained by conditioning. Note that the two plots are shown not to scale, and that the joint and posterior live on different spaces, hence are normalized differently.}
\label{fig:bayes-rule}
\end{figure}

\begin{result}[Bayes' Rule]
Suppose that $\Theta$ is a Polish topological space, and let $\pi_y = \pi_{\theta,y}(\Theta\x\.)$.
Then for every Bayesian model $\pi_{\theta,y}$ there is a $\pi_y\ae[-]$ unique probability kernel $\pi_{\theta\given y}$ satisfying
\[
\pi_{\theta,y}(A_\theta \x A_y) = \int_{A_y} \pi_{\theta\given y}(A_\theta \given y) \d\pi_y(y)
\]
which we call the \emph{posterior distribution}.
\end{result}

\begin{proof}
The claim follows directly from \textcite[Corollary 10.4.15]{bogachev07b}.
See also \textcite[Theorem 5.3 and Theorem 5.4]{kallenberg06}, \textcite[Theorem 5.3.1]{ambrosio08}, and \textcite{chang97} for variations of this result.
\end{proof}

This result is illustrated in \Cref{fig:bayes-rule}, and shows that given a Bayesian model, the posterior distribution describing what was learned from the data exists.
In its full abstract formulation, however, Bayes' Rule is \emph{non-constructive}, and it is not at all clear how to calculate any kind of useful formula from it.
Moreover, the null sets can be problematic: to ensure that $\pi_{\theta\given y}$ is defined pointwise, one needs further properties such as continuity.
Fortunately, in many settings these issues are resolved by virtue of additional structure.

\label{ntn:density}
The simplest such structure occurs when $\pi_{\theta,y}$ admits a density with respect to a product measure $\lambda$.
In this case, it follows from \Cref{lem:ref-density} that $\pi_\theta$ admits the density $f_\theta$ with respect to $\lambda(\.\x Y)$, and similarly for $\pi_y$ and $f_y$.
Define a \emph{conditional density} as the ratio of joint and marginal densities, for instance $f_{\theta\given y} = \frac{f_{\theta,y}}{f_y}$, and let $\propto$ denote equality up to a multiplicative proportionality constant.
Then, we have the following.

\begin{proposition}[Bayes' Rule for densities]
Suppose that $\pi_{\theta,y}$ admits the density $f_{\theta,y}$ with respect to $\lambda_{\theta,y}$.
Then we have
\[
f_{\theta\given y} \propto f_{y\given\theta}f_\theta
.
\]
\end{proposition}

\begin{proof}
We have
\[
f_{\theta,y} &= \frac{f_{\theta,y}}{f_\theta} f_\theta = f_{y\given\theta}f_\theta
&
f_{\theta,y} &= \frac{f_{\theta,y}}{f_y} f_y = f_{\theta\given y}f_y
\]
which, when combined, give the result.
\end{proof}

Sometimes, the knowledge of $f_{\theta\given y}$ up to proportionality is enough to fully deduce its form.
For example, if the likelihood is Gaussian with unknown mean and known variance, and the prior is Gaussian, then posterior is also Gaussian.
\marginnote{Conjugacy}
In cases such as this, where the prior and posterior land in the same parameterized class of distributions, the pair $(\pi_{y\given\theta}, \pi_\theta)$ are called \emph{conjugate}.

Bayes' Rule for densities is remarkable in its generality yet restrictiveness.
On the one hand, we require no direct assumptions about the spaces $\Theta$ and $Y$, and in particular allow for real spaces, discrete spaces, and Riemannian manifolds.
On the other hand, other than in the aforementioned settings, where we can employ the Lebesgue, counting, and Riemannian volume measures, respectively, finding a suitable reference measure can be difficult.

We will work with posterior distributions in infinite-dimensional vector spaces in the sequel---there, densities are either not available or not convenient.
In those cases, it is enough to calculate the posterior on arbitrary finite-dimensional marginal projections to uniquely determine its value on the full infinite-dimensional space.

\begin{proposition}[Conditioning and marginalization]
Conditioning and marginalization commute: given a probability measure $\pi_{\theta,\theta',y}$, if Bayes' Rule holds, then we have $\pi_y\ae[-]$ that
\[
\pi_{\theta \given y} = \pi_{\theta,\theta' \given y}(\. \x \Theta')
.
\]
\end{proposition}

\begin{proof}
Let $\pi_{\theta,\theta'\given y}$ be the $\pi_y\ae[-]$ unique probability kernel given by the Disintegration Theorem satisfying
\[
\pi_{\theta,\theta',y}(A_{\theta,\theta'} \x A_y) = \int_{A_y} \pi_{\theta,\theta'\given y}(A_{\theta,\theta'} \given y) \d\pi_y(y)
.
\]
Plugging in $A_{\theta} \x \Theta'$ for $A_{\theta,\theta'}$ gives
\[
\pi_{\theta,y}(A_\theta \x A_y) = \int_{A_y} \pi_{\theta,\theta'\given y}(A_\theta \x \Theta' \given y) \d\pi_y(y)
.
\]
On the other hand, applying the Disintegration Theorem to $\pi_{\theta,y}$ gives a $\pi_y\ae[-]$ unique probability kernel satisfying
\[
\pi_{\theta,y}(A_\theta \x A_y) = \int_{A_y} \pi_{\theta\given y}(A_\theta \given y) \d\pi_y(y)
.
\]
Since both $\pi_{\theta,\theta'\given y}$ and $\pi_{\theta,y}$ are defined with respect to the same null sets, we conclude by uniqueness that they coincide, and the claim follows.
\end{proof}

This result makes densities into a substantially more powerful tool than they would be otherwise, since it enables us to use them for calculating posterior distributions even where they are not directly available.
In particular, we can map an infinite-dimensional function space into finite-dimensional vector spaces induced by pointwise function evaluations at arbitrary points, enabling us to calculate posterior distributions in such settings, in spite of no suitable densities existing directly in the space of interest.

\label{ntn:space-of-measures}
We now introduce the \emph{variational formulation} of Bayes' Rule, which expresses the posterior as the solution to an infinite-dimensional optimization problem.
Let $\c{M}_1(\Theta)$ be the space of all probability measures over $\Theta$.

\label{ntn:kl-div}
\begin{proposition}[Bayes' Rule in variational form]
\label{prop:variational-bayes}
Let $\pi_{y,\theta}$ be a Bayesian model, and assume it admits the density $f_{\theta,y}$ with respect to a product measure $\lambda$.
Then for every $\gamma\in Y$, the posterior distribution satisfies 
\[
\pi_{\theta\given y}(\.\given \gamma) = \argmin_{\bb{q}_\theta \in \c{M}_1(\Theta)} D_{\f{KL}}(\bb{q}_\theta \from \pi_\theta) - \operatorname*{\E}_{\vartheta\~\bb{q}_\theta} \ln f_{y\given\theta}(\gamma\given\vartheta)
\]
where $D_{\f{KL}}$ is the Kullback--Leibler divergence between probability measures, and the minima does not depend on the choice of $\lambda$.
\end{proposition}

\begin{proof}
Write the quantity being minimized as
\[
D_{\f{KL}}(\bb{q}_\theta \from \pi_\theta) - \operatorname*{\E}_{\vartheta\~\bb{q}_\theta} \ln f_{y\given\theta}(\gamma\given\vartheta) &= \operatorname*{\E}_{\vartheta\~\bb{q}_\theta} \ln \frac{f_{\bb{q}}(\vartheta)}{f_\theta(\vartheta) f_{y\given\theta}(\gamma\given\vartheta)}
\\
&= \operatorname*{\E}_{\vartheta\~\bb{q}_\theta} \ln \frac{f_{\bb{q}}(\vartheta)}{f_{\theta\given y}(\vartheta\given\gamma) f_y(\gamma)}
\\
&= D_{\f{KL}}(\bb{q}_\theta \from \pi_{\theta\given y}(\.\given \gamma)) - \ln f_y(\gamma)
\]
Since $\ln f_y(\gamma)$ does not depend on $\bb{q}_\theta$, and since $D_{\f{KL}}(\mu \from \nu) = 0$ implies $\mu = \nu$, we conclude that the objective is minimized by taking $\bb{q}_\theta = \pi_{\theta\given y}(\.\given\gamma)$.
Since the minima does not depend on the choice of the measure $\lambda$ with respect to which the densities are defined, the claim follows.
\end{proof}

For any given dataset, this result shows that  Bayes' Rule can be viewed in \emph{information-theoretic} manner: among all probability measures, the posterior maximizes predictive power, while retaining as many bits as possible from the prior, in the sense given by the Kullback--Leibler divergence.

It's also possible to prove the result in a different way using the calculus of variations, with variations taken in the Banach space of signed measures under the total variation norm.
That argument also employs Bayes' Rule for densities for its key step, making it largely similar in spirit.

The result suggests a way to approximately compute posterior distributions: restrict the optimization problem to a suitably chosen subspace of the space of all probability measures $\c{M}_1(\Theta)$.
This strategy will be particularly fruitful in the later-described setting of Gaussian processes, where techniques for constructing such approximations with well-understood and favorable accuracy will be considered.

We \emph{always} view variational approximations as minimization of Kullback--Leibler divergences, as this is mathematically sound.
We will eschew standard presentations involving \emph{evidence lower bounds}: the only widely-known mathematical explanations for why maximizing these bounds should improve model performance appeal to Kullback--Leibler divergences---so, then, why talk about evidence lower bounds in the first place?

Once a posterior is calculated, the next step is to extract the relevant quantities from it.
Traditionally, this is often done by displaying summary statistics to be evaluated and interpreted by a person with statistical training, often with a focus on assessing uncertainty.
We will instead focus on settings where the posterior is given as input to an upstream decision-making algorithm: we explore these next.

\subsection{Technical lemmas}

We now prove a number of technical lemmas used in the preceding text, which for completeness are presented here in order to avoid disrupting the reader's flow.

\begin{lemma}
\label{lem:rcrv-rvm-equiv}
Let $b:\Omega\x A\->B$ be a jointly measurable stochastic process.
Then the map $b_*\P : \c{B} \x A \-> \R$, where the pushforward is taken in the first argument, is a probability kernel.
\end{lemma}

\begin{proof}
It is clear that $b_* \P$ is a probability measure for all $a'\in A$, so we only need to prove that the map $a \|> (b(\.,a)_*\P)(A_b)$ is measurable for all $A_b\in\c{B}$.
First, write
\[
(b_*\P)(A_b) = \P(b(\.,a)^{-1}(A_b)) = \int_\Omega \1_{b(\omega,a)\in A_b} \d\bb{P}(\omega)
.
\]
Now, note that the map $\Omega\x A\->\R$ given by $(\omega,a) \|> \1_{b(\omega,a)\in A_b}$ is bounded measurable, since $b$ is measurable in both arguments, and indicators of measurable functions on measurable sets are bounded measurable.
Finally, since for any bounded measurable $f : \Omega \x A \-> \R$, the map $a \|> \int_\Omega f(\omega,a) \d\bb{P}(\omega)$ is measurable by Fubini's Theorem, the claim follows.
\end{proof}

\begin{lemma}
\label{lem:semi-algebra}
Let $\c{S}$ be a \emph{semi-algebra of sets}, which is defined as a family of sets satisfying the following conditions.
\1 $\c{S}$ contains the empty set.
\2 $\c{S}$ is closed under finite intersections.
\3 The complement of any set in $\c{S}$ can be written as a finite union of disjoint sets in $\c{S}$.
\0 
Then any bounded countably additive non-negative function $\mu : \c{S} \-> \R$ satisfying $\mu(\emptyset) = 0$ extends uniquely to a measure defined on the smallest $\sigma$-algebra containing $\c{S}$.
\end{lemma}

\begin{proof}
Define the \emph{algebra of sets} generated by $\c{S}$ to be
\[
a(\c{S}) = \cbr{\U_{i=1}^n S_i : S_i \in \c{S} \t{disjoint}}
\]
and note that the smallest $\sigma$-algebra generated by $\c{S}$ obviously coincides with the smallest $\sigma$-algebra generated by $a(\c{S})$.
We claim every function $\mu : \c{S} \-> \R$ satisfying the given assumptions extends uniquely to a function on $a(\c{S})$.
Every element of $a(\c{S})$ can be written as a finite union of disjoint sets---using this, define 
\[
\mu^{(a)} : a(\c{S}) &\-> \R
&
\mu^{(a)}\del{\U_{i=1}^n S_i} &= \sum_{i=1}^n \mu(S_i)
.
\]
To ensure this is well-defined, we check that the definition is independent of the choice of which disjoint subsets to take the union of---suppose that $\U_{i=1}^n S_i = \U_{j=1}^m T_j$.
Then we have $S_i = \U_{j=1}^m S_i \^ T_j$ and similarly $T_j = \U_{i=1}^n S_i \^ T_j$: plugging these in to $\mu$ yields a double sum of disjoint sets and affirms well-definedness.
Next, note that $\mu^{(a)}(\emptyset) = \mu(\emptyset) = 0$.
To see that $\mu$ inherits countable additivity, take a sequence $S_1,S_2,.. \in a(\c{S})$ and note that
\[
\U_{n=1}^\infty S_n = \U_{n=1}^\infty \U_{m=1}^{p_n} T_{nm}
\]
where $T_{nm} \in \c{S}$ and $p_n \in \N$.
Plugging this into $\mu^{(a)}$ and applying countable additivity of $\mu$ gives the desired property.
We have thus obtained a uniquely defined countably additive function $\mu^{(a)} : a(\c{S}) \-> \R$ defined on an algebra of sets $a(\c{S})$ which extends $\mu$.
This function satisfies the assumptions of Carathéodory's Extension Theorem---see \textcite[Theorem A1.1]{kallenberg06} or \textcite[Theorem 3.1]{billingsley08}---applying this result gives the claim, where we note that uniqueness follows since the range of $\mu^{(a)}$ is bounded.
\end{proof}

\begin{lemma}
\label{lem:cyl-prod}
Let $\pi : \c{A} \x \c{B} \-> \R$ be a countably additive function with $\pi(\emptyset) = 0$.
Then $\pi$ extends uniquely to a measure on the product $\sigma$-algebra.
\end{lemma}

\begin{proof}
By \Cref{lem:semi-algebra}, it suffices to show that $\c{A} \x \c{B}$ is a semi-algebra of sets.
The first required property is immediate, the second and third properties follow by $(A \x B) \^ (A' \x B') = (A \^ A') \x (B \^ B')$ and $(A \x B)^c = (A^c \x B) \u (A \x B^c) \u (A^c \x B^c)$.
The claim follows.
\end{proof}

\begin{lemma}
\label{lem:ref-density}
Let $\pi_{a,b}$ be a measure which admits the density $f_{a,b}$ with respect to a product measure $\lambda_{a,b}$.
Then $\pi_a = \pi_{a,b}(\.\x B)$ admits the density $f_a$ with respect to $\lambda_a = \lambda_{a,b}(\.\x B)$, and similarly in the other argument.
\end{lemma}

\begin{proof}
By assumption and Tonelli's Theorem, we have 
\[
\pi_a(A_a) &= \pi_{a,b}(A_a \x B) 
\\
&= \int_{A_a \x B} f_{a,b}(\alpha,\beta) \d\lambda_{a,b}(\alpha,\beta)
\\
&= \int_{A_a} \int_B f_{a,b}(\alpha,\beta) \d\lambda_b(\beta)\d\lambda_a(\alpha) 
\\
&= \int_{A_a} f_a(\alpha) \d\lambda_a(\alpha) 
\]
where the desired probability density is $f_a(\alpha) = \int_B f_{a,b}(\alpha,\beta) \d\lambda_b(\beta)$.
\end{proof}

\section{Statistical decision-making}

We now use the Bayesian formalism to construct a probabilistic theory of decision-making.
To begin, we formalize the very general concept of an abstract agent making decisions in an environment in pursuit of some goal.
We use \textcite{sutton18,bertsekas19} as our references.

\subsection{Markov decision processes}
A Markov decision process is a stochastic system consisting of a set of states, actions, and transitions between states, together with a rewards that vary depending on states and actions.
This is defined as follows.

\label{ntn:mdp}
\begin{definition}[Discrete-time Markov decision process]
A \emph{discrete-time Markov decision process} is a $4$-tuple consisting of the following.
\1 State space: a measurable space $S$.
\2 Action space: a measurable space $A$.
\3 Reward: a probability kernel $r : \c{B}(\R) \x S \x A \-> \R$. 
\4 Transition kernel: a probability kernel $p : \c{S} \x S \x A \-> \R$.
\0 
\end{definition}

This is a very broad notion---however, a number of variations are also possible.
For instance, one can consider continuous-time, purely deterministic, and partially observed analogs---each of these involve their own subtleties and deserve study in their own right, but we do not pursue them here.

\begin{figure}
\tikzset{external/export next=false}
\begin{tikzpicture}
\begin{scope}[shift={(-1.5,2.125)}, xscale=-0.0085, yscale=0.0085, thick]
\clip[preaction = {draw,ultra thick}] (300, -395) -- (285, -355) .. controls (285, -355) and (269.142, -342.071) .. (255, -335) .. controls (245, -330) and (227.237, -322.04) .. (215, -320) .. controls (185, -315) and (175, -295) .. (175, -295) .. controls (175, -295) and (150, -295) .. (135, -285) .. controls (129.117, -281.078) and (100, -250) .. (100, -235) .. controls (100, -215) and (105, -205) .. (115, -195) .. controls (135.274, -174.726) and (144.617, -159.069) .. (186.751, -142.257) .. controls (230, -125) and (245, -125.009) .. (255, -125) .. controls (270.117, -124.987) and (289.478, -127.154) .. (305.786, -129.965) .. controls (335, -135) and (347.49, -144.511) .. (364.446, -158.01) .. controls (378.439, -169.15) and (389.863, -183.998) .. (397.853, -200) .. controls (405.048, -214.411) and (409.778, -230.664) .. (410, -246.769) .. controls (410.144, -257.217) and (410.264, -270.263) .. (402.959, -277.316) .. controls (395, -285) and (385, -290) .. (385, -290) .. controls (385, -290) and (377.69, -310.274) .. (375.386, -315.536) .. controls (371.018, -325.511) and (360.595, -330.936) .. (350.937, -334.628) .. controls (340.799, -338.503) and (318.451, -336.817) .. (318.451, -336.817) -- (326.08, -380.928) -- cycle;
\draw (300, -400) -- (300, -275);
\draw (275, -350) -- (275, -250) -- (225, -250);
\draw (250, -350) -- (250, -275);
\draw (225, -325) -- (225, -300) -- (200, -275) -- (100, -275);
\draw (325, -350) -- (325, -250) -- (300, -250);
\draw (400, -300) -- (350, -300) -- (350, -275) -- (375, -250) -- (375, -200) -- (400, -200);
\draw (350, -250) -- (350, -225) -- (250, -225) -- (250, -175) -- (225, -150) -- (150, -150);
\draw (350, -200) -- (300, -200) -- (300, -175);
\draw (400, -175) -- (325, -175) -- (300, -150) -- (300, -125);
\draw (200, -250) -- (150, -250) -- (125, -225) -- (75, -225);
\draw (175, -200) -- (225, -200) -- (225, -225) -- (175, -225);
\draw (225, -175) -- (150, -175) -- (150, -200) -- (100, -200);
\draw (275, -200) -- (275, -125);
\draw[fill=white] (200, -250) circle (7.5);
\draw[fill=white] (225, -250) circle (7.5);
\draw[fill=white] (250, -275) circle (7.5);
\draw[fill=white] (300, -275) circle (7.5);
\draw[fill=white] (300, -250) circle (7.5);
\draw[fill=white] (350, -250) circle (7.5);
\draw[fill=white] (350, -200) circle (7.5);
\draw[fill=white] (175, -225) circle (7.5);
\draw[fill=white] (175, -200) circle (7.5);
\draw[fill=white] (225, -175) circle (7.5);
\draw[fill=white] (300, -175) circle (7.5);
\draw[fill=white] (275, -200) circle (7.5);
\end{scope}
\begin{scope}[shift={(2.5,-1)}, scale=0.0125, very thick, line cap=round]
\draw (2, 2) -- (66, 2);
\draw (82, 2) -- (162, 2);
\draw (50, 18) -- (114, 18);
\draw (146, 18) -- (162, 18);
\draw (18, 34) -- (50, 34);
\draw (98, 34) -- (114, 34);
\draw (2, 50) -- (34, 50);
\draw (50, 50) -- (66, 50);
\draw (114, 50) -- (146, 50);
\draw (18, 66) -- (34, 66);
\draw (66, 66) -- (114, 66);
\draw (130, 66) -- (146, 66);
\draw (34, 82) -- (66, 82);
\draw (98, 82) -- (130, 82);
\draw (2, 98) -- (18, 98);
\draw (34, 98) -- (50, 98);
\draw (82, 98) -- (98, 98);
\draw (146, 98) -- (162, 98);
\draw (50, 114) -- (66, 114);
\draw (82, 114) -- (98, 114);
\draw (2, 130) -- (18, 130);
\draw (34, 130) -- (50, 130);
\draw (98, 130) -- (114, 130);
\draw (130, 130) -- (146, 130);
\draw (18, 146) -- (34, 146);
\draw (50, 146) -- (66, 146);
\draw (114, 146) -- (130, 146);
\draw (146, 146) -- (162, 146);
\draw (2, 162) -- (82, 162);
\draw (98, 162) -- (162, 162);
\draw (2, 2) -- (2, 162);
\draw (18, 18) -- (18, 34);
\draw (18, 66) -- (18, 82);
\draw (18, 98) -- (18, 114);
\draw (18, 130) -- (18, 146);
\draw (34, 2) -- (34, 18);
\draw (34, 50) -- (34, 66);
\draw (34, 82) -- (34, 114);
\draw (50, 18) -- (50, 34);
\draw (50, 50) -- (50, 82);
\draw (50, 114) -- (50, 130);
\draw (50, 146) -- (50, 162);
\draw (66, 34) -- (66, 66);
\draw (66, 82) -- (66, 114);
\draw (66, 130) -- (66, 146);
\draw (82, 18) -- (82, 50);
\draw (82, 66) -- (82, 98);
\draw (82, 114) -- (82, 162);
\draw (98, 34) -- (98, 66);
\draw (98, 98) -- (98, 114);
\draw (98, 130) -- (98, 162);
\draw (114, 18) -- (114, 34);
\draw (114, 82) -- (114, 130);
\draw (130, 2) -- (130, 34);
\draw (130, 50) -- (130, 146);
\draw (146, 34) -- (146, 50);
\draw (146, 82) -- (146, 98);
\draw (146, 114) -- (146, 130);
\draw (162, 2) -- (162, 162);
\end{scope}
\node[minimum size=85] at (-3.5,0) (a) {};
\node[minimum size=85] at (3.5,0) (e) {};
\node[anchor=north] at (a.south) {Agent};
\node[anchor=north] at (e.south) {Environment};
\draw[-latex,very thick] (a) to[bend left=22.5] node[midway, above] {action} (e);
\draw[-latex,very thick] (e) to[bend left=22.5] node[midway, below] {reward, next state} (a);
\end{tikzpicture}
\caption[Markov decision processes]{Illustration of the feedback loop induced by a Markov decision process. Here, the agent chooses an action, which results in the environment changing to a new state. The agent observes the new state, as well as the reward given by the previous state and action. The agent's goal is to select actions to maximize long-term rewards.}
\label{fig:mdp}
\end{figure}
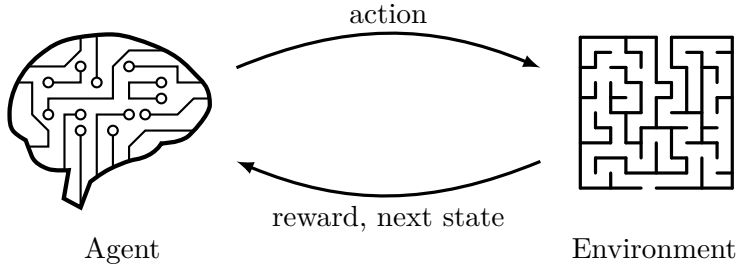

The idea behind this definition is that, at every point in time, the agent observes the current state, chooses an action $a \in A$, and obtains another state $s' \~ p(s\given a)$.
\Cref{fig:mdp} illustrates this.
Note that \emph{time} can, and often will, be part of the state $s$.
The agent's goal is to choose actions so as to control the full trajectory of states in order to obtain maximum rewards.
The choice of action in every state is called a \emph{policy}, and is formalized as follows.

\label{ntn:mdp-policy}
\begin{definition}[Policy]
Define the following.
\1 A measurable function $\pi : S \-> A$ is called a \emph{deterministic policy}.
\2 A probability kernel $\pi : \c{A} \x S \-> \R$ is called a \emph{Markov policy}.
\0 
\end{definition}

Markov policies include deterministic policies as a special case, by taking the conditional distributions to be Dirac and re-interpreting the given expressions appropriately.
As with Markov decision processes, one can also consider more general classes of policies, but we do not do so here.

Different policies yield different state trajectories, and therefore different rewards.
Of these, some obtain more rewards than others.
We therefore introduce notions for distinguishing between policies according to the rewards they obtain.

\label{ntn:mdp-value}
\begin{definition}[Optimal policy]
Let $T \in \N$ be the \emph{time horizon}.
A policy is called \emph{optimal} if it maximizes the \emph{value function} 
\[
V^{(\pi)}(s_0) = \E \sum_{t=0}^T r_t
\]
where $r_t \given s_t,a_t \~ r(s_t,a_t)$, $a_t \given s_t \~ \pi(s_t)$, and $s_{t+1} \given s_t,a_t \~ p(s_t, a_t)$.
\end{definition}

Finding an optimal policy therefore amounts to selecting the best possible actions to maximize expected total rewards.
Note that closely-related alternative notions of optimality, such as minimizing infinite discounted sums, or limits of averages, are also possible.
The most important distinction between different decision problems for finding optimal policies is given by what is assumed known.

\1 If $p$ and $r$ are known, we say we have an \emph{optimal control} problem.
\2 Otherwise, we say we have a \emph{reinforcement learning} problem.
\0 

These classes differ fundamentally from one another. 
Optimal control problems can be viewed as a class of structured optimization problems, where the goal is to compute $\pi$ by evaluating $r$ and $p$ as necessary.
Here, one generally proceeds by proving that $V^{(\pi)}$ and $\pi$ satisfy certain recursive equations, and developing schemes for solving them directly.

Reinforcement learning problems are more complex.
Due to the rewards or dynamics being \emph{unknown}, they cannot simply be maximized and their expectation must be \emph{learned}.
This forces one to consider whether to take advantage of actions known to be good, or to try others in case they might be better---this is known as the \emph{explore-exploit tradeoff}.
For such settings, we need an appropriate solution concept---to obtain one, define the following.

\label{ntn:mdp-regret}
\begin{definition}[Regret]
The \emph{regret} of a policy $\pi$ is defined as 
\[
R^{(\pi)}(s_0) = V^*(s_0) - V^{(\pi)}(s_0)
\]
provided that the value function $V^*$ with respect to an optimal policy exists.
\end{definition}

Minimizing regret is equivalent to maximizing value, but when $p$ and $r$ are unknown doing so directly is impossible.
Instead, the goal is to find an \emph{algorithm}---that is, a way of updating the policy based on observed data---so as to limit growth of regret.

In most settings, one can prove that every algorithm which does not know $p$ and $r$ necessarily incurs some level of regret.
This is done by exhibiting a randomized set of problems and rewards over which regret is lower-bounded in expectation for any algorithm.
In such a class, actions that perform well on one problem will necessarily perform badly on another problem.
Such arguments show that some degree of regret is inevitable.

On the other hand, some algorithms incur more regret than others.
The obviously-bad algorithm which learns nothing and chooses the exact same action over and over again incurs at most linear regret.
An algorithm is said to \emph{solve} a decision problem if its asymptotic regret rate with respect to $T$ matches the respective regret lower bound in the given problem class.
Finding such algorithms is of key interest.

One way to construct algorithms for solving decision problems is to employ a \emph{model-based} approach, which loosely speaking works as follows.

\1 Learn the unknown transitions and/or rewards from observed data using a supervised learning approach.
\2 Use the learned model(s) to find a policy satisfying some criteria.
\0 

The details of such approaches depend on the setting.
We distinguish between two key kinds of reinforcement learning problems.

\1 If $|S| = 1$, we say we have a \emphmarginnote{multi-armed bandit} problem.
\2 Otherwise, we say we have a general reinforcement learning problem.
\0 

Multi-armed bandits possess no variable state $S$, and thus only require one to learn the rewards to determine optimal actions.
In general reinforcement learning, actions can influence transitions between states---these problems require long-term planning, making them much more general, difficult, and important.
One can argue that as a mathematical theory, reinforcement learning is powerful enough to describe many aspects of human and animal intelligence, making it fundamentally interesting to study and develop.

We now restrict ourselves to the bandit setting, which is substantially easier to study and so can be understood more deeply.
Here, even when $A$ is a finite set and the rewards are Gaussian, the model-based approach consisting of (i) estimating rewards using empirical risk minimization and (ii) choosing the policy which maximizes rewards is known to be non-optimal.
This approach fails to explore, and can get stuck chasing inferior rewards.

One way to fix this problem is to replace empirical risk minimization with Bayesian learning, and adopt an appropriate decision rule for selecting actions.
Doing this yields approaches which can be shown optimal in many settings.
We therefore proceed to study multi-armed bandits in more detail.

\subsection{Multi-armed bandits}

The multi-armed bandit problem takes its name from a casino analogy.
In the 1950s, slot machines often had levers one could pull instead of buttons one could press, and were called \emph{one-armed bandits} for their ability to empty gamblers' wallets.
A \emph{multi-armed bandit} is a slot machine, which, for a fixed cost, allows one to pull an arm $x \in X$ and receive a random reward whose distribution depends on $x$.
The goal is to minimize  expected loss, or, equivalently, maximize total expected rewards.

Multi-armed bandits can be viewed as discrete-time Markov decision processes with a one-element state space, but this is not necessarily the most fruitful way to think about them.
We thus begin by introducing formalism and notation better suited to the given setting, which can be viewed as a special case of the notions considered previously.
We use \textcite{slivkins19,lattimore20} as references.

\label{ntn:mab}
\begin{definition}[Multi-armed bandit]
Let $f : X \-> \R$ be a bounded above measurable function, let $\eps : \Omega \x X \-> \R$ be a jointly measurable stochastic process such that $\E(\eps(x)) = 0$ for all $x$, and let $y(\omega,x) = f(x) + \eps(\omega,x)$.
Define the following.
\1 We say that the Markov decision process $(\{1\},X,y_*\P,\delta_{1})$, fully defined below, is a \emph{multi-armed bandit}.
\2 We say that a probability kernel  $\pi : \c{X} \x \bigoplus_{n=1}^\infty (X \x \R)^n \-> \R$ is a \emph{multi-armed bandit algorithm}.
\0
Here, $\delta_1$ is the Dirac measure centered at $1$ for all actions $x\in X$, which is the only possible conditional probability distribution over a one-element set, and $\bigoplus$ denotes the disjoint union of measurable spaces.
\end{definition}

An algorithm, then, assigns every dataset of arbitrary size to a probability measure describing what arms should be picked with what probability.
Each dataset consists of $(x, y)$ pairs where $x$ are the locations chosen by the algorithm, and $y$ are the noisy observed values---recall $\omega$ is the randomness used by the noise.
Some algorithms maximize $f$ faster than others---we consider this next.

\label{ntn:mab-regret}
\begin{definition}[Regret]
For a given multi-armed bandit, let $f(x^*)$ be the global maxima of $f$, which we assume to exist. 
Define the \emph{regret} of an algorithm $\pi$ at time $T$ to be
\[
R(\omega,T) = \sum_{t=1}^T f(x^*) - f(x_t(\omega))
\]
where $x_t \~ \pi(x_1,y_1,..,x_{t-1},y_{t-1})$, $y_t(\omega) = f(x_t) + \eps_t(\omega)$, and $\eps_t \~ \eps(\.,x_t)$.
\end{definition}

\begin{figure}
\begin{subfigure}{0.3\textwidth}
\includegraphics{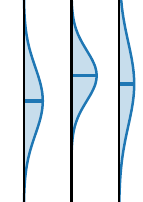}
\caption{Rewards}
\end{subfigure}
\begin{subfigure}{0.3\textwidth}
\includegraphics{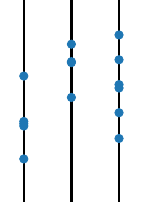}
\caption{Data}
\end{subfigure}
\begin{subfigure}{0.3\textwidth}
\includegraphics{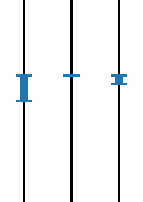}
\caption{Regret}
\end{subfigure}
\caption[Multi-armed bandits]{Here, we illustrate a simple three-armed bandit. Each of the three arms has its own reward distribution, shown on the left. These are unknown to the agent, which only sees the rewards obtained by actions it takes, shown in the center, and must use the obtained information to decide which arm to pick. Each arm's regret is given by its expected decrease in reward compared to the optimal arm, shown on the right.}
\label{fig:mab}
\end{figure}

Regret thus counts the total reward lost by virtue of not knowing the optimal arm in advance---this is illustrated in \Cref{fig:mab}.
Algorithms which achieve small regret therefore learn the optimal rewards effectively, and avoid getting stuck pulling bad arms.

Regret behavior in multi-armed bandit problems is strongly dependent on properties of the underlying function $f$, and in particular its domain $X$.
It is clear by considering for instance $X = \R$ that if $X$ is too large or too unstructured, no algorithm achieves better than linear regret.
We therefore begin studying the simplest non-trivial domain class.

\label{ntn:mab-number-arms}
\begin{definition}[$K$-armed bandit]
We say that a multi-armed bandit defined over a finite set with cardinality $K=|X|$ is a \emph{$K$-armed bandit}.
Moreover, if $y(\.,x)$ is a Bernoulli random variable for all $x$, we say it is a \emph{Bernoulli bandit}.
\end{definition}

For such bandits, regret can be decomposed on a per-arm basis in the manner given as follows.

\label{ntn:mab-regret-arms}
\begin{lemma}
\label{lem:regret-arms}
Regret satisfies 
\[
R(\omega,T) = \sum_{x=1}^K \Delta(x) n_T(\omega,x)
\]
where $n_T(\omega,x) = \sum_{t=1}^T \1_{x_t(\omega) = x}$ and $\Delta(x) = f(x^*) - f(x)$.
\end{lemma}

\begin{proof}
The claim follows directly from the definitions of $R$ and $n_T$ by grouping terms in the sum.
\end{proof}

What kind of performance is possible on such a problem?
One can ask and answer this question as follows.

\begin{theorem}
For any algorithm defined over a class of $K$-armed bandits there is an $f$ such that
\[
\E(R(\.,T)) \geq \Omega(\sqrt{KT})
.
\]
\end{theorem}

\begin{proof}
\textcite[Theorem 2.11]{slivkins19}.
\end{proof}

A similar lower bound, but where the right-hand-side depends on an extra term controlling the difficult of the problem, was originally proved by \textcite{lai85}.
Such bounds are called \emph{instance-dependent}: in that case, the rate obtained is logarithmic, owing to the presence of the extra term.
In our analysis, we focus on \emph{instance-independent} bounds such as the one presented above.

This result tells us that regret is necessarily incurred as consequence of not knowing the expected rewards of each arm given by $f$.
The next step, then, is to ask: is there an algorithm which achieves this rate?
We first consider simply evaluating each arm once, and then making choices according to the empirical averages.

\label{ntn:mab-means}
\begin{proposition}
For a Bernoulli bandit, the algorithm which chooses actions by first trying each arm out once, and then selecting arms according to the maximum empirical average
\[
x_{t+1} &= \argmax_{x\in X} \mu_t(x)
&
\mu_t(x) &= \frac{\sum_{t=1}^T \1_{y_t = 1} \1_{x_t = x}}{\sum_{t=1}^T \1_{x_t = x}}
\]
of the random observed data achieves linear regret.
\end{proposition}

\begin{proof}
We first show that it suffices to exhibit a bandit along with an event $S$, which we call the \emph{stuck event}, which occurs with constant probability and induces linear regret.
To see this, for an event $S$, write
\[
\E(R(\.,T)) &= \E(R(\.,T) \given S) \P(S) + \E(R(\.,T) \given S^c) \P(S^c) 
\\
&\geq \E(R(\.,T) \given S) \P(S)
.
\]
Now, take $S$ to be the event that during the initial first arm pulls, the optimal arm yields zero reward, while some other arm yields a reward: clearly $\P(S) > 0$.
On the other hand, $\E(R(\.,T) \given S) = \c{O}(T)$, because the optimal arm conditional on $S$ has empirical mean zero, which is always smaller than some alternative, and will never be selected again.
The claim follows.
\end{proof}

This algorithm fails to resolve the explore-exploit tradeoff, and gets stuck with suboptimal choices.
While the setup is obviously contrived---for instance, giving each arm some non-zero selection probability even if it previously gave bad rewards would seemingly break the lower bound---it also illustrates an important principle: the algorithm must explore.
The question, then, is how should it explore?
As an alternative, consider a simple uncertainty-based decision rule for selecting arms.

\begin{figure}
\begin{subfigure}{0.3\textwidth}
\includegraphics{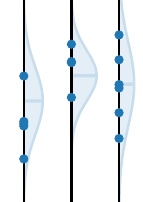}
\caption{Data}
\end{subfigure}
\begin{subfigure}{0.3\textwidth}
\includegraphics{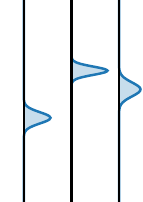}
\caption{Posterior}
\end{subfigure}
\begin{subfigure}{0.35\textwidth}
\includegraphics{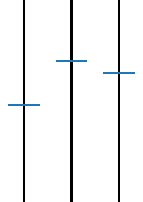}
\caption{Upper confidence bound}
\end{subfigure}
\caption[Upper confidence bound algorithm]{The idea behind the upper confidence bound algorithm is to use the observed data to construct a set of error bars that reflect what has been learned about the mean rewards.
Here, we construct these using the posterior distribution under a Gaussian prior and likelihood.
The next arm is chosen as the maximum of the quantile-based error bars.
This process is repeated iteratively as additional data is obtained, with the quantile becoming more strict over time, ensuring the bounds hold increasingly often.}
\label{fig:bb-ucb}
\end{figure}

\label{ntn:mab-ucb}
\begin{definition}[Hoeffding upper confidence bound algorithm]
Define the \emph{Hoeffding upper confidence bound} algorithm for Bernoulli bandits, which tries each arm once, and then, for a total of $T$ rounds, selects actions by
\[
x_{t+1} &= \argmax_{x\in X} f^+_t(x) 
&
f^+_t(x) &= \mu_t(x) + c_t\sigma_t(x)
\]
where $\mu_t$ are the empirical means, $\sigma_t(x) = \sqrt{\frac{1}{n_t(x)}}$, and $c_t = \sqrt{2\ln(T)}$.
\end{definition}

Many algorithms within the upper confidence bound class have been proposed \cite{lai85,agrawal95,auer02}: we focus on the variant analyzed by \textcite{auer02}, but illustrate an alternative in \Cref{fig:bb-ucb}.
We study the Hoeffding-based algorithm's regret, and present a simplified but less sharp analysis compared to the one given in that work.
It turns out the simple modification of adding a set of error bars, with a threshold growing in time and decreasing with data size, is enough to result in favorable regret behavior.

\begin{theorem}
The Hoeffding upper confidence bound algorithm achieves an expected regret of
\[
\E(R(\.,T)) \leq \tl{\c{O}}(\sqrt{KT})
\]
uniformly for all $f$, where $\tl{\c{O}}$ denotes asymptotics up to logarithmic factors.
\end{theorem}

\begin{proof}
We adapt the argument presented by \textcite{slivkins19}.
First, let $B$ denote the event that the \emph{bound holds}, namely $f^-_t(\.,x) \leq f(x) \leq f^+_t(\.,x)$ for all $x\in X$ and all $t \leq T$, where $f^-_t(\.,x) = \mu_t(\.,x) - c_t \sigma_t(\.,x)$, where all stochasticity is defined with respect to the data.
Under this event, the upper confidence bound is a true bound on the rewards.
Thus, we have
\[
\E(R(\.,T)) &= \E(R(\.,T) \given B) \P(B) + \E(R(\.,T) \given B^c) \P(B^c)
\\
&\leq \E(R(\.,T) \given B) + T \P(B^c)
\]
since $\P(B) \leq 1$, and a regret of $T$ is the maximum possible value as there are $T$ rounds and rewards are in $\{0,1\}$.
Our strategy will be to bound $\E(R(\.,T) \given B)$, while choosing the scaling factor $c_t$ in the width of the upper confidence bound to ensure that $\P(B^c)$ decays fast enough that the latter term is negligible.
We begin with the latter.
By the union bound, we have
\[
\P(B^c) &\leq \sum_{t=1}^T \sum_{x\in X} \P(p_x \leq f^-_t(\.,x)) + \P(p_x \geq f^+_t(\.,x))
\\
&\leq 2KT \max_{\substack{x\in X\\1 \leq t \leq T}} \P(p_x \geq f^+_t(\.,x))
\]
where $p_x$ is the Bernoulli parameter for arm $x$, and we have used symmetry of the error bars to combine terms.
For each $n > 0$, if $y$ is a sum of independent Bernoulli random variables, the probability we want to bound is
\[
\P\del{p_x \geq \frac{y(\.)}{n} + \sqrt{\frac{2\ln(T)}{n}}} \leq \exp\del{-2 \del{\sqrt{\frac{2\ln(T)}{n}}}^2 n} = \frac{1}{T^4}
\]
using Hoeffding's inequality.
Since holds uniformly in $n$, it follows that
\[
\P(p_x \geq f^+_t(\.,x)) = \P\del{p_x \geq \frac{\sum_{x_t = x} y_t(\.)}{n_t(\.,x)} + \sqrt{\frac{2\ln(T)}{n_t(\.,x)}}} \leq \frac{1}{T^4}
.
\]
We therefore conclude that 
\[
T\P(B^c) \leq \frac{2KT}{T^4} = \c{O}(1)
\]
which shows that the upper confidence bounds are exceeded sufficiently rarely that this possibility incurs no regret in the asymptotic limit, and completes this part of the argument.
We now proceed to analyze the $\E(R(\.,T) \given B)$ term---assume henceforth that $B$ holds.
We have that 
\[
\Delta(x_t(\omega)) &= f(x^*) - f(x_t(\omega))
\\
&\leq f^+_t(\omega,x^*) - f^-_t(\omega,x_t(\omega))
\\
&\leq f^+_t(\omega,x_t(\omega)) - f^-_t(\omega,x_t(\omega))
\\
&= 2\sigma_t(\omega,x_t(\omega))
\]
using the definition of the event $B$.
Now, if we consider the rescaled error bar width $c_t\sigma_t(\omega,x_t(\omega))$ of the arm chosen at time $t$, this is
\[
\Delta(x_t(\omega)) \leq \c{O}(c_t\sigma_t(\omega,x_t(\omega))) = \tl{\c{O}}\del{\frac{1}{\sqrt{n_t}}}
.
\]
By \Cref{lem:regret-arms}---which we note still holds conditional on $B$ via the exact same argument---we have
\[
\E(R(\.,T)\given B) &= \sum_{x=1}^K \Delta(x) n_T(\.,x) 
\\
&\leq \sum_{x=1}^K \tl{\c{O}}\del{\sqrt{n_T(x)}}
\\
&\leq \tl{\c{O}}\del{\sqrt{K} \sqrt{\sum_{x\in X} n_T(x)}}
\\
&= \tl{\c{O}}(\sqrt{KT})
\]
using that by definition, every $x$ for which $n_T(\omega, x) > 0$ is equal to $x_t(\omega)$ for some $t$, enabling us to apply the preceding inequality.
The claim follows.
\end{proof}

This shows that the proposed algorithm, which uncertainty built via the error bars, effectively balances exploration and exploitation in this setting.
The above behavior is not unique to the given form of the upper confidence bound, nor to the upper confidence bound rule itself.
For example, we can consider a variation of the above algorithm, where the width of the error bars is constructed via a Bayesian model.

\begin{definition}[Beta--Bernoulli upper confidence bound algorithm]
Define a Bayesian model via the likelihood $\gamma(x) \~[Ber](\mu(x))$ and prior $\mu(x) \~[Beta](a,b)$.
Define the \emph{beta--Bernoulli upper confidence bound} algorithm which selects actions by maximizing the function
\[
x_{t+1} &= \argmax_{x\in X} f^+_t(x) 
&
f^+_t(x) &= \mu_t(x) + c_t \sigma_t(x)
\]
where $(\mu_t, \sigma_t)$ are the mean and standard deviation of the posterior distribution $\mu\given\gamma(x_1) = y_1, .., \gamma(x_t) = y_t$, and $c_t$ is defined as previously.
\end{definition}

One can show this algorithm, illustrated in \Cref{fig:bb-ucb}, achieves the same regret as its Hoeffding-based analogue: the proof is similar, but instead uses the empirical Bernstein inequality, and is more messy owing to the complicated form of the upper confidence bound.
Even more generally, one can select arms by optimizing a function $\alpha : X \-> \R$, called an \emph{acquisition function}, built from a posterior distribution, confidence set, or other appropriate construction.
Many different acquisition functions have been proposed.

We have made no attempt to optimize the bound, and significant improvements are possible, particularly when considering variations of the upper confidence bound algorithm---see \textcite{lattimore20} for state-of-the-art results.
Of these, algorithms built using \emph{confidence sets} rather than posterior distributions, such as the variant presented previously, are particularly important in the $K$-armed bandit setting.
We focus on the Bayesian view because it generalizes well to more complex settings.

We now proceed to explore a more general setting which will enable us to employ the ideas developed so far to develop efficient black-box optimization algorithms.
This will enable us to use Bayesian methods to solve a broad class of decision problems of practical interest.

\subsection{Bayesian optimization}
\label{sec:bayesian-optimization}

We now describe the formalism of \emph{Bayesian optimization} for using ideas built on Bayesian learning and multi-armed bandits to design global optimization algorithms.
We use \textcite{frazier18} as our main reference.
Our goal now is to minimize a black-box function
\[
f : X \-> \R    
\]
which is assumed continuous, and defined on a compact set $X \subseteq \R^d$.
Such a function is automatically bounded.
Our goal is to minimize $f$ with as few evaluations as possible.
To measure performance, we again introduce a notion of regret.
One choice is to define
\[
R(T) = \sum_{t=1}^T f(x_t) - f(x^*)    
\]
where we have used the opposite sign convention compared to bandits and reinforcement learning, because our goal is to minimize $f$ rather than maximizing rewards.
Note that unlike before, for a deterministic algorithm this is now a purely deterministic quantity, and not a random variable.

As before, we can approach this problem by building a Bayesian model.
For an arbitrary sequence of points $x_1,..,x_t$, define the likelihood
\[
y_t(\omega) &= f(x_t) + \eps_t(\omega)
&
\eps_t &\~[N](0,\sigma^2)
.
\]
Here, the space of observed values is $\R$, and our quantity of interest is the actual \emph{function} $f$, which we at least implicitly view as an element of a space of functions, such as for instance the Banach space of continuous functions $C^0(X;\R)$.

\begin{figure}
\begin{subfigure}{0.98\textwidth}
\includegraphics{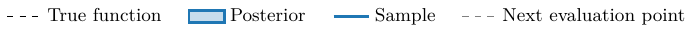}
\end{subfigure}
\begin{subfigure}{0.49\textwidth}
\includegraphics{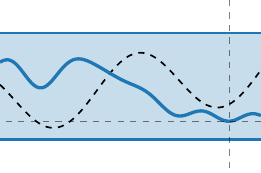}
\caption{$t = 1$}
\end{subfigure}
\begin{subfigure}{0.49\textwidth}
\includegraphics{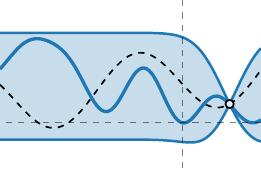}
\caption{$t = 2$}
\end{subfigure}
\begin{subfigure}{0.49\textwidth}
\includegraphics{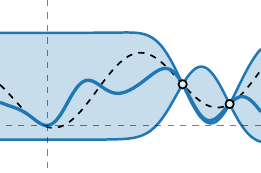}
\caption{$t = 3$}
\end{subfigure}
\begin{subfigure}{0.49\textwidth}
\includegraphics{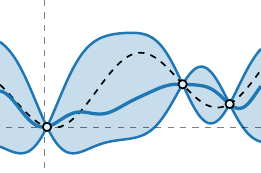}
\caption{$t = 4$}
\end{subfigure}
\caption[Bayesian optimization using Thompson sampling]{Bayesian optimization using Thompson sampling. 
Here, we sample a random function from the posterior distribution over possible functions, and choose the next evaluation point to be the minima of the sampled random function. 
As additional data is obtained, the posterior distribution concentrates around the true function. 
Unlike the upper confidence bound acquisition function considered previously, Thompson sampling makes use of a full probability distribution, rather than just a set of error bars.}
\label{fig:ts}
\end{figure}

It is now clear why we bothered with setting up a dense, abstract formalism for Bayesian learning in general measure spaces: this formalism is rich and powerful enough to enable us to properly define priors on spaces like this---we explore these in the sequel.
Suppose for the moment that this is possible.
Then, we can calculate the posterior distribution
\[
f \given y_1,..,y_t
\]
which is now a probability measure supported on $C^0(X;\R)$.
To determine the next point to query, we introduce and optimize an acquisition function.
Typical acquisition functions include the \emph{upper confidence bound} \cite{auer02} acquisition function considered previously, as well as \emph{Thompson sampling} \cite{thompson33,russo18}, first proposed half a century before Bayesian optimization was otherwise popularized. 
This is defined as a random acquisition function given by
\[
x_{t+1}(\omega) &= \argmin_{x\in X} \alpha_t(\omega,x)
&
\alpha_t&\~ f \given y_1,..,y_t
.
\]
We show an example of Bayesian optimization using Thompson sampling in \Cref{fig:ts}.
Another typical choice is given by \emph{expected improvement} \cite{mockus75,jones98,snoek12}, which is defined as
\[
x_{t+1} = \argmax_{x\in X} \E \max(0,f_t(\.,x) - f(x^*_t))
\]
where $f_t = f \given y_1,..,y_t$ is the posterior, and $x^*_t = \argmin_{t \in \{1,..,t\}} f(x_t)$ is location with the smallest value observed so far.
Many other effective acquisition functions, including \emph{probability of improvement} \cite{kushner64}, \emph{predictive entropy search} \cite{hernandezlobato14} and \emph{information directed sampling} \cite{russo14}, are also possible.

Regret analysis for all of these choices can be performed, and will in general depend on the detailed properties of the Bayesian model, acquisition function, and unknown function $f$ \cite{srinivas09}.
In particular, regularity and smoothness properties may play a role, as well as structure present in the domain $X$.
In settings where the function $f$ is unknown, it's also possible to analyze
\[
BR(T) = \E \sum_{t=1}^T f(\omega,x_t) - f(\omega,x^*(\omega))    
\]
where regret is now considered in expectation with respect to the prior---this is called the \emph{Bayesian regret}.
Some acquisition functions, such as Thompson sampling, admit particularly simple analyses in this setting \cite{lattimore20}.

This concludes our development and showcasing of decision-making algorithms.
We now proceed to develop the final piece of the puzzle not yet studied in detail: how to place priors on function spaces, in order to build Bayesian models for settings such as Bayesian optimization.

\section{Gaussian processes}

In the preceding section, considerations arising from Bayesian decision-making algorithms motivated us to find a way to place priors on spaces of functions $f : X \-> \R$.
Gaussian processes are a broad class of random functions capable of this.
In \Cref{fig:ts}, to perform Bayesian optimization using Thompson sampling, we used a Gaussian process model to define the posterior distribution over random functions, as needed by the decision rule.

To develop Gaussian processes, we will need to start with simpler notions and gradually work towards increasing levels of generality.
Gaussian processes are defined by the key property that no matter what angle one views them from, they yield Gaussian marginal distributions.
We therefore begin by studying properties of Gaussian random variables and Gaussian random vectors, which will be used as the basic building blocks for the general case.

The notion of a Gaussian process as a random function will turn out to be too restrictive for all of our settings of interest: it is not possible to view certain random variables, which do deserve to be called Gaussian processes, in this way.
The obstructions can be geometric or analytic in nature.
For example, a vector field on a manifold is not a vector-valued continuous function, it is a section of a vector bundle: what, then, should the term \emph{Gaussian} actually mean?
Difficulties also occur when considering white noise processes.

To handle these issues, we work with the notion of a Gaussian process $f : \Omega \-> V$ where $V$ is a real vector space equipped with additional structure arising from the setting at hand.
Gaussian process are fundamentally \emph{linear} objects which reflect this structure.
The simplest possible settings one can study arise when $V$ is smallest.

\1  Choosing $V = \{0\}$ to be the trivial vector space, there is exactly one $V$-valued random variable, whose distribution is the Dirac measure centered at $0$, which can trivially be called Gaussian.
This random variable is not very interesting, so we do not consider it further.
\2 Choosing $V = \R$ to coincide with the underlying scalar field yields the setting of \emph{Gaussian random variables}.
This is the next-simplest setting and the first one we explore in detail.
\3 Choosing $V = \R^d$ yields the setting of \emph{Gaussian random vectors}, whose components are multivariate Gaussian---or, equivalently, whose \emph{dot products} are all scalar-valued Gaussian.
\4 Choosing $V$ to be a space of functions $f : X \-> \R$ yields the setting of Gaussian processes, whose finite-dimensional marginals are multivariate Gaussian.
This setting is well-studied when $X$ is itself a Euclidean space: in \Cref{ch:noneuclidean}, we will examine cases where $X$ instead possesses various kinds of geometric structure.
\5 Finally, choosing $V$ to be a possibly infinite-dimensional topological vector space yields the most general setting we examine.
This level of generality will be important for two reasons: (i) to provide a formalism for studying stochastic partial differential equations, and (ii) to develop a coordinate-free notion of Gaussian random vectors that will be useful in differential-geometric settings.
\0 

In what follows, our goal will be to lay the groundwork for subsequent development.
Thus, we will not discuss Bayesian learning with Gaussian processes, which will instead be presented in \Cref{ch:pathwise}.
We use \textcite{lifshits12} as our standard reference.
We first study the scalar case, before generalizing to other settings.

\subsection{Review of vector spaces}
Since Gaussian processes are closely connected to vector spaces and related mathematical structures, here we review ideas from functional analysis that are useful in understanding them.
We use \textcite{lang12} as our reference on these topics.

\parmarginnote{Topological vector space}
A set equipped with a notion of addition and multiplication by scalars taking values in an underlying field is called a \emph{vector space}.
We work with vector spaces where the underlying field is the real numbers.
A \emph{topological vector space} is a vector space equipped with a \emph{topology} under which the vector operations are continuous---the topology gives rise to notions of \emph{convergence}, \emph{continuity}, \emph{compactness}, and many others.
Many interesting spaces of functions and generalizations thereof are topological vector spaces.

\label{ntn:banach-hilbert}
A \emphmarginnote{Banach space} is a topological vector space equipped with a \emph{norm}, which assigns a notion of size to all vectors, and induces a notion of distance given by a topologically complete metric, which in turn induces the space's topology.
A \emphmarginnote{Hilbert space}, similarly, is a topological vector space equipped with an \emph{inner product} which induces a complete norm.
Many, but not all, useful spaces of functions are Banach or Hilbert spaces.
Similarly, many, but not all, useful notions of convergence come from norms or inner products.

\parmarginnote{Basis}
Every vector space admits a \emph{Hamel basis}, which allows arbitrary vectors to be decomposed into linear combinations of linearly independent vectors in the basis.
Bases are generally highly non-unique.
The \emphmarginnote{dimension} of a vector space is the cardinality of such a basis, which does not depend on which basis is chosen.
Hamel bases of infinite-dimensional vector spaces are usually poorly-behaved, and alternative notions, such as that of a \emph{Schauder basis} of a Banach space, are often used instead.

\label{ntn:direct-sum}
Given two vector spaces $V$ and $W$, we can form the \emphmarginnote{direct sum} vector space $V \oplus W$ by taking the Cartesian product of both spaces, and defining vector operations for $V \oplus W$ componentwise using the vector operations on $V$ and $W$.
This yields a vector space whose dimension is the sum of the dimensions of $V$ and $W$.
If $V$ and $W$ are topological vector spaces, equipping $V \oplus W$ with the product topology makes it into a topological vector space.
In this work, we only consider finite direct sums.

\parmarginnote{Linear operator}
Maps between vector spaces are called \emph{linear operators} if they preserve the vector operations.
By linearity, continuous maps between Banach spaces are automatically bounded with respect to their inputs, and vice versa, so in this context we refer to boundedness and continuity interchangeably.
Given two respective bases, linear maps between finite-dimensional vector spaces can be viewed as matrices via their action on a vector's basis coefficients.
In doing so, composition of linear maps becomes matrix multiplication.

\label{ntn:dual-space}
\parmarginnote{Dual space}
For a topological vector space $V$, the space $V^*$ consisting of continuous linear maps $\phi : V \-> \R$ into the underlying scalar field is called its \emph{dual space}.
The dual space can be made into a vector space: for many topological vector spaces, including Banach and Hilbert spaces, it can canonically be assigned a topology making it into a topological vector space.
\marginnote{Functional}
We generally call elements of $V^*$ \emph{linear functionals}.

\label{ntn:bounded-operators}
\parmarginnote{Space of bounded linear operators}
The spaces of linear operators between two vector spaces can also be given the structure of a vector space.
Using this, if $V$ and $W$ are Banach spaces, one can define the space of $L(V;W)$ of bounded linear operators.
This space can be made into a Banach space via the \emphmarginnote{operator norm}, defined as the largest possible size of a unit-norm vector in $V$ after it is mapped into $W$ by the given operator. 
The \emphmarginnote{adjoint} $\c{A}^* : H \-> G$ of a bounded linear operator $\c{A} : G \-> H$ between Hilbert spaces is defined by $\innerprod{\c{A}g}{h}_H = \innerprod{g}{\c{A}^* h}_G$.

\label{ntn:conts-fns}
\parmarginnote{Space of continuous functions}
We now describe some particularly important function spaces.
The space of \emph{continuous functions} is denoted $C^0(X;\R)$: if $X$ is a compact set, this space can be equipped with the supremum norm to obtain a Banach space.
This is not the only useful topology on this space: in many settings, for instance, the topology induced by convergence of all bounded linear functionals is also used.
The space of infinitely differentiable functions is denoted by $C^\infty(X;\R)$: this space, in contrast, is generally not a Banach space.

\label{ntn:leb-space}
For a measure space $X$, the \emphmarginnote{Lebesgue space} $L^p(X;\R)$, defined for $1 \leq p < \infty$, is the Banach space of equivalence classes of measurable functions $f : X \-> \R$, identified up to almost sure equality, whose norm, given by integration of absolute $p$th powers, is finite.
Of these spaces, only $L^2(X;\R)$ is a Hilbert space.
The space $L^\infty(X;\R)$ is the Banach space of essentially bounded equivalence classes of functions equipped with the essential supremum norm.

\parmarginnote{Space of distributions}
Many useful topological vector spaces are not spaces of functions.
The \emph{space of distributions} $D'(X)$ is defined as the dual of the space $D(X)$ of infinitely differentiable functions with compact support, equipped with the topology of uniform convergence of the function and all derivatives on compact sets.
Locally integrable functions embed into $D'(X)$, so we can view it as a space of generalized functions, containing both functions and other, less regular elements such as the Dirac delta function, which is not a classical function.

\subsection{Gaussian random variables}

A Gaussian random variable is a map which takes in an abstract random number, and returns a real scalar.
The basic object from which other Gaussians will be constructed is the standard scalar Gaussian, defined as follows.

\begin{definition}[Standard Gaussian random variable]
A random variable $z : \Omega\->\R$ is called \emph{standard Gaussian} if it admits the Lebesgue density
\[
f(z) = \frac{1}{\sqrt{2\pi}} \exp\del{-\frac{z^2}{2}}
.
\]
\end{definition}

From this, we define general scalar Gaussians.

\begin{definition}[Gaussian random variable]
A random variable $y : \Omega\->\R$ is called \emph{Gaussian} if there are scalars $\mu, \sigma\in\R$, and a standard Gaussian $z$, such that
\[
y = \sigma z + \mu
.
\]
\end{definition}

\begin{figure}
\includegraphics{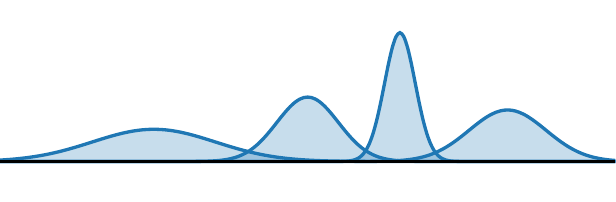}
\caption[Gaussian distributions]{Probability density functions for a set of Gaussian random variables, each with different mean and standard deviation parameters. All of these can be obtained by shifting and rescaling any of the others via affine transformations.}
\label{fig:norm}
\end{figure}

\label{ntn:norm-dist}
Note that we do \emph{not} require $\sigma \geq 0$: hence, every Gaussian random variable is determined uniquely in 
distribution by the pair $(\mu,\sigma^2)$, which we call its \emph{mean} and \emph{variance}, respectively. 
We write $y \~[N](\mu,\sigma^2)$.
True to these parameter names, we have
\[
\E(y) &= \mu
&
\E\del{(y - \mu)^2} &= \sigma^2
.
\]
A Gaussian random variable is called \emph{centered} if $\mu = 0$.
For a given variance $\sigma^2$ and standard Gaussian $z$, the expressions $\sigma z$ and $-\sigma z$ define two \emph{different} centered Gaussians with the same distribution.
At this stage, pointing these distinctions may appear needlessly pedantic: they will become more pronounced and important once we consider more general objects.
The density of a Gaussian random variable, if it exists, takes on a form analogous to that of a standard Gaussian, namely
\[
f(y) = \frac{1}{\sqrt{2\pi}\sigma} \exp\del{-\frac{(y-\mu)^2}{2\sigma^2}}
.
\]
This density is visualized in \Cref{fig:norm}.
Note that the density will not exist if $\sigma^2 = 0$: the distributions of such Gaussians are Dirac measures centered at $\mu$.
By examining this density, one sees that Gaussian random variables respect the additive and multiplicative structures of the reals.

\begin{proposition}[Affine maps between Gaussians]
Let $y \~[N](\mu,\sigma^2)$.
Then for $a,b \in \R$ we have that
\[
a y + b \~[N](a\mu+b, a^2\sigma^2)
.
\]
\end{proposition}

\begin{proof}
Immediate by definition.
\end{proof}

This compatibility with linear structure will be true at all levels of generality we consider.
We now lift this definition to construct multivariate analogs.

\subsection{Gaussian random vectors}

A multivariate Gaussian random vector is a random variable taking values in $\R^d$.
We write vectors defined in $\R^d$ in bold italics to emphasize this distinction, and similarly distinguish matrices from linear maps by writing the former in bold upface letters.
As before, we begin by defining a standard Gaussian.

\begin{definition}[Standard multivariate Gaussian]
A random variable $\v{z} : \Omega\->\R^d$ is called \emph{standard multivariate Gaussian} if its distribution is the product measure of the distributions of $d$ standard Gaussians.
\end{definition}

Once the notion of a standard Gaussian is available, we can again define multivariate Gaussians as transformations of standard Gaussians.

\begin{definition}[Multivariate Gaussian]
A random variable $\v{y} : \Omega\->\R^d$ is called \emph{multivariate Gaussian} if there is a vector $\v\mu\in\R^d$, matrix $\m{L}\in\R^{d\x d}$, and standard multivariate Gaussian $\v{z}$, such that
\[
\v{y} = \m{L}\v{z} + \v\mu
.
\]
\end{definition}

\begin{figure}
\begin{subfigure}{0.49\textwidth}
\includegraphics[scale=0.25]{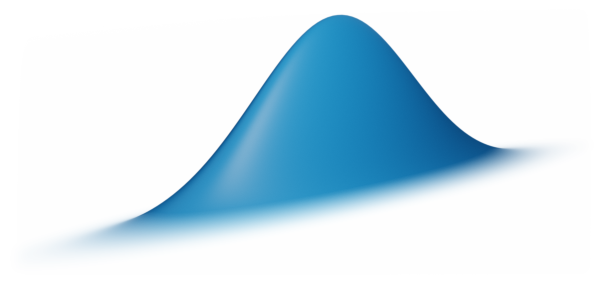}
\caption{$\rho = 0.9$}
\end{subfigure}
\begin{subfigure}{0.49\textwidth}
\includegraphics[scale=0.25]{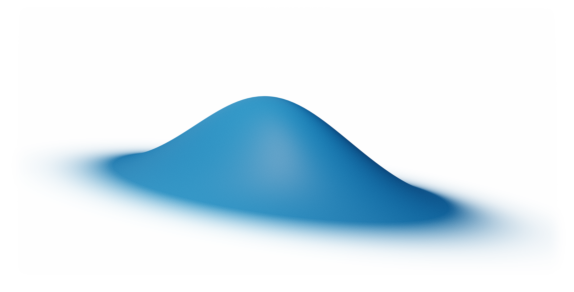}
\caption{$\rho = -0.6$}
\end{subfigure}
\caption[Multivariate Gaussian densities]{Two multivariate Gaussian densities, in dimension two, with unit variances and different correlation coefficients $\rho$. As $|\rho| \to 1$, the individual components of the multivariate Gaussian become more and more dependent.}
\label{fig:mvn-density}
\end{figure}

\begin{figure}
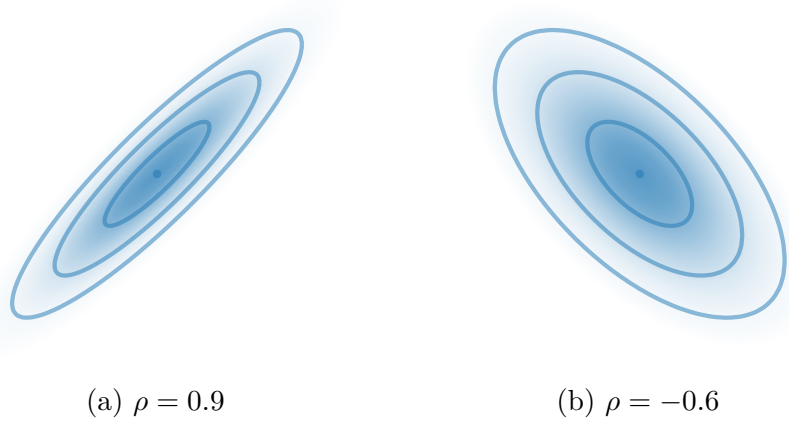

\begin{subfigure}{0.49\textwidth}
\input{figures/tex/mvn-pos.tex}
\caption{$\rho = 0.9$}
\end{subfigure}
\begin{subfigure}{0.49\textwidth}
\input{figures/tex/mvn-neg.tex}
\caption{$\rho = -0.6$}
\end{subfigure}
\caption[Multivariate Gaussian quantile ellipsoids]{Quantile ellipsoids for two multivariate Gaussian densities in dimension two with unit variances, different correlation coefficients $\rho$, and quantile levels equal to $0.8$, $0.95$, and $0.99$.}
\label{fig:mvn-contour}
\end{figure}

\label{ntn:mvn-dist}
Every multivariate Gaussian is determined by its \emph{mean vector} $\v\mu$ and positive semi-definite \emph{covariance matrix} $\m\Sigma = \m{L}\m{L}^T$, and, as before, is called \emph{centered} if $\v\mu = \v{0}$.
We write $\v{y}\~[N](\v\mu,\m\Sigma)$.
We can obtain a centered multivariate Gaussian with a given distribution by calculating a \emph{matrix square root} of $\m\Sigma$, multiplying it with a standard Gaussian.
The mean and covariance are
\[
\E(\v{y}) &= \v\mu    
&
\Cov(\v{y}) &= \E\del{(\v{y}-\v\mu)(\v{y}-\v\mu)^T} = \m\Sigma
\]
which mirror the previous situation.
If the determinant $|\m\Sigma|$ is non-zero, the density, displayed in two different ways in \Cref{fig:mvn-density,fig:mvn-contour}, is
\[
f(\v{y}) = \frac{1}{\sqrt{(2\pi)^d|\m\Sigma|}} \exp\del{-\frac{1}{2}(\v{y}-\v\mu)^T\m\Sigma^{-1}(\v{y}-\v\mu)}
\]
which now might not exist even if the distribution of $\v{y}$ is not Dirac.
In such cases, one can see that at least some eigenvalues of $\m\Sigma$ must be zero, and so Gaussians which do not admit densities must, when viewed in an appropriate basis, be products of Dirac measures with Gaussians which do admit densities.
Already in the multivariate case, then, we see the technical power of densities weakening: this will become more pronounced as we consider more general settings.
Affine maps, however, behave as before.

\begin{proposition}[Affine maps between multivariate~Gaussians]
Let $\v{y}\~[N](\v\mu,\m\Sigma)$. Then for $\m{A}\in\R^{d\x d}$ and $\v{b}\in\R^d$ we have 
\[
\m{A} \v{y} + \v{b} \~[N](\m{A}\v\mu + \v{b}, \m{A}\m\Sigma\m{A}^T)
.
\]
\end{proposition}

\begin{proof}
Immediate by definition.
\end{proof}

We now pause and reflect.
First, we distinguish $\R^d$ from a generic $d$-dimensional vector space: the former comes with a product structure $\R^d = \R \x .. \x \R$, including projection maps onto each coordinate, which in turn induce a \emph{canonical} choice of inner product given by the Euclidean dot product.
A generic finite-dimensional vector space lacks this structure: it admits many different inner products, and provides for no canonical choice.

With this in mind, we observe that we have not actually used the product structure of $\R^d$: suppose that $V$ is a finite-dimensional inner product space, and define
\[
y(\omega)  &= \sum_{i=1}^d z_i(\omega)  e_i
&
z_i \~[N](0,1)
\]
where $e_i$ is any orthonormal basis.
By noting that the choice of basis $e_i$ induces a Borel isomorphism $V \isom \R^d$, it is easy to see that this definition is basis-independent, since linear maps associated with changes of orthonormal bases are represented by orthogonal matrices.
This expression therefore defines a standard Gaussian with respect to the given inner product.

Subtle difference such as the ones considered become increasingly important in the sequel.
Therefore, we introduce an alternative definition to help build intuition for later.

\begin{definition}[Multivariate Gaussian (duality)]
A random variable $\v{y} : \Omega \-> \R^d$ is called \emph{multivariate Gaussian} if, for any vector $\v\phi \in \R^d$, the Euclidean dot product 
\[
\v\phi\.\v{y} : \Omega \-> \R 
\]
is univariate Gaussian.
\end{definition}

This definition turns out to be equivalent to the original one.

\begin{proposition}
The notions of multivariate Gaussians in the sense of transformations and in the sense of duality coincide.
\end{proposition}

\begin{proof}
Since the dot product is a linear map, it is clear that multivariate Gaussians in the sense of transformations are also Gaussian in the sense of duality.
To see the other direction, consider unit vectors $\v\phi$ where all coordinates except one are zero. 
By using dot products with such vectors to reassemble the mean vector and covariance matrix, one sees that the claim follows from eigenvalue factorization of positive semi-definite matrices.
\end{proof}

This definition allows one to begin imagining what a substantially more general notion of Gaussianity might look like: dot products become linear functionals, and covariance matrices become bilinear forms.
The preceding argument even suggests that such random vectors can be studied using spectral theory.
Of course, in infinite-dimensional settings, topological and analytic considerations come into play, making theory more difficult.
We develop these ideas subsequently, but first consider a simpler setting.

\subsection{Gaussian random functions}

We now consider Gaussian random functions, which are the first notion of a Gaussian random variable that is generally called a \emph{Gaussian process}.
Here, we will adopt a \emph{bottom-up} view which, from a technical perspective, departs slightly from the notions introduced so far.

Recall that for a set $X$, an $\R$-valued \emphmarginnote{stochastic process} is a map $f : \Omega \x X \-> \R$ measurable in its first argument.
If we have a set of points $x_1,..,x_n \in X$, we can plug them into $f$ to obtain a map $(f(\.,x_1),..,f(\.,x_n)) : \Omega \-> \R^n$, which, by virtue of being a product of measurable maps, is a random variable.
We call its distribution a \emphmarginnote{finite-dimensional marginal distribution}.
Using this notion, we define Gaussian processes.

\begin{figure}
\begin{subfigure}{0.49\textwidth}
\includegraphics{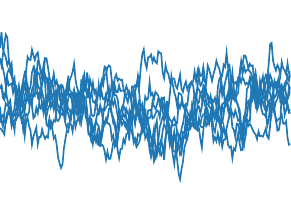}
\caption{Exponential}
\end{subfigure}
\begin{subfigure}{0.49\textwidth}
\includegraphics{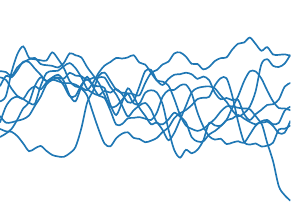}
\caption{Matérn-3/2}
\end{subfigure}
\begin{subfigure}{0.49\textwidth}
\includegraphics{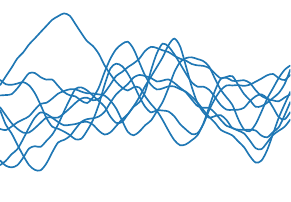}
\caption{Matérn-5/2}
\end{subfigure}
\begin{subfigure}{0.49\textwidth}
\includegraphics{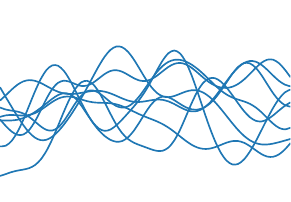}
\caption{Squared exponential}
\end{subfigure}
\caption[Gaussian processes]{Random functions generated from Gaussian processes with four different covariance kernels, which differ in particular according to their regularity, ranging from nowhere-differentiable in the exponential case to infinitely-differentiable in the squared exponential case.}
\label{fig:gp-smoothness}
\end{figure}

\begin{definition}[Gaussian process (stochastic process)]
Let $X$ be a set. 
A random process $f: \Omega \x X \-> \R$ is called a \emph{Gaussian process} if, for any finite set of points $x_1,..,x_n \in X$, the random variable $(f(\.,x_1),..,f(\.,x_n)) : \Omega \-> \R^n$ is multivariate Gaussian.
\end{definition}

Immediately upon writing this definition, one is left to wonder: does it actually make sense?
In particular, do Gaussian processes in the sense given here exist?

Under general conditions on the finite-dimensional marginals, Kolmogorov's Consistency Theorem states that there exists a unique probability measure on a suitable $\sigma$-algebra, called the cylindrical $\sigma$-algebra and defined below, whose finite-dimensional projections coincide with the distributions of the random variables written above.
Thus, Gaussian processes exist so long as a family of multivariate Gaussians satisfying the conditions of Kolmogorov's Consistency Theorem can be found.
We visualize such processes in \Cref{fig:gp-smoothness}.

The same results also allow us to reinterpret Gaussian processes as \emph{function-valued random variables}, which, in many ways, are a more natural point of view to take.
Define $\R^X = \{f : X \-> \R\}$ and equip it with the cylindrical $\sigma$-algebra, namely the smallest $\sigma$-algebra containing all sets of the form $\{f \in \R^X : f(x_1),..,f(x_n) \in A\}$, where $A$ is a measurable set on $\R^n$, for all $n$, to make it into a measurable space. 
Let $V \subseteq \R^X$, and equip it with the subset $\sigma$-algebra.
We can then reinterpret Gaussian processes as follows.

\begin{definition}[Gaussian process (random function)]
Let $X$ be a set. 
A random variable $f: \Omega \-> V \subseteq \R^X$ is called a \emph{Gaussian process} if, for any finite set of points $x_1,..,x_n \in X$, the random variable $(f(\.)(x_1),..,f(\.)(x_n)) : \Omega \-> \R^n$ is multivariate Gaussian.
\end{definition}

It is clear that every Gaussian process in the random process sense induces a Gaussian process in the random variable sense, and vice versa.
Thus, these two notions are simply two different ways of viewing the same object.

We can then ask, what properties of multivariate Gaussians are still true in this setting?
Since a Gaussian process is uniquely determined by its finite-dimensional marginals, we need to find functions that generate a family of mean vectors and covariance matrices which are positive semi-definite.
The former is straightforward: every function $\mu : X \-> \R$ can be evaluated at a finite set of points $x_1,..,x_n$ to obtain a mean vector $\v\mu = \mu(x_1,..,x_n)$.
The latter requires only slightly more thinking.

\label{ntn:kernel}
\begin{definition}[Positive semi-definite kernel]
A symmetric function $k : X \x X \-> \R$ is called a \emph{positive semi-definite kernel} if, for any finite set of points $x_1,..,x_n\in X$, the kernel matrix
\[
\begin{bmatrix}
k(x_1,x_1) & \dots &k(x_1,x_n)
\\
\vdots & \ddots & \vdots 
\\
k(x_n,x_1) & \dots & k(x_n,x_n)
\end{bmatrix}
\]
is positive semi-definite.
\end{definition}

We can recover such a kernel from a given Gaussian process.

\begin{definition}[Covariance kernel]
Define the \emph{covariance kernel} of a Gaussian process to be
\[
k(x,x') = \Cov(f(\.,x),f(\.,x'))    
.
\]
\end{definition}

\label{ntn:gp-dist}
It is then clear that the pair $(\mu,k)$ uniquely define a Gaussian process, and so we write $f\~[GP](\mu,k)$.
Defining a centered Gaussian process then amounts to defining a positive semi-definite kernel. 

In the Euclidean case, this can be done straightforwardly by noting that (i) the linear kernel $k(x,x') = \innerprod{x}{x'}$ is positive semi-definite by non-degeneracy of the inner product, (ii) that sums, powers, and limits of positive semi-definite kernels are positive semi-definite.
By this technique, we see that the widely-used exponential and squared exponential kernels are positive semi-definite.

For many spaces of functions $V \subseteq \R^X$, we can define addition and scalar multiplication, thereby making $V$ into a vector space.
If we do so, then, as before, affine maps preserve Gaussianity.

\begin{proposition}[Affine maps between Gaussian processes]
Let $f\~[GP](\mu, k)$, and equip $V$ and $W \subseteq \R^X$ with the topology of pointwise convergence.
Suppose that $f$ can be written as an infinite sum of deterministic basis functions as $f(\omega,x) = \sum_{n=0}^\infty w_i(\omega) f_i(x)$ where $w_i$ are independent Gaussian random variables.
If $\c{A} : V \-> W$ is a continuous linear map, and $b \in W$, then $\c{A} f + b$ is a Gaussian process.
\end{proposition}

\begin{proof}
It is clear by considering a set of finite-dimensional marginal distributions that adding $b$ preserves Gaussianity. 
Writing 
\[
(\c{A}f)(x) = \c{A} \del{\sum_{i=0}^\infty w_i(\omega) f_i(\.)}(x) = \sum_{i=0}^\infty w_i(\omega) (\c{A}f_i)(x)
\]
gives the statement for all finite-dimensional marginals, since sums and limits of Gaussian random variables are Gaussian. 
The claim follows.
\end{proof}

This is a substantially weaker assertion than previously: though we can conclude that affine maps of Gaussian processes are Gaussian, for a generic affine map we cannot immediately determine the form of the resulting kernel.
Moreover, the conditions are unnatural: by supposing them, we have more-or-less \emph{assumed} the resulting kernel to exist.
The problem here is that the notion of a \emph{kernel} is too rigid to permit a broad statement---an analogous property will hold if this is replaced with a different notion of covariance.

The situation for the other properties considered previously is even worse.
In particular, it is clear that, as an infinite-dimensional object, $f$ does not admit a probability density analogous to the ones considered previously, because an infinite-dimensional Lebesgue measure, in the sense of a locally finite translation invariant measure, does not exist.
It is also not clear what a \emph{standard} Gaussian process, nor what the analog of a matrix square root of a positive semi-definite kernel, might be.

The loss of these technical tools has consequences on what can be said about Gaussian processes.
In \Cref{ch:noneuclidean}, we will study Gaussian processes whose domain $X$ is a Riemannian manifold.
In that setting, one cannot begin by proving positive semi-definiteness of linear kernels, because there is simply no useful analog of this concept.
Defining positive semi-definite kernels there turns out to be non-trivial, and the kernels we study do not admit simple expressions like they do in the Euclidean case.

We therefore proceed to adopt a function-analytic perspective which is more technical, but significantly more powerful, and will allow us to recover the previous set of tools in a much more pure and abstract form.

\subsection{Gaussian processes in the sense of duality}
\label{sec:abstract-gp}

In the preceding section, we established a notion of a \emph{Gaussian process}, generalizing the notion of Gaussianity to random functions.
In developing this viewpoint, some of the appeal of Gaussianity was lost: in particular, unlike in the finite-dimensional setting, no simple covariance kernel transformation rule was available.
Here, we consider an alternative function-analytic view which avoids these limitations at cost of a higher degree of abstraction.

\label{ntn:dual-pair}
\begin{definition}[Gaussian process (duality)]
Let $V$ and $W$ be a pair of measurable real vector spaces equipped with a jointly measurable non-degenerate bilinear form $\dualprod{\.}{\.} : W \x V \-> \R$.
We say that a random variable $f : \Omega \-> V$ is a \emph{Gaussian process in the sense of duality} if, for any $\phi \in W$, the scalar-valued random variable $\dualprod{\phi}{f}$ is Gaussian.
\end{definition}

This is an exceedingly broad definition, which is more useful as an organizing framework connecting different concepts than as a technical tool in its own right.
At first glance, it is not clear what, if anything, this notion has to do with the Gaussian processes we have considered previously.
To better understand this, an illustrative example is in order.

Let $f \~[GP](\mu,k)$ be a Gaussian process defined on $[0,1]$ whose sample paths are almost surely continuous, and note by compactness that they are automatically bounded.
Our Gaussian process can therefore be viewed as a random variable $f : \Omega \-> C^0([0,1];\R)$. 
We endow the latter space with the supremum norm, making it into a Banach space.
If we take another function $\phi \in C^\infty([0,1],\R)$, called the \emph{test function}, we can define the pairing
\[
\dualprod{\phi}{f} = \int_0^1 \phi(x)f(x) \d x
\]
which by boundedness is almost surely finite.
Now, note that since sums of Gaussians are Gaussian, Riemann sums of Gaussian processes are Gaussian.
Since limits of Gaussians are Gaussian, the quantity $\dualprod{\phi}{f}$ will be a \emph{Gaussian} scalar.
Therefore, a Gaussian in the sense of duality can loosely be thought of as a Gaussian process whose integral against arbitrary test functions is always Gaussian---an adaptation of the dot product notion encountered previously to the infinite-dimensional setting.

Of course, the above only describes how one can intuitively think about Gaussians in the sense of duality.
For general Gaussians, there is no defined notion of an \emph{integral}---only of a dual pairing.
Such a Gaussian also need not be a random real-valued function: it's possible to consider random distributions and other objects generalizing the usual notion of a function.
It is therefore clear this view is very general.

This generality comes with a price: the resulting random vectors become more abstract, and it is much more difficult to understand whether or not they actually exist or say anything useful about them.
This difficulty is handled by specializing to less-general settings: for example, \textcite{bogachev98} studies the setting where $V$ is a locally convex topological vector space and $W$ is its dual, and \textcite{hairer09,lifshits12} study cases where $V$ is a Banach or Hilbert space.
In those cases, a number of results are available.

Provided we are considering a Gaussian $f$ which does exist, what objects play the role of a mean and covariance, and uniquely characterize $f$?
Since the canonical pairing $\dualprod{\.}{\.}$ is linear and non-degenerate, we know that $f = g$ holds if and only if $\dualprod{\phi}{f} = \dualprod{\phi}{g}$ for all $\phi\in W$, with both equalities used in the same sense.
It is therefore clear that the \emph{mean} of a Gaussian process is simply a vector $\mu\in V$, since such a vector uniquely determines all expectations $\E\dualprod{\phi}{f}$.
The covariance is only slightly more subtle.

\begin{definition}[Covariance form]
We say that a symmetric positive semi-definite bilinear form $k : W \x W \-> \R$ is a \emph{covariance form}.
\end{definition}

As before, we can construct a covariance form from a given Gaussian process.

\begin{definition}[Covariance form of a Gaussian process]
The \emph{covariance form} of a Gaussian in the sense of duality is defined by 
\[
k(\phi,\psi) = \Cov(\dualprod{\phi}{f},\dualprod{\psi}{f})
.  
\]
\end{definition}

It is clear that two Gaussian processes $f$ and $g$ are equal in distribution if and only if they have the same mean and covariance form.
This can be seen by noting these requirements force $\dualprod{\phi}{f} = \dualprod{\phi}{g}$ to hold for all $\phi\in W$.

We now ask: what relationship does the covariance form have with the covariance kernel defined previously?
Consider again our Gaussian process $f$ defined on the space of continuous functions. 
For any pair of test functions $\phi,\psi \in C^\infty([0,1];\R)$, the symmetric positive semi-definite bilinear form
\[
(\phi,\psi) \|> \int_0^1\int_0^1 \phi(x)k(x,x')\psi(x') \d x \d x'
\]
is the covariance form of $f$.

So far, this perspective mirrors the preceding ones, albeit with a more abstract presentation.
We have only introduced definitions.
These definitions, however, suffice to recover a notion of affine maps.

\begin{proposition}[Affine maps between Gaussian processes]
Let $f \~[GP](\mu, k)$. For $\c{A} : V \-> V'$ and $b\in V'$ we have 
\[
\c{A} f + b \~[GP](\c{A}\mu + b, k(\c{A}^*(\.),\c{A}^*(\.)))
\]
where $V'$ and $W'$ are a dual pair, and $\c{A}^* : W' \-> W$ is the adjoint operator defined by $\dualprod{\c{A}^*\phi}{v} = \dualprod{\phi}{\c{A}v}$.
\end{proposition}

\begin{proof}
Immediate by definition.
\end{proof}

Suppose now that $V$ is a locally convex topological vector space, and $W = V^*$ its topological dual.
Since the covariance form is a map $k : V^* \x V^* \-> \R$, it gives rise to an operator $\c{K} : V^* \-> V^{**}$, called the \emphmarginnote{covariance operator}, which in the case that $V$ is reflexive becomes an operator $\c{K} : V^* \-> V$ via composition with the canonical isomorphism given by reflexivity.
For our running example, this operator is given by 
\[
\phi \|> \int_0^1 \phi(x)k(x,\.) \d x
.
\]
One can easily see that the transformation rule for covariance operators under affine maps is given by
\[
\c{K} \|> \c{A} \c{K}\c{A}^*
.    
\]
If it is often unclear whether or not a Gaussian process with a given covariance form exists, it can be even less clear whether or not a Gaussian process with a given covariance operator exists.
Still, this notion is important, as these operators can potentially be studied using spectral theory.
We now move to the final concept we consider at this stage: that of a \emph{standard Gaussian}.

\begin{definition}[Gaussian white noise]
Let $W \subseteq H$ where $H$ is a Hilbert space.
We say that $\c{W} : \Omega \-> V$ is a \emph{Gaussian white noise} process if its covariance form coincides with the inner product in all cases where the former is defined.
\end{definition}

In cases where $\c{K}$ admits a sufficiently rich spectral theory, this can potentially provide a suitable notion of an \emph{operator square root} which relates Gaussian white noise to other Gaussian processes.

With these definitions in hand, we have recovered the two technical notions lost when transitioning from multivariate Gaussians to Gaussian processes in the sense of random functions.
The introduced machinery gives a way of constructing Gaussian processes that does not rely on defining a kernel: (i) construct a Gaussian white noise process, and (ii) define the Gaussian process of interest as an affine map of white noise.

We employ a variation of this strategy in \Cref{ch:noneuclidean} to construct Gaussian processes on Riemannian manifolds, where defining a kernel directly leads to difficulties.
Loosely speaking, we do so by taking $V$ to be a space of distributions, and $\c{A}$ to be the inverse of a differential operator.
This is chosen to ensure that $\c{A}$ smooths its inputs, so that its output is regular enough to be understood as a random function.

This line of thought leads one to the theory of stochastic partial differential equations driven by Gaussian white noise.
The main point swept here under the rug is that there is a convenient way to sidestep much of the analysis needed to carry out the above calculations using the theory of reproducing kernel Hilbert spaces.
Roughly, this amounts to instead working with certain vector spaces that uniquely determine the Gaussian processes of interest.

We also use the theory of Gaussians in the sense of duality for a second purpose: to construct a \emph{coordinate-free} notion of Gaussianity as a suitable building block for constructing Gaussian vector fields on Riemannian manifolds.
The issue here is that the Gaussian process \emph{cannot} be understood as a real-valued random function due to topological obstructions---it is instead a \emph{random section}, which we define in \Cref{ch:noneuclidean}.
Finite-dimensional Gaussians in the sense of duality end up being the right tool for this setting.

For this, we prove a general existence theorem on Gaussians in the sense of duality in finite-dimensional settings.
Aside from giving an intrinsic reinterpretation of the multivariate Gaussians described previously, this ensures our framework is suited for its purpose in the coordinate-free setting.

\begin{proposition}
Let $V$ be a finite-dimensional real topological vector space, and let $W = V^*$ be its topological dual.
Then for any vector $\mu \in V$ and covariance form $k : V^* \x V^* \-> \R$, there exists a unique-in-distribution random vector $y \~[N](\mu, k)$.
\end{proposition}

\begin{proof}
To prove this, choose a basis $e_i$ on $V$, and let $e^i$ be the dual basis. 
Let $\c{E} : V \-> \R^d$ be the continuous linear isomorphism induced by the basis, and define $y = \c{E}^{-1} \v{y}$ with $\v{y}\~[N](\v\mu,\m{K})$ defined by 
\[
\v\mu &= \begin{bmatrix}
\dualprod[0]{e^1}{\mu}
\\
\vdots
\\
\dualprod[0]{e^d}{\mu}
\end{bmatrix}
&
\m{K} &= \begin{bmatrix}
k(e^1, e^1) & \hdots & k(e^1, e^d)
\\
\vdots & \ddots & \vdots
\\
k(e^d, e^1) & \hdots & k(e^d, e^d)
\end{bmatrix}  
.  
\]
It is clear by direct calculation using Gaussians on $\R^d$ that the resulting vector is Gaussian with the right mean and covariance form.
The claim follows by noting that the assumed non-degeneracy of the dual pairing forces the distribution of every Gaussian in the sense of duality to be uniquely determined by its mean and covariance form.
\end{proof}

Here, we see the key difference between the coordinate-free view and the matrix-vector view considered previously: the Gaussian random vector can be viewed as a real-valued multivariate Gaussian in any basis, but itself is defined on $V$ independent of this choice.
The value of considering this distinction in the first place will become clear in \Cref{ch:noneuclidean}.
Here, finite-dimensionality suffices to ensure existence: the story in infinite-dimensional settings is completely different and requires case-by-case analysis.

To conclude, we reflect on the introduced ideas. 
We began by studying Gaussian random variables and random vectors, before generalizing these to Gaussian random functions, which offered a concrete framework where certain aspects of Gaussianity were seemingly lost.
Adopting a function-analytic view restored these aspects, at cost of increased abstraction.
This also gave a coordinate-free way to reinterpret the preceding multivariate constructions.

While much of the technical power of the duality framework is extraneous for our purposes, studying it nonetheless helps provide a unified conceptual perspective from which to interpret our developments.
By considering these notions, it becomes much clearer how one should understand the constructions encountered later, which might otherwise appear as if they arise out of thin air.
This completes our study of Gaussianity for its own sake, independent of Bayesian learning and other machine-learning-related considerations.

\section{Discussion}

The preceding sections paint a rich and detailed picture of what a mathematical theory of decision-making under uncertainty looks like.
We now recap the steps taken so far, and reflect on them, before proceeding to describe contributions to be presented.

We began with the concept of probability, constructed in the language of measure theory. 
We used this language to formalize the concept of learning via the notion of conditional probability, thereby obtaining the theory of \emph{Bayesian learning}.
By working in an abstract measure-theoretic setting, we obtained a formalism suitable for learning about very general unknown quantities of interest.

We then took a step back, examining how to formalize the notion of an agent selecting actions in an unknown environment on basis of interactions, obtaining the key concept of a \emph{Markov decision process}.
We then immediately restricted to the simpler setting of \emph{multi-armed bandits}.
We saw that model-based algorithms built atop Bayesian learning yielded decision systems that perform effectively.
With these notions, we described how to efficiently solve global optimization problems using \emph{Bayesian optimization}.

To transform the preceding ideas into a workable class of methods, we proceeded to study \emph{Gaussian process} models in depth.
We developed these models in sequence, starting from the simplest settings, and ending with the highest generality.
These ideas provide us with key tools for understanding Gaussian processes, so that we can use them as building blocks of Bayesian models and high-performance decision systems atop those models.

It is worth pausing to reflect on the merits of the setting chosen, within a broader context of artificial intelligence.
In choosing to work with Bayesian learning, we opted to represent uncertainty using probability---a powerful but computationally limiting choice.
This choice was counterbalanced by working with simple models in bandit-like settings, and is most effective when the decisions of interest must be made in a data-efficient manner that only algorithms with near-asymptotically-optimal regret can achieve.

Not all settings fit these criteria well.
In many reinforcement learning problems of interest in robotics, the complexity of the dynamics---which, for multi-armed bandits, are totally absent from the problem---is a key difficulty.
Gaussian processes are largely not expressive enough to represent multi-object collision dynamics and related phenomena.
We have also not addressed partial observations---another key difficulty in that setting.

On the other hand, no other currently known theoretical framework comes close to understanding decision to the degree of command one can obtain from the notions described.
In the absence of a probabilistic framework, it is unclear how to assess, represent, and propagate uncertainty to resolve explore-exploit tradeoffs and minimize regret in non-trivial settings.
Thus, when probabilistic methods can be considered, they absolutely should be.

As a step towards building increasingly sophisticated decision systems, it seems fruitful to expand probabilistic approaches built via Bayesian learning to more general settings.
Improved understanding of these phenomena may yield lessons of broad interest to the application of decision systems.
Contributions presented here include development of \emph{pathwise conditioning} methods for making Gaussian process models easier to work with, and a variety of \emph{non-Euclidean Gaussian processes}, both described next.

\chapter{Pathwise Conditioning}
\label{ch:pathwise}

\lettrine{G}{aussian processes} admit analytic conditional distributions, making them a key model class for Bayesian learning.
The standard way of viewing these conditional distributions mirrors the general measure-theoretic setup common to all Bayesian models, and has strongly influenced how people think about Gaussian processes.

In the early 1970s, an alternative view emerged in the geostatistics community.
Miraculously, in the Gaussian case it is also possible to develop conditioning in a manner not purely based on \emph{distributions}, but on \emph{random variables} directly.
This view turns out to lift from the multivariate to the Gaussian process setting, yielding \emph{pathwise} representations of posterior Gaussian processes, which have largely been overlooked in machine learning until now.

The pathwise perspective turns out to be a powerful point of view with wide-ranging consequences.
We will show how to use it to resolve a long-standing difficulty in Bayesian optimization: constructing a posterior approximation whose computational cost is linear both at training time and at test time, with excellent approximation properties and error control.

One of the key ingredients used within the construction will be basis function expansions of \emph{prior} Gaussian processes.
We will thus examine a number of methods for constructing such expansions for different classes of priors.
We will also reinterpret sparse approximations in a function-based manner simpler than the typical viewpoint.
We conclude by benchmarking Bayesian optimization using pathwise sampling.
We proceed to these developments.

\section{Conditioning multivariate Gaussians}

We now describe conditioning of multivariate Gaussians.
Recall that using Bayes' Rule, a prior and likelihood combine into a joint distribution, which factorizes into the marginal distribution of the data and the posterior.
The posterior is the conditional distribution of the parameters given the data, which is unique almost everywhere with respect to the marginal distribution.
We now study how to represent this distribution for the case of interest.

\subsection{Distributional conditioning}

The most obvious way to represent a Gaussian conditional distribution is to simply calculate its distribution as a closed-form analytic expression.
This is given below.

\begin{proposition}[Multivariate Gaussian conditionals]
\label{prop:mvn-cond}
Let
\[
\begin{bmatrix}
\v\theta
\\
\v{y}
\end{bmatrix} 
\~[N]\del{
\begin{bmatrix}
\v\mu_{\v\theta}
\\
\v\mu_{\v{y}}
\end{bmatrix}
,
\begin{bmatrix}
\m\Sigma_{\v\theta\v\theta} & \m\Sigma_{\v\theta\v{y}}
\\
\m\Sigma_{\v{y}\v\theta} & \m\Sigma_{\v{y}\v{y}}
\end{bmatrix} 
}
\]
be non-singular.
Then we have that
\[
(\v\theta\given\v{y})(\.,\v\gamma) \~[N]\del{\v\mu_{\v\theta} + \m\Sigma_{\v\theta\v{y}}\m\Sigma_{\v{y}\v{y}}^{-1}(\v\gamma-\v\mu_{\v{y}}), \m\Sigma_{\v\theta\v\theta} - \m\Sigma_{\v\theta\v{y}}\m\Sigma_{\v{y}\v{y}}^{-1}\m\Sigma_{\v{y}\v\theta}}
.
\]
\end{proposition}

\begin{proof}
By non-singularity, $(\v\theta,\v{y})$ admits a Lebesgue density, and the claim follows by direct calculation via applying Bayes' Rule for densities.
\end{proof}

\begin{figure}
\begin{subfigure}{0.98\textwidth}
\includegraphics{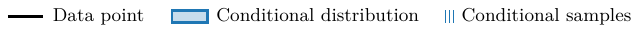}
\end{subfigure}
\begin{subfigure}{0.49\textwidth}
\input{figures/tex/mvn-dist-joint.tex}
\caption{Calculate conditional}
\end{subfigure}
\begin{subfigure}{0.49\textwidth}
\includegraphics{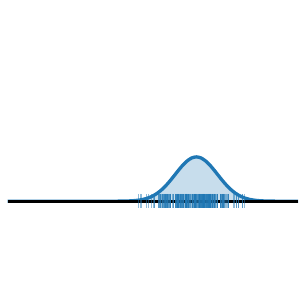}
\caption{Draw samples}
\end{subfigure}
\caption[Distributional conditioning of multivariate Gaussians]{Illustration of distributional conditioning of a bivariate Gaussian. Here, we form the joint distribution, calculate the conditional distribution, and then draw samples from it. Note that all steps except the very last one are \emph{distributional} in nature and do not involve the use of random variables, which only appear at the very end of the process.}
\label{fig:mvn-cond}
\end{figure}

See \textcite[Appendix A]{rasmussen06}, for a full derivation.
The non-singularity requirement here is more-or-less necessary: otherwise, the marginal distribution of $\v{y}$ may admit too many null sets, rendering the desired conditional distribution non-unique, except in regions where one can apply a linear map to recover a suitable Lebesgue density and apply the above argument on the obtained subspace.

If one wants to work with this expression numerically, then it's possible to simply compute the desired conditional mean and covariance. 
Then, one can generate conditional samples via the expression
\[
(\v\theta\given\v{y})(\omega,\v\gamma) &= \m{L}_{\v\theta\given\v{y}}\v{z}(\omega) + \v\mu_{\v\theta\given\v{y}}
&
\v{z} &\~[N](\v{0},\m{I})
\]
where $\v\mu_{\v\theta\given\v{y}}$ is the conditional mean, and $\m{L}_{\v\theta\given\v{y}}$ is a Cholesky factor of the conditional covariance, or any other matrix square root obtained numerically.
We illustrate this procedure in \Cref{fig:mvn-cond}.
This enables one to calculate any quantity of interest depending on the conditional distribution numerically via the Monte Carlo method.
The computational costs will in general be cubic in the dimension of both $\v\theta$ and $\v{y}$, owing to the need to compute $\m{L}_{\v\theta\given\v{y}}$ and invert $\m\Sigma_{\v{y}\v{y}}$, respectively.

\subsection{Pathwise conditioning}

The preceding considerations gave closed-form analytic expressions for Gaussian conditionals in terms of matrix-vector expressions that can be computed numerically.
From this, one might be tempted to conclude that there is nothing more to say about conditioning multivariate Gaussians---this, however, would miss an alternative view: Gaussian conditionals, which in general are a purely \emph{distributional} notion, can also be described in a \emph{pathwise} manner using \emph{random variables}.

\begin{restatable}[Matheron's update rule]{theorem}{thmmvnpw}
\label{thm:mvn-pw}
For $\v\theta,\v{y}$ defined in \Cref{prop:mvn-cond}, we have that
\[
(\v\theta\given\v{y})(\omega,\v\gamma) = \v\theta(\omega) + \m\Sigma_{\v\theta\v{y}}\m\Sigma_{\v{y}\v{y}}^{-1}(\v\gamma - \v{y}(\omega))
.    
\]
\end{restatable}

\begin{proof}
By direct calculation,
\[
\E(\v\theta + \m\Sigma_{\v\theta\v{y}}\m\Sigma_{\v{y}\v{y}}^{-1}(\v\gamma - \v{y})) = \v\mu_{\v\theta} + \m\Sigma_{\v\theta\v{y}}\m\Sigma_{\v{y}\v{y}}^{-1}(\v\gamma - \v\mu_{\v{y}}) = \E(\v\theta\given\v{y}=\v\gamma)
\]
and 
\[
\Cov(\v\theta + \m\Sigma_{\v\theta\v{y}}\m\Sigma_{\v{y}\v{y}}^{-1}(\v\gamma - \v{y})) &= \m\Sigma_{\v\theta\v\theta} + \m\Sigma_{\v\theta\v{y}}\m\Sigma_{\v{y}\v{y}}^{-1}  \m\Sigma_{\v{y}\v\theta} - 2\m\Sigma_{\v\theta\v{y}}\m\Sigma_{\v{y}\v{y}}^{-1} \m\Sigma_{\v{y}\v\theta}
\\
&= \m\Sigma_{\v\theta\v\theta} - \m\Sigma_{\v\theta\v{y}}\m\Sigma_{\v{y}\v{y}}^{-1}  \m\Sigma_{\v{y}\v\theta} = \Cov(\v\theta\given\v{y}=\v\gamma)
\]
where we have cancelled a factor of $\m\Sigma_{\v{y}\v{y}}\m\Sigma_{\v{y}\v{y}}^{-1}$ in the middle term.
\end{proof}

\begin{figure}
\begin{subfigure}{0.98\textwidth}
\includegraphics{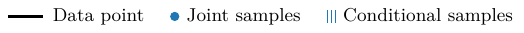}
\end{subfigure}
\begin{subfigure}{0.49\textwidth}
\includegraphics{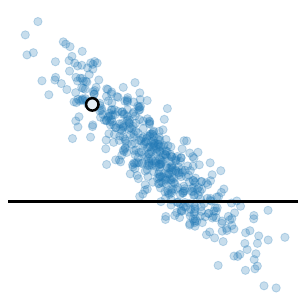}
\caption{Sample jointly}
\end{subfigure}
\begin{subfigure}{0.49\textwidth}
\includegraphics{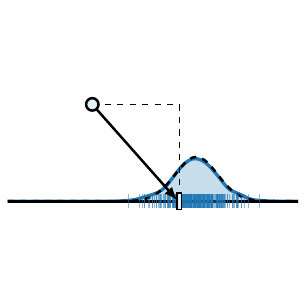}
\caption{Transform into conditional}
\end{subfigure}
\caption[Pathwise conditioning of multivariate Gaussians]{Illustration of pathwise conditioning of a bivariate Gaussian. Here, we first sample a random vector from the joint distribution, then transform it into a sample from the conditional distribution.
These steps are called \emph{pathwise} because they are defined directly using random variables, rather than indirectly through probability distributions.}
\label{fig:mvn-pw}
\end{figure}

This is illustrated in \Cref{fig:mvn-pw}.
The above argument affirms the claim, but gives few hints on where this expression originates or how to obtain it from first principles.
To better understand this, we now prove \Cref{thm:mvn-pw} in a different way.
To do so, we first prove a plug-in property of conditioning.

\begin{lemma}
\label{lem:cond-repr}
Consider three random vectors $\v{a} : \Omega \-> \R^m$, $\v{b} : \Omega \-> \R^n$, and $\v{c} : \Omega \-> \R^m$ such that 
\[
\v{a} = f(\v{b}) + \v{c}    
\]
where $f : \R^n \-> \R^m$ is a measurable function, and where the random variables $\v{b}$ and $\v{c}$ are independent. 
Then we have 
\[
(\v{a} \given \v{b} = \v\beta) = f(\v\beta) + \v{c}    
.
\]
\end{lemma}

\begin{proof}
This follows by direct calculation by writing
\[
\int_{A_{\v{b}}} \pi_{\v{a}\given\v{b}}(A_{\v{a}}\given\v\beta) \d\pi_{\v{b}}(\v\beta) &= \P(\v{a} \in A_{\v{a}}, \v{b} \in A_{\v{b}}) 
\\
&= \P(f(\v{b}) + \v{c} \in A_{\v{a}}, \v{b} \in A_{\v{b}})
\\
&= \int_{\R^m\x\R^n} \1_{f(\v\beta) + \v\varsigma \in A_{\v{b}}, \v\beta\in A_{\v{b}}} \d(\pi_{\v{c}}\ox\pi_{\v{b}})(\v\varsigma,\v\beta)
\\
&= \int_{A_{\v{b}}} \int_{\R^m} \1_{f(\v\beta) + \v\varsigma \in A_{\v{b}}} \d\pi_{\v{c}}(\v\varsigma) \d\pi_{\v{b}}(\v\beta)
\\
&= \int_{A_{\v{b}}} \P(f(\v\beta) + \v{c} \in A_a) \d\pi_{\v{b}}(\v\beta)
\]
where we have used independence to represent the probability as an integral over a product measure, followed by Tonelli's Theorem.
By the Disintegration Theorem, $\pi_{\v{a}\given\v{b}}$ is $\pi_{\v{b}}\ae[-]$ unique, implying that the conditional distributions of interest are equal, and the claim follows.
\end{proof}

The key idea behind our argument will be to choose the \emph{conditional expectation} for our function $f$, or more precisely, the map $\v{y} \|> \E(\v\theta\given\v{y})$.
We now recall this notion and some of its key properties.

Conditional expectation is defined as the orthogonal projection from the Lebesgue space $L^2(\Omega,\c{F},\P;\R^n)$ onto the subspace $L^2(\Omega,\sigma(\v{y}),\P;\R^n)$ where $\sigma(\v{y})$ is the smallest $\sigma$-algebra containing all preimages $\v{y}^{-1}(A_{\v{y}})$ where $A_{\v{y}}\in\c{B}(\R^n)$. 
Recall that the preimage is a map $\v{y}^{-1} : \c{B}(\R^n) \-> \c{F}$ between $\sigma$-algebras.
This definition is reasonably intuitive: we can think of this as projecting onto the subspace induced by all collections of random numbers in $\Omega$ which play a role in determining what $\v{y}$ does.

Recall that $L^2(\Omega,\c{F},\P;\R^n)$ is the Hilbert space of equivalence classes of random variables with inner product given by $\innerprod{\v{a}}{\v{b}} = \E(\v{a}\cdot\v{b})$.
Using this, it follows from the Projection Theorem for Hilbert spaces that the terms $\E(\v\theta\given\v{y})$ and $(\v\theta - \E(\v\theta\given\v{y}))$ are uncorrelated---and, in the Gaussian case, that they are independent.
This gives us the candidate random variables to use for $\v{b}$ and $\v{c}$, if we choose conditional expectation for $f$.

Finally, recall that for multivariate Gaussians, the conditional expectation is given by $\E(\v\theta\given\v{y}) = \v\mu_{\v\theta} + \m\Sigma_{\v\theta\v{y}}\m\Sigma_{\v{y}\v{y}}^{-1}(\v{y} - \v\mu_{\v{y}})$, where we note in our setting that the inverse always exists by non-singularity of $\v{y}$.
With these preparations, we are ready to revisit \Cref{thm:mvn-pw}.

\thmmvnpw*

\begin{proof}
Write 
\[
\v\theta = \E(\v\theta\given\v{y}) + (\v\theta - \E(\v\theta\given\v{y}))
\]
and note that since $\E(\v\theta\given\v{y})$ and $(\v\theta - \E(\v\theta\given\v{y}))$ are uncorrelated and jointly Gaussian, they are independent.
Applying \Cref{lem:cond-repr} yields
\[
(\v\theta\given\v{y})(\omega,\v\gamma) &= \m\Sigma_{\v\theta\v{y}}\m\Sigma^{-1}_{\v{y}\v{y}} \v\gamma + (\v\theta(\omega) - \m\Sigma_{\v\theta\v{y}}\m\Sigma^{-1}_{\v{y}\v{y}} \v{y}(\omega))
\\
&= \v\theta(\omega) + \m\Sigma_{\v\theta\v{y}}\m\Sigma^{-1}_{\v{y}\v{y}}(\v\gamma - \v{y}(\omega))
\]
where we have substituted $\E(\v\theta\given\v{y}) = \v\mu_{\v\theta} + \m\Sigma_{\v\theta\v{y}}\m\Sigma_{\v{y}\v{y}}^{-1}(\v{y} - \v\mu_{\v{y}})$ and immediately cancelled the mean terms. 
The claim follows.
\end{proof}

From \Cref{thm:mvn-pw}, we obtain a second way of representing multivariate Gaussian conditionals.
This entails two steps: (i) sample $\v\theta,\v{y}$ jointly, and (ii) transform $\v\theta,\v{y}$ into $\v\theta\given\v{y}=\v\gamma$ by employing the given expression.
For an illustration of this procedure, see \Cref{fig:mvn-pw}.

Remarkably, this result is seemingly missing from every machine learning textbook on Gaussian processes in widespread use, and appears almost entirely unknown within the field.
It's possible this is because the expression's computational costs are cubic in the combined dimension, which is more expensive than the previous costs.
While this holds for general Gaussians, we show it can be avoided for many cases of practical interest.

On the other hand, \Cref{thm:mvn-pw} is certainly known in other communities.
In a tribute to Georges Matheron, who pioneered the expression's use in geostatistics, \textcite{chiles05} say that:

\begin{quotation}
[Matheron's update rule] is nowhere to be found in Matheron's entire published works, as he merely regarded it as an immediate consequence of the orthogonality of the [conditional expectation] and the [residual process].
\end{quotation}

More recently, \textcite{doucet10} describes the algorithm in a technical report which begins with the remark: 

\begin{quotation}
This note contains no original material and will never be submitted anywhere for publication. However, it might be of interest to people working with [Gaussian processes] so I am making it publicly available.
\end{quotation}

Additionally, \Cref{thm:mvn-pw} is reasonably well-known in geostatistics \cite{journel78,defouquet94,emery07,oliver96}.
In parallel, these ideas were rediscovered in the astrophysics community, with \textcite{hoffman91} describing approximations similar in spirit to the ones we study in the sequel.

The present state of affairs therefore seems to be that a small set of technical experts are aware of \Cref{thm:mvn-pw} but believe it to be too trivial to write about, while practitioners working in areas such as Bayesian optimization do not know that it exists.
While for multivariate Gaussians the result certainly is trivial, we subsequently show that using it in the right manner yields significant progress towards resolving certain issues in decision-making settings.
To do so, we now consider Gaussian processes.

\section{Conditioning Gaussian processes}

We now study conditioning in Gaussian processes.
For this, we use \textcite{rasmussen06} as our canonical reference.
In what follows, we develop and showcase the two points of view---distributional and pathwise---introduced in the Gaussian process setting.
We begin with the former.

\subsection{Distributional conditioning}

The standard way of representing Gaussian process conditionals is to use the finiteness of the data to pick a suitable set of locations and work with finite-dimensional marginals.
Since conditioning and marginalization commute, conditioning the Gaussian process thus reduces to conditioning multivariate Gaussians.
We now describe this.

To ease notation, we now set the prior mean to zero: the general case can be recovered by adding and subtracting mean functions.
Additionally, in what follows, we let $\v{x} = (x_1,..,x_n)$ denote the data locations.
For this usage specifically, we use bold italics to indicate a product structure rather than a linear structure, and in particular do not require $X$ to be a vector space.
Finally, functions applied to product spaces are understood componentwise.
With this notation in place, we are ready to state the claim.

\label{ntn:gp-model}
\begin{proposition}[Posterior Gaussian process]
\label{prop:gp-cond}
The Bayesian model
\[
\v{y} \given f &\~[N](f(\v{x}), \m\Sigma)
&
f &\~[GP](0,k)
\]
admits the Gaussian process
\[
(f\given\v{y})(\.,\v\gamma)\~[N](\m{K}_{(\.)\v{x}} (\m{K}_{\v{x}\v{x}} + \m\Sigma)^{-1}\v\gamma, \m{K}_{(\.,\.)} - \m{K}_{(\.)\v{x}} (\m{K}_{\v{x}\v{x}} + \m\Sigma)^{-1} \m{K}_{\v{x}(\.)})
\]
as its posterior. 
\end{proposition}

\begin{proof}
Apply \Cref{prop:mvn-cond} to a set of finite-dimensional marginals.
\end{proof}

\begin{figure}
\begin{subfigure}{0.98\textwidth}
\includegraphics{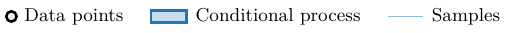}
\end{subfigure}
\begin{subfigure}{0.49\textwidth}
\includegraphics{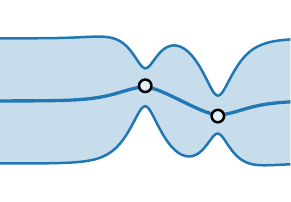}
\caption{Calculate conditional}
\end{subfigure}
\begin{subfigure}{0.49\textwidth}
\includegraphics{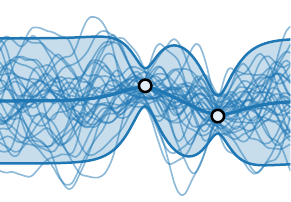}
\caption{Draw samples}
\end{subfigure}
\caption[Distributional conditioning of Gaussian processes]{Illustration of distributional conditioning of a Gaussian process. Here, we calculate the conditional distribution at a finite set of locations, and then draw samples from a multivariate Gaussian, obtaining the value of the posterior Gaussian process at a set of points. This view of conditioning is called \emph{distributional}, since random variables appear only at the very end of the algorithmic process.}
\label{fig:gp-cond}
\end{figure}

This result, illustrated in \Cref{fig:gp-cond}, gives us a way of carrying out Bayesian learning using Gaussian processes, given a finite set of data.
Note the \emph{bottom-up} nature of this perspective: we describe the posterior Gaussian process---which is an actual \emph{process} defined everywhere---using its posterior finite-dimensional marginals.

We now consider computational costs of the above formula.
Calculating posterior moments clearly entails cubic costs with respect to the data size $n$, owing the need to invert $n\x n$ matrices.
Now, suppose we are are interested in computing quantities involving the posterior distribution.
Consider for instance computing the Thompson sampling acquisition function considered in \Cref{sec:bayesian-optimization}, which we recall is 
\[
x_{t+1}(\omega) &= \argmin_{x\in X} \alpha_t(\omega,x)
&
\alpha_t&\~ f \given \v{y}
.
\]
Given the minimization involved in this objective, there is no chance in finding an analytic expression for $x_{t+1}$, and we must resort to numerical methods.
Just about any numerical procedure one can imagine---for instance, gradient descent---will involve drawing random samples from $f\given\v{y}$ at different locations, and performing the necessary algorithmic operations on them.
Summarizing, the computational costs of this expression are as follows.

\label{ntn:gp-complexity}
\1 Data: $\c{O}(n^3)$ where $n$ is the size of the training set, due to the intermediate term $(\m{K}_{\v{x}\v{x}} + \m\Sigma)^{-1}$ in the posterior covariance.
\2 Predictions: $\c{O}(n_*^3)$ where $n_*$ is the number of locations the posterior Gaussian process needs to be jointly evaluated at, due to the need to factorize the posterior covariance $\m{K}_{(\.,\.)} - \m{K}_{(\.)\v{x}} (\m{K}_{\v{x}\v{x}} + \m\Sigma)^{-1} \m{K}_{\v{x}(\.)}$.
\0

This becomes more difficult if one considers evaluation locations that are not known in advance, and might be determined using previous points.
Due to the need to iteratively re-condition on sampled process values and factorize matrices at every intermediate computation step, roundoff errors accumulate as computations proceed.
Thus, even if we are willing to pay cubic costs, we then face the secondary issue of numerical instability.
The situation if one needs to differentiate through objectives such as $\alpha_t$ is even worse.

Without additional considerations, these costs are disastrous, and illustrate typical difficulties in building practical decision-making systems powered by Gaussian processes.
Fortunately, a wide variety of techniques to deal with them are available: in particular, \emph{inducing point} methods provide a broad set of approximations for reducing the $\c{O}(n^3)$ costs.
We will complement these ideas by introducing techniques to tackle the $\c{O}(n_*^3)$ costs.
For this, we proceed to develop a pathwise view of conditioning.

\subsection{Pathwise conditioning}

Given the pathwise view of conditioning multivariate Gaussians given by Matheron's update rule, one can ask: is there an analogous statement for Gaussian processes?
Does the purely distribution notion of a Gaussian conditional have an analogous description in terms of random functions?
We answer this affirmatively below.

\begin{corollary}[Posterior Gaussian process (pathwise)]
\label{cor:gp-pw}
For $\v{y}\given f$, $f$, and $\v{f}$ defined in \Cref{prop:gp-cond}, we have
\[
(f \given\v{y})(\omega,\v\gamma) = f(\omega,\.) + \m{K}_{(\.)\v{x}} (\m{K}_{\v{x}\v{x}} + \m\Sigma)^{-1}(\v\gamma - f(\omega,\v{x}) - \v\eps(\omega))
\]
where $\v\eps \~[N](\v{0},\m\Sigma)$.
\end{corollary}

\begin{proof}
Apply \Cref{thm:mvn-pw} to a set of finite-dimensional marginals.
\end{proof}

\begin{figure}
\begin{subfigure}{0.98\textwidth}
\includegraphics{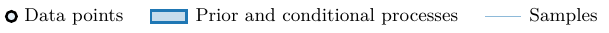}
\end{subfigure}
\begin{subfigure}{0.49\textwidth}
\includegraphics{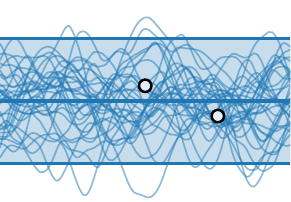}
\caption{Sample from prior}
\end{subfigure}
\begin{subfigure}{0.49\textwidth}
\includegraphics{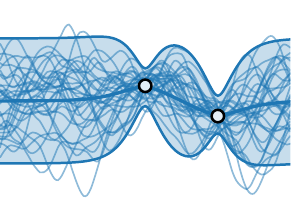}
\caption{Transform into conditional}
\end{subfigure}
\caption[Pathwise conditioning of Gaussian processes]{Illustration of pathwise conditioning of a Gaussian process. Here, we first sample a set of random functions from the prior, along with random noise variables at each of the data locations, then transform these into a samples from the posterior distribution.
This view of conditioning is termed \emph{pathwise}, since it is defined directly at the level of random functions.}
\label{fig:gp-pw}
\end{figure}

Note that in this expression, the prior is evaluated jointly at all locations---the random variables $f(\omega,\.)$ and $f(\omega,\v{x})$ are \emph{dependent}.
Similarly, equality holds in distribution, because the random variable $(f\given\v{y})(\.,\v\gamma)$ is only defined in distribution to begin with.

\Cref{cor:gp-pw} gives an alternative way of representing posterior Gaussian processes: (i) sample the prior and all auxillary random variables such as the noise term $\v\eps$, and (ii) transform the sampled function to form the posterior as a random function.
This is shown in \Cref{fig:gp-pw}.

This strategy can be carried out if we know the locations we wish to evaluate the posterior at.
In this case, we sample the prior at the data and evaluation locations jointly, and transform the resulting samples into posterior samples.
Examining the computational costs, we see these are $\c{O}(n^3)$ with respect to data size, and $\c{O}(n_*^3)$ with respect to the number of evaluation locations.
At this stage, then, we have seemingly only gained numerical stability.

\label{ntn:gp-approx-prior}
\Cref{cor:gp-pw}, however, is not merely a computational result: it gives us a powerful way of thinking about posterior Gaussian processes. 
In particular, we can use the point of view it offers to construct posterior approximations.
Observe that the cubic costs $\c{O}(n_*^3)$ occur entirely due to the need to jointly sample the prior at all evaluation locations.
Suppose that we can approximately express the prior using a set of \emph{finite basis functions} as 
\[
f(\omega,\.) &\approx \tilde{f}(\omega,\.) = \ubr{\sum_{i=1}^\ell w_i(\omega)\phi_i(\.)}_{\t{finite basis functions}}
&
w_i &\~[N](0,1)
.
\]
We can substitute this approximation into \Cref{cor:gp-pw} to obtain
\[
(f \given\v{y})(\omega,\v\gamma) \approx \tilde{f}(\omega,\.) + \m{K}_{(\.)\v{x}} (\m{K}_{\v{x}\v{x}} + \m\Sigma)^{-1}(\v\gamma - \tilde{f}(\omega,\v{x}) - \v\eps(\omega))
.
\]
Once the random weights $\v{w}$ are sampled, the posterior becomes a deterministic function, which can be evaluated at $\c{O}(n_*)$ costs.
Thus, under this class of approximations, our cubic costs with respect to the number of evaluation locations become \emph{linear}.
We show in the sequel that for appropriate choices of $\tilde{f}$, such approximations can achieve excellent error control, ensuring they perform effectively in practice.

\begin{figure*}
\includegraphics{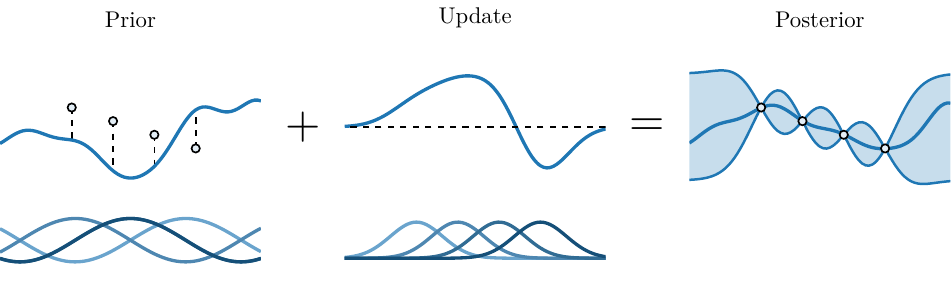}
\caption[Approximate pathwise conditioning]{Approximate pathwise conditioning, with bases on bottom row.}
\label{fig:gp-pw-approx}
\end{figure*}

\label{ntn:canonical-basis-fns}
We now examine a different aspect of the pathwise representation of posterior Gaussian processes: the role of the data.
By re-expressing the matrix-vector product of the kernel matrix term $\m{K}_{(\.)\v{x}}$ as a sum, we obtain
\[
(f \given\v{y})(\omega,\v\gamma) = f(\omega,\.) + \ubr{\sum_{j=1}^n v_j(\omega) k(x_j,\.)}_{\t{canonical basis functions}}
\]
where $\v{v} = (\m{K}_{\v{x}\v{x}} + \m\Sigma)^{-1}(\v\gamma - f(\omega,\v{x}) - \v\eps(\omega))$.
This identifies the effect of the data in a posterior Gaussian process as the addition of \emph{canonical basis functions} $k(x_i,\.)$ to the prior.
We explore this interpretation further in what follows.
From this viewpoint, if we define $\v{\tilde{v}}$ analogously to $\v{v}$, but with $f$ replaced with $\tilde{f}$, our approximate posterior is
\[
(f \given\v{y})(\omega,\v\gamma) \approx \sum_{i=1}^\ell w_i(\omega)\phi_i(\.) + \sum_{j=1}^n \tilde{v}_j(\omega) k(x_j,\.)
\]
which shows that our proposed approximation involves writing the posterior Gaussian process of interest as a sum of \emph{two} finite sets of basis functions---one for the prior, another for the data.
The basis coefficients $w_i$ and $\tilde{v}_j$ here are dependent random variables.
We illustrate the overall approximation in \Cref{fig:gp-pw-approx}.
Summarizing, pathwise representations of posterior Gaussian processes provide a useful framework for constructing posterior approximations which are actual random functions.

In Bayesian optimization, such approximations offer a promising avenue to avoid the computational difficulties that result from working with finite-dimensional marginals.
Computing the Thompson sampling acquisition function
\[
x_{t+1}(\omega) &= \argmax_{x\in X} \alpha_t(\omega,x)
&
\alpha_t&\~ f \given \v{y}
\]
or any other quantities that require evaluating the Gaussian process at arbitrary locations can be done using a simple Monte Carlo approach as follows.

\1 Sample the random weights $(\v{w}, \v{v})$ to form the approximate posterior.
\2 Maximize the approximate posterior using any numerical procedure.
\0

The key advantage of this approach is that once the random weights are sampled, the posterior---a random function---effectively becomes deterministic.
This allows us to not only evaluate it in linear time, but also to differentiate through posterior samples using automatic differentiation---here, enabling us to maximize them using gradient descent without computing gradient processes or employing special considerations of any kind.

In total, we obtain a simple, accurate, and efficient way to compute acquisition functions such as Thompson sampling.
To obtain a complete algorithm, all that remains is to find finite basis approximations to Gaussian process priors: this will require additional structure present in specific classes of kernels.
We proceed to explore a number of techniques for doing so.

\section{Sampling from prior Gaussian processes}

In the preceding section, we explored a class of approximate posterior Gaussian processes constructed by plugging a finite-basis-function-based approximate prior into the pathwise update.
Such approximate priors can be constructed in many ways, including by expressing the true prior within a basis of an appropriate space of functions, and truncating the resulting infinite sum.
Different choices of bases will result in different applicability, ease of use, and approximation error.
We now explore possible choices.

\subsection{Random feature methods}

One of the most popular classes of kernels is the class of stationary kernels defined on Euclidean spaces, which are kernels that factorize through a one-argument function depending only on the difference between points.
For such kernels, random feature methods, originally proposed by \textcite{rahimi08}, can be used to construct approximate priors whose properties make them a particularly attractive choice.
We examine these now.

To construct a basis function expansion, our strategy will be to find a \emph{feature map} $\varphi : X \-> H_k$ that maps states into vectors in the \emph{reproducing kernel Hilbert space} induced by $k$.
This is defined as follows.

\label{ntn:rkhs}
\begin{definition}[Reproducing kernel Hilbert space]
Let $X$ be a set, and let $H \subseteq \R^X$ be a Hilbert space of functions. 
We say that $H$ is a \emph{reproducing kernel Hilbert space} if, for any $x\in X$, we have $\f{ev}_x \in H^*$ where $\f{ev}_x : H \-> \R$ is called the \emph{evaluation map} and is defined by $\f{ev}_x f = f(x)$.
\end{definition}

Note that in this definition, the statement $\f{ev}_x \in H^*$ means that $\f{ev}_x$ is a bounded linear functional.
Reproducing kernel Hilbert spaces therefore impose continuity requirements on their pointwise evaluation functionals.

Ostensibly, this definition has nothing to do with kernels, and it is unclear what a reproducing kernel Hilbert space \emph{induced by} $k$ actually means.
A consequence of the above definition is that given a reproducing kernel Hilbert space $H$, we can define the function $k_{H}(x,x') = \innerprod[1]{\Psi_{H}^{-1}\f{ev}_x}{\Psi^{-1}_{H}\f{ev}_{x'}}$, called the \emphmarginnote{reproducing kernel}, where $\Psi_{H} : H \-> H^*$ is the bijective linear isometry given by the Riesz Representation Theorem.
It is easy to see that $k_{H}$ is positive semi-definite.
It turns out a converse statement also holds.

\begin{result}[Moore--Aronszajn Theorem]
Let $k : X \x X \-> \R$ be a symmetric positive semi-definite kernel.
Then there is a unique Hilbert space $H_k \subseteq \R^X$ of real-valued functions for which $k$ is the reproducing kernel.
\end{result}

\begin{proof}
\textcite[Proposition 2.13 and Theorem 2.14]{paulsen16}.
\end{proof}

This gives another point of view from which one can study and understand kernels and, by proxy, Gaussian processes.
The reproducing kernel Hilbert space induced by the covariance kernel of a Gaussian process is sometimes called its \emph{Cameron--Martin space} or \emph{native space}, and encodes its mathematical properties.
This perspective can be studied abstractly, leading to the theory of \emph{isonormal Gaussian processes}.
See \textcite{wendland04, lifshits12, legall16} for details on these and related ideas.

\begin{figure}
\begin{subfigure}{0.49\textwidth}
\includegraphics{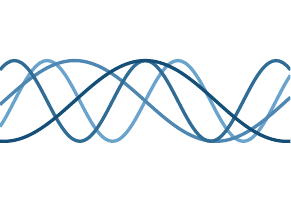}
\caption{Fourier basis functions}
\end{subfigure}
\begin{subfigure}{0.49\textwidth}
\includegraphics{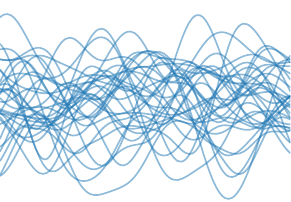}
\caption{Approximate prior samples}
\end{subfigure}
\caption[Random Fourier feature prior approximations]{Illustration of random Fourier feature methods for sampling from approximate priors. Here, we show a small subset of randomly sampled Fourier basis functions, along with approximate prior samples constructed using the random Fourier basis functions.}
\label{fig:gp-rff}
\end{figure}

\label{ntn:approx-feature-map}
We will need one final notion.
We say that a function $\varphi : X \-> H_k$ is a \emphmarginnote{feature map} if $k(x,x') = \innerprod{\phi(x)}{\phi(x')}$.
Now, we introduce the key idea for constructing our approximate Gaussian prior: suppose we have a \emph{finite-dimensional} approximation for such a feature map, namely a vector-valued function $\v\phi : X \-> \R^\ell$ such that $\v\phi(x)^T \v\phi(x') \approx \innerprod{\varphi(x)}{\varphi(x')}$.
Then
\[
\tilde{f}(\omega,\.) &= \sum_{i=1}^\ell w_i(\omega) \phi_i(\.)
&
w_i &\~[N](0,1)
\]
by direct calculation has covariance approximately equal to that of $f$.
An example can be seen in \Cref{fig:gp-rff}.
Therefore, to construct an approximate prior, it suffices to find a finite-dimensional approximate feature map.

For stationary kernels, techniques for constructing approximate feature maps are well-studied, originally motivated by questions arising in kernel support vector machines.
We now introduce the \emph{random Fourier feature} method for constructing such maps, beginning with a brief description of the stationary setting.

\parmarginnote{Stationary kernel}
Let $X = \R^d$.
In the remaining subsection, we use bold italic $\v{x}$ to denote linear rather than product structure.
We say that a kernel $k(\v{x},\v{x}')$ is called \emph{stationary} if $k(\v{x},\v{x}') = k(\v{x} - \v{x})$ for a function $k : \R^d \-> \R$.
A stationary kernel, then, is a two-argument positive definite function which factorizes through a one-argument function depending only on the difference between two points.
Note that such a kernel is invariant under translation and can be characterized as such.
We will need a result known as \emph{Bochner's Theorem}.

\begin{result}[Bochner's Theorem]
For every stationary continuous positive definite kernel with $k(\v{x},\v{x}) = 1$ there is a symmetric probability measure $\rho$ on $\R^d$ which we call the \emph{spectral measure} of $k$.
Moreover, $k$ admits the representation
\[
k(\v{x} - \v{x}') = \int_{\R^d} e^{2\pi i \v\varpi^T (\v{x} - \v{x}')} \d\rho(\v\varpi)
.
\]
Conversely, every probability measure on $\R^d$ gives rise to such a kernel via its Fourier transform.
\end{result}

\begin{proof}
\textcite[Theorem 10.4]{paulsen16}.
\end{proof}

Here, \emph{symmetry} of $\rho$ refers to invariance under reflection about the origin.
This result is true more generally if $X$ is replaced by a locally compact Abelian group, for which the corresponding spectral measure will be supported on the Pontryagin dual group---see \textcite{paulsen16}. 
We omit this level of generality because we will not need it.
We do briefly note, however, that this means that spectral measures of kernels on compact spaces are \emph{discrete}---this behavior will reappear under a different guise in \Cref{ch:noneuclidean}.

From here, it is clear that the feature map we seek can be constructed by Monte Carlo approximation of the integral representation given by Bochner's Theorem.
To maintain order, we introduce a second probability space $(\Xi,\c{G},\bb{Q})$ to distinguish the stochasticity associated with the random feature expansion from stochasticity associated with the Gaussian process.
Letting $k(\v{x},\v{x}) = \sigma^2$, write
\[
k(\v{x} - \v{x}') &= \sigma^2 \int_{\R^d} e^{2\pi i \v\varpi^T (\v{x} - \v{x}')} \d\rho(\v\varpi)
\\
&= \sigma^2 \int_{\R^d} e^{2\pi i \v\varpi^T \v{x}} \conj{e^{2\pi i \v\varpi^T \v{x}'}} \d\rho(\v\varpi)
\\
&\approx \frac{\sigma^2}{\ell} \sum_{j=1}^\ell e^{2\pi i \v\varpi_j(\xi)^T \v{x}} \conj{e^{2\pi i \v\varpi_j(\xi)^T \v{x}'}}
\\
&= \frac{\sigma^2}{\ell} \sum_{i=1}^\ell \begin{aligned}
\cos&(2\pi \v\varpi_i(\xi)^T \v{x})\cos(2\pi \v\varpi_i(\xi)^T \v{x}') 
\\
+&\sin(2\pi \v\varpi_i(\xi)^T \v{x})\sin(2\pi \v\varpi_i(\xi)^T \v{x}')
\end{aligned}
\\
&= \v\phi(\xi,\v{x})^T\v\phi(\xi,\v{x}')
\]
where $\conj{\,\.\,}$ denotes complex conjugation, and
\[
\phi_i(\xi,\v{x}) = \lbr{\begin{aligned}
&\frac{\sigma}{\sqrt{\ell}} \cos(2\pi \v\varpi_i(\xi)^T \v{x}), 
&
&i \t{odd}
\\
&\frac{\sigma}{\sqrt{\ell}} \sin(2\pi \v\varpi_{i-1}(\xi)^T \v{x}), 
&
& i \t{even}
\end{aligned}}
\]
which, for $\ell$ even, gives our approximate prior, shown in \Cref{fig:gp-rff} as
\[
\tilde{f}(\.) &= \sum_{i=1}^\ell w_i(\omega) \phi_i(\xi,\.)
&
w_i &\~[N](0,1)
&
\v\varpi_i &\~ \rho
.
\]
One remarkable property of random feature methods is that their approximation error decays at a \emph{dimension-free} rate in the Monte Carlo sense: for details, see \textcite{sutherland15}.
This limits the effect of the curse of dimensionality to constant factors.
Better yet, random feature methods are well-understood owing to their widespread use in other areas such as support vector machines \cite{rahimi08,liu21}.
For this reason, random feature methods are often the technique of choice for stationary Euclidean kernels.

A number of different random Fourier feature approximations have been proposed in the literature \cite{rahimi08,yu16,choromanski16,liu21}, along with techniques for their theoretical analysis \cite{sutherland15,sriperumbudur15,choromanski18,li19,liu21}.
For a comprehensive review of random feature methods, see \textcite{liu21}.
We now consider other classes of approximations.

\subsection{Karhunen--Loève expansions}

Among all techniques for constructing approximate priors, different techniques will generally yield different amounts of error for the same number of basis functions.
One can then ask: is there an optimal choice?
Of course, the answer to this question will depend on what one actually means by the word \emph{optimal}.

If we take $X \subset \R^d$ to be compact, and we use expected mean squared error---which we recall is equivalent to the $L^2(\Omega;L^2(X;\R))$ norm---as our notion of optimality, one can affirmatively answer the above question.
Recall that a continuous kernel $k : X \x X \-> \R$ induces a covariance operator
\[
\c{K} : L^2(X;\R) &\-> L^2(X;\R)
&
\c{K} : \phi &\|> \int_X \phi(x) k(x,\.) \d x
\]
where by writing $\norm{\c{K} \phi}_{L^2(X;\R)} \leq \vol(X) \norm{k}_{C^0(X\x X;\R)} \norm{\phi}_{L^2(X;\R)}$ using compactness and boundedness of $k$ we affirm correctness of the operator's domain and range.
By compactness, $\c{K}$ will admit a countable set of eigenvalues and eigenfunctions.
It turns out, that, by the Karhunen--Loève Theorem, the Gaussian process itself can be written in terms of the same eigenvalues and eigenfunctions as well.

\begin{result}[Karhunen--Loève Theorem]
Let $X \subseteq \R^d$ be compact, and let $f$ be a Gaussian process with continuous covariance function.
Then we have
\[
f(\omega,\.) &= \sum_{i=1}^\infty w_i(\omega) \phi_i(\.)
&
w_i &\~[N](0,1)
\]
where convergence holds almost surely, and $\phi_i$ are an orthogonal basis on $L^2(X;\R)$ given by rescaled eigenfunctions of the covariance operator.
Moreover, for every $\ell\in \N$, truncating this series yields an $L^2(\Omega;L^2(X;\R))$-optimal approximation among all $\ell$-term sums of $L^2(X;\R)$-orthogonal functions.
\end{result}

\begin{proof}
\textcite[Theorem 5.28]{lord14}, as well as \textcite[Section 2.3.2]{ghanem91}.
\end{proof}

More generally, an analogous result, albeit with a weaker form of convergence, also holds for general \emph{square-integrable} stochastic processes which may be non-Gaussian.
Since we will not consider such processes, we do not pursue this direction here.

\begin{figure}
\begin{subfigure}{0.49\textwidth}
\includegraphics{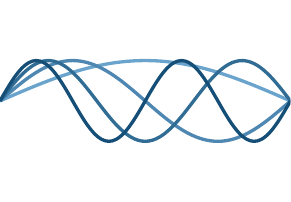}
\caption{Karhunen--Loève basis}
\end{subfigure}
\begin{subfigure}{0.49\textwidth}
\includegraphics{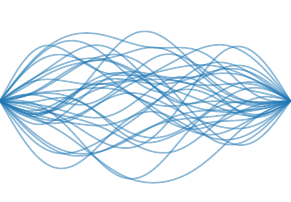}
\caption{Approximate prior samples}
\end{subfigure}
\caption[Karhunen--Loève prior approximations]{Illustration of a Karhunen--Loève expansion for a boundary-constrained squared exponential kernel on the unit interval.
For this kernel, the basis functions coincide with the eigenfunctions of the Dirichlet Laplacian.
We show the first four eigenfunctions, along with approximate prior samples using the first 87 terms in the expansion. This truncation was chosen because the remaining terms are zero in floating-point arithmetic, highlighting the approximation's efficiency.}
\label{fig:gp-kl}
\end{figure}

For a general kernel, finding the eigenfunctions of the covariance operator may be difficult.
In many cases, however, they can be understood by utilizing additional structure present in the problem.
For example \textcite{solin19} study certain classes of Gaussian processes constructed to satisfy boundary conditions.
Here, the eigenfunctions of the covariance operator coincide with the eigenfunctions of a boundary-constrained Laplacian, allowing them to be obtained numerically.
These are shown in \Cref{fig:gp-kl}.

Kernels induced by eigenvalues and eigenfunctions of appropriately-defined Laplace operators are also a common tool in non-Euclidean settings \cite{solin18,solin20,coveney20}, and we explore them further in \Cref{ch:noneuclidean}.
For the moment, however, we consider a third class of approximations.

\subsection{Finite element methods}

We now introduce the third class of approximations we will consider: those constructed via \emph{finite element approximations} of solutions of stochastic partial differential equations \cite{lord14,lototsky17,krainski18}.
Many Gaussian process priors, including the widely-used Matérn class, following \textcite{whittle54,whittle63,lindgren11}, can be expressed in such a manner.
Such Gaussian processes are also of direct interest in the non-Euclidean settings of \Cref{ch:noneuclidean}.

Let $X$ be a subset of a Euclidean space, or a Riemannian manifold.
Suppose our Gaussian process $f$ satisfies the equation 
\[
\c{L}f = \c{W} 
\]
where $\c{L} : H \-> L^2(X;\R)$ is a bounded linear operator acting on a certain reproducing kernel Hilbert space which uniquely determines the Gaussian process, and $\c{W}$ is a white noise process over $L^2(X;\R)$---a random distribution whose properties are determined by the Lebesgue space.
Slightly more precisely, the equation is meant as either an almost sure or a distributional equality
\[
f(\omega,\c{L}^* h) = \c{W}(\omega, h)
\]
between generalized Gaussian fields---a formal treatment is given in \Cref{ch:noneuclidean}.
Assume that the generalized Gaussian field $f : \Omega \x H \-> \R$ can be written as the integral of a Gaussian process $f : \Omega \x X \-> \R$ as
\[
f(\omega, h) = \int_X f(\omega, x) h(x) \d x
.
\]
This amounts to requiring the stochastic partial differential equation's solution $f$ to possess some regularity, and in particular on compact domains $X$ follows from sample-continuity of $f$.
If we define the bilinear form 
\[
a(\phi,\psi) = \int_X \phi(x) (\c{L}^*\psi)(x) \d x    
\]
then our stochastic partial differential equation becomes
\[
a(f(\omega,\.), h) = \c{W}(\omega, h)
\]
which is an equation between a Gaussian process, bilinear form, and random linear functional.
Now, we introduce approximations: suppose that $f(\omega,x) \approx \tilde{f}(\omega,x) = \sum_{i=1}^\ell w_i(\omega) \phi_i(x)$ and that $h(x) \approx \sum_{j=1}^\ell v_j \psi_j(x)$.
Plugging this in and differentiating to remove the coefficients $v_j$ yields the system of equations
\[
\sum_{i=1}^\ell w_i(\omega) a(\phi_i,\psi_j) = \c{W}(\omega, \psi_j)
\]
which by defining $A_{ij} = a(\phi_i,\psi_j)$ and $b_j(\omega) = \c{W}(\omega, \psi_j)$ can be recognized as a random linear system, more compactly written
\[
\m{A} \v{w}(\omega) = \v{b}(\omega)
.
\]
If we let $\m{M} = \Cov(\v{b})$ with $\Cov(b_i,b_j) = \innerprod{\psi_i}{\psi_j}$---where we note that since $\c{W}$ is a white noise process, the matrix $\m{M}$ coincides with the finite-element mass matrix---we obtain the distribution for the basis coefficients. 
Our approximate prior can therefore be written
\begin{figure}
\begin{subfigure}{0.49\textwidth}
\includegraphics{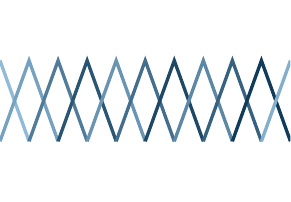}
\caption{Finite element basis}
\end{subfigure}
\begin{subfigure}{0.49\textwidth}
\includegraphics{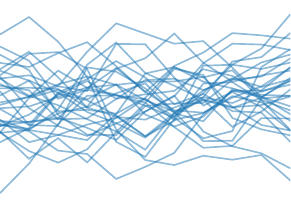}
\caption{Approximate prior samples}
\end{subfigure}
\caption[Finite element prior approximations]{Illustration of finite element approximate prior samples for a Matérn kernel with smoothness $3/2$ and length scale $\kappa$.
In one dimension, the bilinear form for this kernel is $a(f,h) = \int_X \frac{3}{\kappa^2} f(x)h(x) + \grad f(x) \. \grad h(x) \d x$. 
Since this bilinear form is of first order in both arguments, we employ a piecewise linear finite element basis consisting of compactly supported triangle-shaped functions, using it to represent both $f$ and $h$.
This results in matrices $\m{A}$ and $\m{M}$ which are symmetric tridiagonal.}
\label{fig:gp-fe}
\end{figure}
\[
\tilde{f}(\omega,x) &= \sum_{i=1}^\ell w_i(\omega) \phi_i(x)
&
\v{w} &\~[N](\v{0},\m{A}^{-1}\m{M}\m{A}^{-T})\
.
\]
What is particularly powerful about this technique is that it gives a significant degree of freedom for what kinds of finite sets of basis functions we can choose for $\phi_i$ and $\psi_j$.
If $\c{L}$ is a differential operator of sufficiently low order, a fruitful choice, shown in \Cref{fig:gp-fe} for the model studied by \textcite{lindgren11}, is to take $\phi_i$ to be compactly supported piecewise linear functions, since this will cause the matrices $\m{A}$ and $\m{M}$ to be sparse.
This can enable one to use a much larger number of basis functions compared to alternative methods.

The main issue with finite element methods is that they typically demand more from practitioners compared to alternatives.
In particular, one usually needs to understand the stochastic partial differential equations governing the Gaussian process of interest with a reasonable degree of detail to know how to choose basis functions well.
Additionally, sparse linear algebra in a parallel environment with automatic differentiation can be cumbersome.

This concludes our presentation of techniques for constructing approximate priors.
In general, the best choice for a particular application will depend on the detailed requirements of the situation.
We now proceed to study other aspects of the pathwise viewpoint.

\section{Approximating pathwise data-dependent terms}

In the preceding section, we discussed techniques for approximating the prior using a finite set of basis functions with random coefficients.
This enabled us to construct approximate pathwise representations of posterior Gaussian processes with $\c{O}(n_*)$ computational complexity, where we recall again that $n_*$ is the number of points where we wish to evaluate the posterior.
We now consider the other computational costs in the formula: $\c{O}(n^3)$, where $n$ is the size of the training data.
These costs arise because computing
\[
(f \given\v{y})(\omega,\v\gamma) = f(\omega,\.) + \m{K}_{(\.)\v{x}} (\m{K}_{\v{x}\v{x}} + \m\Sigma)^{-1}(\v\gamma - f(\omega,\v{x}) - \v\eps(\omega))
\]
requires us to invert the $n\x n$ matrix $\m{K}_{\v{x}\v{x}} + \m\Sigma$.
We now ask: from a pathwise perspective, what techniques are available for reducing these costs?

\subsection{Inducing points}

\label{ntn:gp-num-inducing}
Consider a simple approach to reducing the above cubic costs: instead of conditioning the Gaussian process on the full data $(x_1,y_1),..,(x_n,y_n)$, find a different data set $(z_1,\mu_1),..,(z_m,\mu_m)$, for which $m \ll  n$ and yet the posterior is approximately the same.
This idea underlies \emph{inducing point} approximations \cite{snelson06,titsias09,opper09,hensman13}.

\begin{figure}
\begin{subfigure}{0.98\textwidth}
\includegraphics{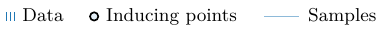}
\end{subfigure}
\begin{subfigure}{0.49\textwidth}
\includegraphics{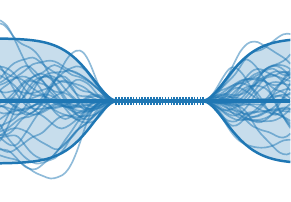}
\caption{Posterior Gaussian process}
\end{subfigure}
\begin{subfigure}{0.49\textwidth}
\includegraphics{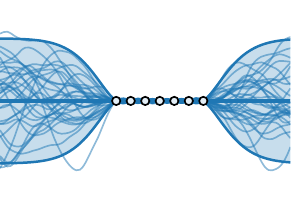}
\caption{Variational approximation}
\end{subfigure}
\caption[Variational approximations of Gaussian processes]{Inducing point approximation using a variation family built using randomized process values. Here, we use seven inducing points to represent the posterior distribution under thirty-one data points. 
The inducing point approximation approximates the posterior as accurately as possible, using sparsified Gaussian processes as the variational family.}
\label{fig:gp-ip-approx}
\end{figure}

\begin{figure}
\begin{subfigure}{0.49\textwidth}
\includegraphics{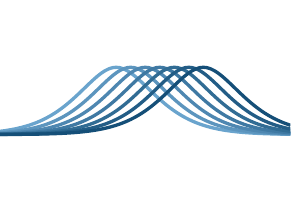}
\caption{Canonical basis functions}
\end{subfigure}
\begin{subfigure}{0.49\textwidth}
\includegraphics{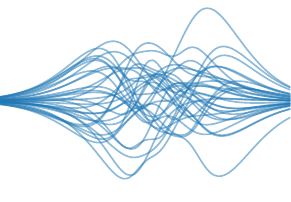}
\caption{Update terms}
\end{subfigure}
\caption[Canonical basis functions]{The type of approximation made using inducing points can be understood in a pathwise manner: the inducing approximation employs the seven displayed canonical basis functions, rather than the thirty-one used in the true posterior, which possess significantly more overlap.
The update term fades to zero as we move away from the data, highlighting the role of the prior in representing uncertainty.}
\label{fig:gp-ip}
\end{figure}

To ensure the approximation is expressive enough, we can either (i) make $\v\mu$ random with a learnable mean and covariance, or, following \textcite{opper09}, (ii) introduce a learned noise covariance $\m\Lambda$ rather than reusing the one from the original model.
From a pathwise perspective, the latter approach gives
\[
(f \given\v{u})(\omega,\v\mu) = f(\omega,\.) + \m{K}_{(\.)\v{z}} (\m{K}_{\v{z}\v{z}} + \m\Lambda)^{-1}(\v\mu - f(\omega,\v{z}) - \v\epsilon(\omega))
\]
where $f\given\v{u}$ is the posterior under the modified likelihood employing the noise covariance $\m\Lambda$.
For $\v{z}$, as with $\v{x}$, we also use bold italics to denotes product structure.
This approximation has a particularly simple interpretation: we can re-write it as
\[
(f \given\v{u})(\omega,\v\mu) = f(\omega,\.) + \ubr{\sum_{j=1}^m v_j(\omega) k(z_j,\.)}_{\t{sparse basis functions}}
\]
where $\v{v}(\omega) = (\m{K}_{\v{z}\v{z}} + \m\Lambda)^{-1}(\v\mu - f(\omega,\v{z}) - \v\epsilon(\omega))$.
Thus, we see that all we have done is replaced a large sum of data-dependent functions $k(x_j,\.)$ with $j=1,..,n$ induced by the kernel with a much smaller sum involving a sparsified set of functions $k(z_j,\.)$ with $j=1,..,m$ and $m \ll n$.
This interpretation applies to most variational approximations, including the classical framework of \textcite{titsias09}, which can be seen in \Cref{fig:gp-ip-approx,fig:gp-ip}.

Given such an approximate posterior, how should we select the introduced hyperparameters $\v{z},\v\mu,\m\Lambda$ to ensure quality?
At minimum, we should attempt to guarantee consistency of the approximation, in the sense that if $m = n$ then one can choose $\v{z} = \v{x}$, $\v\mu = \v\gamma$, and $\m\Lambda = \m\Sigma$ to recover the desired posterior.
We therefore seek to minimize some notion of \emph{distance} on the space of probability measures.

The most natural choice one can consider is arguably the one that emerged from the analysis of Bayes' Rule presented in \Cref{ch:intro}: namely, the \emph{Kullback--Leibler divergence}.
Minimizing this amounts to replacing the measure space in the variational formulation of Bayes' Rule of \Cref{prop:variational-bayes} with the parameterized subspace of measures induced by the set of approximate posteriors with different parameter values

Assume mutual absolute continuity between the distributions of the prior $\pi_f$ and true posterior $\pi_{f\given\v{y}}$.
Define the \emphmarginnote{variational family} $\bb{Q}$ to be the set of all measures equal to the distribution of $f\given\v{u}$ for some choice of variational parameter values $\v{z},\v\mu,\m\Lambda$.
Restricting the variational formulation of Bayes' Rule to $\bb{Q}$ yields the optimization problem 
\[
\argmin_{\bb{q}_f\in\bb{Q}} D_{\f{KL}}(\bb{q}_f \from \pi_f) + \frac{1}{2}\operatorname*{\E}_{f\~\bb{q}_f} (\v\gamma - f(\v{x}))^T \m\Sigma^{-1} (\v\gamma - f(\v{x}))
\]
where we have dropped constant terms from the likelihood density because they do not affect the optima.
Our \emphmarginnote{variational approximation} is obtained by solving this optimization problem.
By the chain rule for Kullback--Leibler divergences, $D_{\f{KL}}(\bb{q}_f \from \pi_f) = D_{\f{KL}}(\bb{q}_{f(\v{z})} \from \pi_{f(\v{z})})$ reduces to the Kullback--Leibler divergence between the respective finite-dimensional marginal distributions at the inducing locations $\v{z}$.
It follows that the optimization objective of the variational approximation is finite.

We thus see that the \emph{inducing point} approximation studied by \textcite{opper09}, when re-interpreted using the variational inference framework of \textcite{titsias09}, coincides with the pathwise variational approximation constructed here.
The same is true for other classes of inducing points, including the variational family originally proposed by \textcite{titsias09}, where $\v\mu$ is randomized.

Better yet, the constructions above are straightforward, principled, and mathematically sound.
In particular, we do not rely on heuristic arguments involving \emph{evidence lower bounds} which, in the absence of connections with the Kullback--Leibler divergence, require us to \emph{posit} that optimizing certain quantities will result in improved posterior approximation.
Instead, by virtue of the Kullback--Leibler divergence generating a Hausdorff topology on the space of probability measures, this is proven.

Inducing point approximations perform well in many settings, and are particularly effective in cases involving large quantities of data that contain redundant information about the posterior.
In certain asymptotic regimes, \textcite{burt19} have shown that the number of inducing points can be taken logarithmic in the size of the data while maintaining approximation accuracy.

Among such approximations, the variational family presented above is particularly interesting because the matrix $\m{K}_{\v{x}\v{x}} + \m\Lambda$ which needs to be inverted is generally better-behaved numerically compared to matrices present in other approximations, owing to the presence of $\m\Lambda$, whose learned values tend to concentrate away from zero.
This makes this choice particularly attractive for use in combination with iterative solvers \cite{dong17,gardner18,pleiss18,meanti20,pleiss20}: we discuss this idea, along with other avenues for future work, in \Cref{ch:discussion}.

This concludes our study of inducing point methods from the pathwise perspective, which are identified with sparsified pathwise approximations where instead of a sum of $n$ kernel terms, we have a smaller sum of $m \ll n$ kernel terms.
We now consider another class of approximations that one can consider using in practice.

\subsection{Approximate priors}
In the preceding sections, we presented a number of posterior approximations which were built by approximating individual terms within the pathwise formalism.
We considered approximations where the prior term is replaced with a finite basis function approximation, and where the data term is replaced with a sparser analog.
An obvious question one can ask is, instead of approximating the prior term, why not just change the model by using a finite basis function prior to begin with?

The answer to this question is that the resulting models become fundamentally finite-dimensional, and lose expressive capacity, often resulting in reduced performance \cite{rasmussen05}.
This is often particularly pronounced in Bayesian optimization using Fourier feature methods \cite{wang18,mutny18}, which have nonetheless attracted attention in that setting owing to their convenience \cite{hernandezlobato14,shahriari15}.
This can be seen precisely by comparing the Fourier feature pathwise update 
\[
\tilde{f}(\omega,\.) + \ubr{\v\phi(\.)^T\v\phi(\v{x})}_{\t{replaces}\,\m{K}_{(\.)\v{x}}} (\ubr{\v\phi(\v{x})^T\v\phi(\v{x})}_{\t{replaces}\,\m{K}_{\v{x}\v{x}} } + \m\Sigma)^{-1}(\v\gamma - \tilde{f}(\omega,\v{x}) - \v\eps(\omega))
\]
to the original pathwise update
\[
f(\omega,\.) + \m{K}_{(\.)\v{x}} (\m{K}_{\v{x}\v{x}} + \m\Sigma)^{-1}(\v\gamma - f(\omega,\v{x}) - \v\eps(\omega))
\]
under the infinite-dimensional prior.
Observe that all that has changed is that the prior $f$ has been replaced with $\tilde{f}$, and the kernel matrices $\m{K}_{(\.)\v{x}}$ and $\m{K}_{\v{x}\v{x}}$ have been replaced with the approximations $\v\phi(\.)^T\v\phi(\v{x})$ and $\v\phi(\v{x})^T\v\phi(\v{x})$.
Observe further that since $\tilde{f}(\omega,\.) = \v\phi(\.)^T\v{w}(\omega)$ where $w_i\~[N](0,1)$, we can write
\[
(f\given\v{y}) \approx \v\phi(\.)^T(\v{w}(\omega) + \v{v}'(\omega))
\]
where $\v{v}'(\omega) = \v\phi(\v{x}) (\v\phi(\v{x})^T\v\phi(\v{x}) + \m\Sigma)^{-1}(\v\gamma - \tilde{f}(\omega,\v{x}) - \v\eps(\omega))$ are random weights.
Therefore, if we change the model, then the posterior becomes a sum of the specified finite basis functions, which do not grow in quantity as data size increases.

\begin{figure}
\begin{subfigure}{0.98\textwidth}
\includegraphics{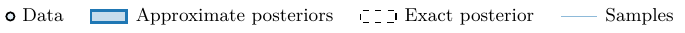}
\end{subfigure}
\begin{subfigure}{0.49\textwidth}
\includegraphics{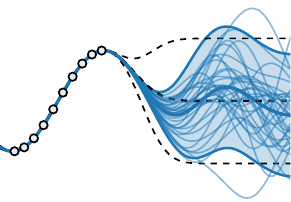}
\caption{Sine and cosine}
\end{subfigure}
\begin{subfigure}{0.49\textwidth}
\includegraphics{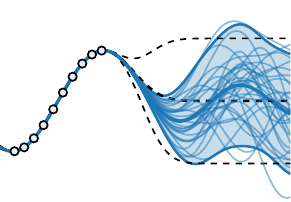}
\caption{Random phase}
\end{subfigure}
\caption[Variance starvation in Fourier feature models]{Example of the \emph{variance starvation} phenomenon in two different random Fourier feature models, compared to the true posterior.
We see two problems: the mean spuriously oscillates and the error bars grow too slowly away from the data.
This can cause the upper confidence bound acquisition, which is the global minima of the error bars, to appear in the completely wrong location.
With the given hyperparameter choices, $n=10$ data points is enough to exhibit considerable approximation error.}
\label{fig:gp-vs}
\end{figure}

If we think of the dimension of the vector space spanned by all basis functions as the representational capacity of the finite-dimensional model, we see that for Fourier feature posteriors this capacity does not grow as data size increases.
This is to be contrasted with the approximate pathwise update, for which the representational capacity grows in spite of its finite-dimensional nature due to the presence of $n$ kernel functions $k(x_j,\.)$ in the sum.

The result is that if the true posterior takes on a difficult-to-represent shape---which will often be the case if $n$ is large enough, but might also occur even if it is not---then performance can degrade disastrously.
This can be seen in \Cref{fig:gp-vs}.
This phenomenon has been called \emph{variance starvation} \cite{wang18,mutny18} in the literature for its tendency to produce error bars that are too narrow due to lack of representational capacity---we note this name is misleading because it is also capable of producing error bars which are far too large.

Variance starvation can be alleviated in a number of ways.
One way is to avoid using Fourier bases for representing the posterior, and rely instead on basis functions which are compactly supported or otherwise possess some sense of locality.
This mirrors ideas in numerical analysis surrounding \emph{Runge's phenomenon} \cite{epperson87,dahlquist08}, where similar behavior occurs in polynomial interpolation.
From this angle, the pathwise viewpoint offers a canonical way to select the functions used for representing the data.

By not approximating terms that do not need to be approximated, pathwise approximations avoid limiting representational capacity and thus retain performance.
Variance starvation has been an ongoing difficulty in Bayesian optimization algorithms for some time---in our view, pathwise approximations of the kind described here, when applicable, largely resolve the issue.

\section{Error analysis}

Pathwise posterior approximations are, ultimately, approximations.
Therefore, a key question one can ask is: how accurate are they?
To quantify this, we need an appropriate notion of \emph{distance} to quantify how far away the two random variables of interest are.
This notion should possess a number of key properties.

\1  It should be \emph{distributional} in nature to reflect the fact that it is the information contained in the posterior that is of interest to us, rather than the precise way in which the random variables are generated.
\2 The distance between the true posterior and pathwise approximations should be finite, in order to facilitate meaningful comparisons.
\0 

\label{ntn:wass-dist}
The \emphmarginnote{Wasserstein distance} \cite{villani08} between probability measures is defined as 
\[
W_{p,d}(\pi,\pi')^p = \inf_{\gamma\in\Gamma(\pi,\pi')} \int_{A\x A} d(a,a')^p \d\gamma(a,a')
\]
where $p \geq 1$, $d$ is a metric on $A$, and $\Gamma(\pi,\pi')$ is this set of all \emph{couplings} of $\pi$ with $\pi'$, namely probability measures $\gamma$ supported on $A\x A$ whose marginals equal $\pi$ and $\pi'$.
We adopt this expression as our notion of distance.

This distributional distance can be understood as the expected distance between two random variables $a$ and $a'$, with distributions $\pi$ and $\pi'$, where the random numbers used to generate $a$ and $a'$ are linked in order to make the expectation as small as possible.
The smallest possible expected distance is zero, which only occurs if the random variables can be made identical to one another---or, in other words, if their distributions coincide.

This definition satisfies both requirements. 
By virtue of varying random numbers---or, more precisely, optimizing over couplings---it compares distributions.
Moreover, unlike alternatives such as the Kullback--Leibler divergence or total variation distance, Wasserstein distances metrize the topology of weak convergence \cite{villani08} , do not impose restrictive absolute continuity requirements, and are finite in the cases of interest.

We also analyze error in a second way: using the supremum norm between kernels.
Observe that pathwise approximations under mean-zero priors possess the exact same mean as the true posterior.
Therefore, when restricted to pathwise approximations under mean-zero priors, this notion gives an actual metric between probability distributions in the given class.
We now state our main results, suppressing ranges from function spaces to ease notation.

\begin{proposition}[Pathwise posterior Wasserstein error bound]
\label{prop:wasserstein-bound}
Assume $X \subseteq \R^d$ is compact with volume $\vol(X)$ and that $f \~[GP](0,k)$ is almost surely continuous.
Let $\tilde{f}(\omega,x) = \sum_{i=1}^\ell w_i(\omega) \phi_i(x)$, and let $(\tilde{f}\given\v{y})(\omega,x) = \tilde{f}(\omega,x) + \m{K}_{(\.)\v{x}}\m{K}_{\v{x}\v{x}}^{-1}(\v{y}- \tilde{f}(\v{x}))$.
Then we have 
\[
W_{2,L^2(X)}(\tilde{f}\given\v{y}, f\given\v{y}) \leq C_1 W_{2,C^0(X)}(\tilde{f},f)
\]
where $C_1 = \del{2\vol(X)\del{1 + \norm{k}^2_{C^0(X \x X)} \norm{\m{K}_{\v{x}\v{x}}^{-1}}^2_{L(\ell^\infty;\ell^1)}}}^{1/2}$.
\end{proposition}

\begin{proof}
The idea is to first use the pathwise update to prove a pointwise bound, then take expectations with respect to a minimizing coupling to obtain a Wasserstein bound.
Using Hölder's inequality with $p=1$ and $q=\infty$, write 
\[
&\abs[1]{(\tilde{f}\given\v{y})(\.) - (f\given\v{y})(\.)}^2 
\\
&\qquad\leq 2 \abs[1]{\tilde{f}(\.) - f(\.)}^2 + 2 \abs[1]{\m{K}_{(\.)\v{x}}\m{K}_{\v{x}\v{x}}^{-1} (\tilde{f}(\v{x}) - f(\v{x}))}^2
\\
&\qquad\leq 2\norm[1]{\tilde{f} - f}^2_{L^\infty(X)} + 2 \norm[1]{\m{K}_{(\.)\v{x}}\m{K}_{\v{x}\v{x}}^{-1}}^2_{\ell^1} \norm[1]{\tilde{f}(\v{x}) - f(\v{x})}^2_{\ell^\infty}
\\ 
&\qquad\leq 2 \del{1 + \norm[1]{\m{K}_{(\.)\v{x}}}^2_{\ell^\infty} \norm[1]{\m{K}_{\v{x}\v{x}}^{-1}}^2_{L(\ell^\infty;\ell^1)}} \norm[1]{\tilde{f} - f}^2_{L^\infty(X)}
\\ 
&\qquad\leq \ubr{2 \del{1 + \norm[1]{k}^2_{C^0(X \x X)} \norm[1]{\m{K}_{\v{x}\v{x}}^{-1}}^2_{L(\ell^\infty;\ell^1)}}}_{C_0} \norm[1]{\tilde{f} - f}^2_{C^0(X)}
\]
where $\norm{\.}_{L(A;B)}$ is the operator norm between $A$ and $B$, $\norm{\.}_{\ell^p}$ is the Euclidean $p$-norm, and where we have used almost sure continuity of sample paths to replace $\norm{\.}_{L^\infty(X)}$ with $\norm{\.}_{C^0(X)}$.
We now lift this bound to a bound on the Wasserstein distance by integrating both sides with respect to an optimal coupling $\gamma\in\Gamma(\tilde\pi,\pi)$, where $\tilde\pi$ and $\pi$ are the distributions of $\tilde{f}$ and $f$, respectively.
Writing
\[
W_{2,L^2(X)}(\tilde{f}\given\v{y}, f\given\v{y})^2 &= \inf_{\gamma\in\Gamma(\tilde\pi, \pi)} \E_\gamma \norm[1]{(\tilde{f}\given\v{y}) - (f\given\v{y})}^2_{L^2(X)}
\\
&\leq C_0 \vol(X) \inf_{\gamma\in\Gamma(\tilde\pi, \pi)} \E_\gamma \norm[1]{\tilde{f} - f}_{C^0(X)}
\\
&\leq C_1^2 W_{2,C^0(X)}(\tilde{f},f)^2
\]
and noting that $C^0(X)$ is a separable metric space, which ensures that the Wasserstein distance over it is well-defined, gives the claim.
\end{proof}

\begin{proposition}[Pathwise posterior kernel error bound]
\label{prop:kernel_bound}
Under the same assumptions as \Cref{prop:wasserstein-bound}, letting $k^{(f\given\v{y})}$, $k^{(\tilde{f}\given\v{y})}$, and $k^{(\tilde{f})}$ be the covariance kernel of these respective processes, we have
\[
\norm[1]{k^{(\tilde{f}\given\v{y})} - k^{(f\given\v{y})}}_{C^0(X \x X)} \leq C_2 \norm[1]{k^{(\tilde{f})} - k}_{C^0(X \x X)}
\]
where $C_2 = n\del{1 + \norm[1]{k}_{C^0(X \x X)} \norm[1]{\m{K}_{\v{x}\v{x}}^{-1}}_{L(\ell^\infty;\ell^1)}}^2$.
\end{proposition}

\begin{proof}
The idea is again to apply standard function-analytic inequalities to the pathwise update.
For a kernel $k$, define the linear operator $M_k : C^0(X \x X) \-> C^0(X \x X)$ by
\[
(M_k c)(\.,\.') &= c(\.,\.') - c(\.,\v{x})\m{K}_{\v{x}\v{x}}^{-1}\m{K}_{\v{x}(\.)} - \m{K}_{(\.)\v{x}}\m{K}_{\v{x}\v{x}}^{-1}c(\v{x},\.)
\\
&\qquad+ \m{K}_{(\.)\v{x}}\m{K}_{\v{x}\v{x}}^{-1} c(\v{x},\v{x}) \m{K}_{\v{x}\v{x}}^{-1}\m{K}_{\v{x}(\.)}
.
\]
By construction, we have that 
\[
k^{(f\given\v{y})} &= M_k k
&
k^{(\tilde{f}\given\v{y})} &= M_k k^{(\tilde{f})}
.
\]
Thus, it suffices to prove that $M_k$ is bounded and calculate its operator norm.
To do so, write 
\[
\norm[1]{M_k c}_{C^0(X \x X)} &\leq \norm[1]{c}_{C^0(X \x X)} + 2 \norm[1]{c(\.,\v{x})\m{K}_{\v{x}\v{x}}^{-1}\m{K}_{\v{x}(\.)}}_{C^0(X \x X)}
\\
&\qquad+ \norm[1]{\m{K}_{(\.)\v{x}}\m{K}_{\v{x}\v{x}}^{-1} c(\v{x},\v{x}) \m{K}_{\v{x}\v{x}}^{-1}\m{K}_{\v{x}(\.)}}_{C^0(X \x X)}
.
\]
We bound these term-by-term.
For the second term, using Hölder's inequality with $p=1$ and $q=\infty$, write 
\[
&\norm[1]{c(\.,\v{x})\m{K}_{\v{x}\v{x}}^{-1}\m{K}_{\v{x}(\.)}}_{C^0(X \x X)}
\\
&\qquad= \sup_{\v{x}',\v{x}'' \in X} c(\v{x}',\v{x})\m{K}_{\v{x}\v{x}}^{-1}\m{K}_{\v{x}\v{x}''}
\\
&\qquad\leq \sup_{\v{x}',\v{x}'' \in X} \norm[1]{c(\v{x}',\v{x})}_{\ell^\infty} \norm[1]{\m{K}_{\v{x}\v{x}}^{-1}}_{L(\ell^\infty;\ell^1)} \norm[1]{\m{K}_{\v{x}\v{x}''}}_{\ell^\infty}
\\
&\qquad\leq \norm[1]{c}_{C^0(X \x X)} \norm[1]{\m{K}_{\v{x}\v{x}}^{-1}}_{L(\ell^\infty;\ell^1)} \norm[1]{k}_{C^0(X \x X)}
\]
where we have used continuity of $k$ to obtain the last inequality.
For the third term, similarly, write 
\[
&\norm[1]{\m{K}_{(\.)\v{x}}\m{K}_{\v{x}\v{x}}^{-1} c(\v{x},\v{x}) \m{K}_{\v{x}\v{x}}^{-1}\m{K}_{\v{x}(\.)}}_{C^0(X \x X)}
\\
&\qquad= \sup_{\v{x}',\v{x}'' \in X} c(\v{x}',\v{x}) \m{K}_{\v{x}\v{x}}^{-1} c(\v{x},\v{x}) \m{K}_{\v{x}\v{x}}^{-1} c(\v{x},\v{x}'')
\\
&\qquad \leq \norm[1]{k}^2_{C^0(X \x X)} \norm[1]{\m{K}_{\v{x}\v{x}}^{-1} c(\v{x},\v{x}) \m{K}_{\v{x}\v{x}}^{-1}}_{L(\ell^\infty;\ell^1)}
\\
&\qquad \leq n \norm[1]{k}^2_{C^0(X \x X)} \norm[1]{\m{K}_{\v{x}\v{x}}^{-1}}^2_{L(\ell^\infty;\ell^1)} \norm{c}_{C^0(X \x X)}
\]
where $n$ factor appears due to use of $\norm[1]{\.}_{L(\ell^\infty;\ell^1)}$.
Putting these inequalities together gives 
\[
\frac{\norm[1]{M_k c}_{C^0(X \x X)}}{\norm{c}_{C^0(X \x X)}} &\leq \Big(1 + 2 \norm[1]{\m{K}_{\v{x}\v{x}}^{-1}}_{L(\ell^\infty;\ell^1)} \norm[1]{k}_{C^0(X \x X)} 
\\
&\qquad+ n \norm[1]{\m{K}_{\v{x}\v{x}}^{-1}}^2_{L(\ell^\infty;\ell^1)} \norm[1]{k}^2_{C^0(X \x X)}\Big) 
\\
&\leq n \del{1 + \norm[1]{\m{K}_{\v{x}\v{x}}^{-1}}_{L(\ell^\infty;\ell^1)} \norm[1]{k}_{C^0(X \x X)}}^2
\]
and so the claim follows.
\end{proof}

The above arguments shows that, for a given dataset size, posterior approximation error is controlled by prior approximation error.
Thus, if we increase approximation accuracy in the prior, we are guaranteed to also increase approximation accuracy in the posterior.

The key idea in the above arguments was to analyze the pathwise approximation by applying standard function-analytic inequalities to the pathwise representation itself.
By varying the resulting norms, a wide set of similar inequalities follow: we make no attempt to optimize bounds, and instead focus on presenting the technique in the simplest manner.

\parmarginnote{Approximation error and asymptotic posterior contraction}
The bounds presented are not tight, particularly in large-data settings where the inverse matrix norm in the constant becomes large.
This may tempt one to conclude that approximation error will grow, but this is often false---instead, the bounds become loose.
One way to see this is by considering posterior contraction: in large-data settings, the posterior will be concentrated, and since our approximations are exact with respect to the mean, there is little room left for error, in the absolute sense, to accumulate.

Different prior approximations possess different error behavior.
The random Fourier feature approach is particularly attractive because its Monte Carlo nature makes it decay at a dimension-free rate of $\sqrt{\ell}$: \textcite{sutherland15} provide bounds on this term.
Note also that the terms in our kernel bounds depend on the dimension only through the kernel matrix, which itself depends on the dimension of the data and not the ambient space.
In this sense, our bounds are very well-behaved with respect to dimension.

This concludes our theoretical presentation and analysis.
We now move to evaluating these ideas in practical settings, and discussing the broad picture painted by the ideas presented here and in the preceding sections.

\section{Parallel Thompson sampling}

We now study empirically how pathwise approximations affect downstream performance of Gaussian process models, using a simple experiment to showcase the key behavior.
For this, we perform a Bayesian optimization experiment using parallel Thompson sampling for acquisition.
This variant of Thompson sampling consists of sampling $p$ independent posterior draws, then computing the acquisition and evaluating the target function at each location simultaneously.
Specifically, for $s=1,..,p$, define
\[
x_{t+1,s}(\omega) &= \argmin_{x\in X} \alpha_{t,s}(\omega,x)
&
\alpha_{t,s}&\~ f \given y_1,..,y_t
\]
where $f$ is the Gaussian process.
Our goal is to understand how approximation error affects performance of Bayesian optimization using this acquisition function, particularly among domains of different dimension.
We focus on data-efficiency rather than computational costs.

We now define the Bayesian model.
Let the domain $X$ to be the unit hypercube in dimensions $d = 2$, $d = 4$, and $d = 8$.
Let $f$ be assigned a Matérn prior with unit variance, length scale $\kappa = \sqrt{d/100}$, and smoothness $\nu = 5/2$.
Each observation is assigned an independent Gaussian likelihood with error variance $\sigma^2 = 10^{-3}$.
From these, we obtain the posterior distribution.

We select target functions to minimize by sampling from a random Fourier feature approximation of the prior, and obtain their minima for purposes of regret using multi-start gradient descent.
To account for variable problem difficulty, we allow the number of evaluations to vary according to dimension, setting $n=64$ for $d=2$, $n = 256$ for $d = 4$, and $n = 1024$ for $d = 8$.
Similarly, we set the degree of parallelism to allow $p = d$ simultaneous evaluations.
We repeat each experiment $32$ times to assess variability.

We consider two sequential baselines and three approximate acquisition strategies using the posterior Gaussian process.
For the sequential baselines, we use \emph{random search} \cite{bergstra12} and \emph{dividing rectangles} \cite{jones93}, selected for their simplicity.
For the Gaussian process baselines, we consider (a) a \emph{Cholesky} grid-based approach, (b) a \emph{random Fourier feature} posterior process, and (c) a \emph{pathwise} approximate posterior using the same Fourier feature approximation for the prior term.
We now describe these in more detail.

\begin{figure*}[p!]
\includegraphics{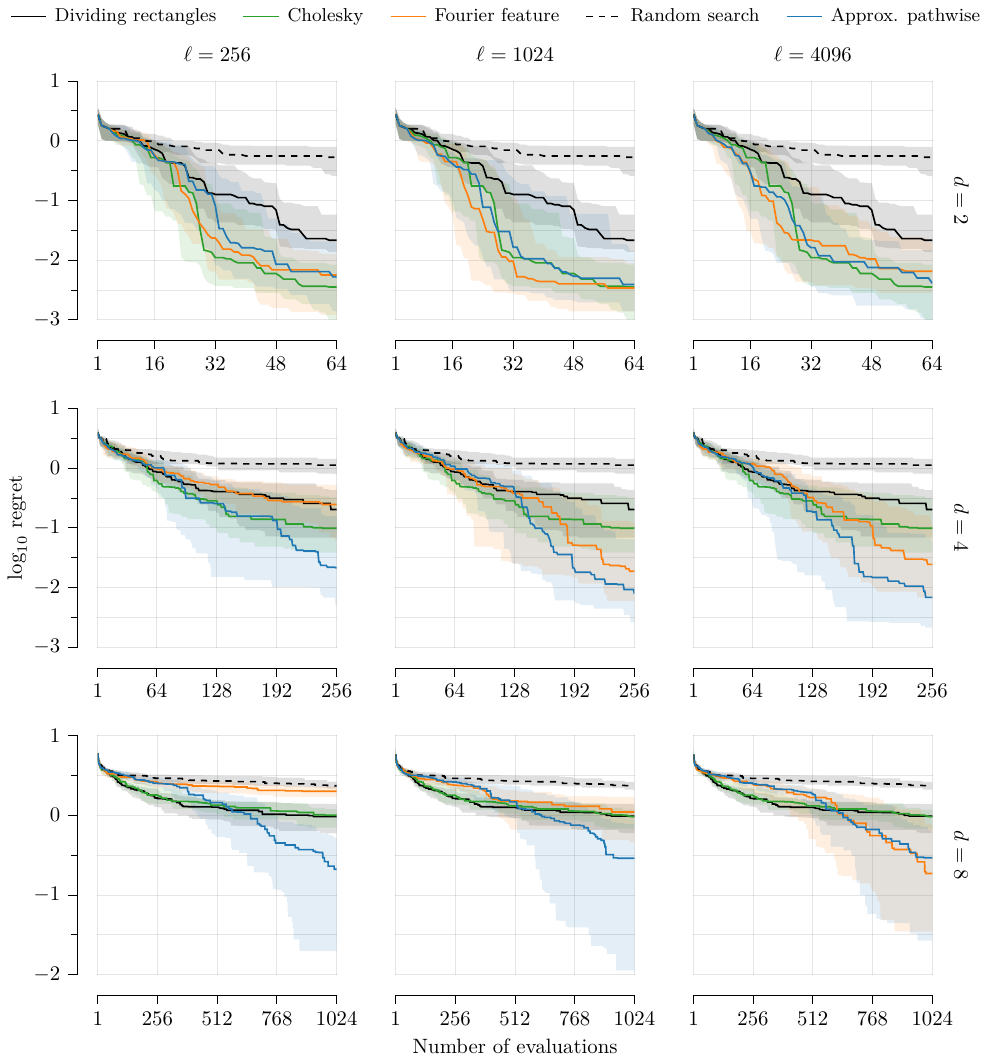}
\caption[Parallel Thompson sampling benchmark results]{Parallel Thompson sampling benchmark results.}
\label{fig:parallel-ts}
\end{figure*}

The Cholesky grid-based approach works by (i) drawing a set with $250,000$ points uniformly at random, (ii) sampling a vector of independent Gaussians whose mean and variance are equal to the posterior predictive distribution at those points, (iii) choosing the $2048$ smallest elements, (iv) sampling the posterior Gaussian process jointly at the chosen locations, and (v) choosing the smallest value.
This gives an approximation to the true acquisition locations using the candidate locations on the grid.

The random Fourier feature and pathwise approximate posterior approaches do not use grids.
Instead, we (i) draw a set of $250,000$ points uniformly at random, (ii) evaluate $f \given \v{y}$ at those locations to choose the $32$ smallest points, and (iii) run multi-start optimization using L-BFGS-B \cite{byrd95} from each candidate location to find the minima.
This procedure is selected so as to be relatively similar to the grid-based approach and ensure a fair comparison.

To understand the effect of posterior approximation, we examine both basis-function-based Gaussian processes with a total of $\ell = 256$, $\ell = 1024$, and $\ell = 4096$ basis functions.
For the pathwise approximate posterior, this is the sum of both the number of basis functions used for the prior with the number of canonical basis functions---this choice helps avoid unfairly penalizing the random fourier Feature model.
We use random-phase-based Fourier features in both approximate posteriors.

Results can be seen in \Cref{fig:parallel-ts}, where we plot median regret curves, along with first and third quartiles.
Immediately, we see that the effect of the strategy used for computing the acquisition function varies according to dimension.

In the two-dimensional setting, all Gaussian process methods perform comparably, outperforming the random search and dividing rectangles baselines.
This occurs even in the regime with $\ell = 256$ basis functions, and shows that posterior approximation error does not play a particularly significant role for the given comparisons and parameters.

In the four-dimensional setting, the behavior is different.
This time, in the $\ell = 256$ case, the pathwise approximate posterior outperforms both other Gaussian process baselines.
Here, the random Fourier feature baseline performs worse than the Cholesky baseline.
If we increase the amount of basis functions, the random Fourier feature baseline recovers this difference, yielding relatively comparable performance to the pathwise approximation for $\ell = 1024$ and $\ell = 4096$, and outperforming the Cholesky baseline.

In the eight-dimensional setting, the pathwise approximation outperforms all baselines for $\ell = 256$ and $\ell = 1024$.
In those cases, both the random Fourier feature and Cholesky baselines are hampered by the curse of dimensionality, and perform no better than the dividing rectangles baseline---for $\ell = 256$, random Fourier features are comparable to random search.
With $\ell = 4096$ basis functions, the performance of Fourier features recovers.
In contrast, the pathwise approximation is insensitive to the number of basis functions.

In summary, these results largely confirm the viewpoint developed using the preceding theory.
We saw previously that the pathwise approximate posterior can be seen as modifying the random Fourier feature posterior by replacing certain Monte Carlo approximations with their expected values.
It is therefore reasonable to suppose that this gives a more accurate approximation of the true posterior.
Our results are thus consistent with the true Gaussian process being a better-performing model for Bayesian optimization.

For the Matérn prior used, this mirrors modern understanding: low-rank and other finite-dimensional models are generally less expressive than true Gaussian processes and mostly favored in cases where their properties help control computational costs.
Using the true Gaussian process, or a more accurate approximation thereof, can result in better performance.

The benefits of using a method that avoids solving arbitrarily large linear systems at test-time are clear from the point of view of both computational complexity and numerical stability.
From these perspectives, our results show that already in $d=4$ the improvements are enough to make a noticeable difference in downstream tasks.

\section{Conclusion}

In the Gaussian case, the distributional notion of \emph{conditioning} can be re-formulated using random variables, yielding a notion of \emph{pathwise conditioning}.
In the preceding sections, we developed and studied this point of view for Gaussian processes, allowing us to express a posterior Gaussian process as the sum of a \emph{prior term}, and a \emph{data-dependent term}.

Pathwise conditioning gives a powerful way to think about posterior Gaussian processes.
Using this notion, we re-interpreted classical methods such as \emph{random Fourier features} and \emph{inducing points}, by observing that they correspond to approximations made to individual terms within the pathwise update.
Crucially, we observed that one could approximate different terms within the formula individually.

Using this observation, we constructed accurate finite-dimensional posterior approximations which nonetheless yield actual \emph{random functions}.
These functions are sums of two sets of finite basis functions with dependent random coefficients.
These coefficients can be sampled in advance, after which the functions effectively become deterministic and can be evaluated at arbitrary points, as well as differentiated using standard techniques.

The resulting approximations are particularly useful in decision-making settings such as \emph{Bayesian optimization}, where acquisition functions constructed from Gaussian process sample paths need to be optimized or evaluated at arbitrary locations.
This makes it possible to implement Thompson sampling using automatic differentiation in a straightforward manner without any sophisticated bookkeeping.
In particular, there is no need to track at what points the Gaussian process has already been evaluated at.

Using the presented technique, minimizing a Gaussian process sample path can be done in linear time and without incurring large approximation error that degrades performance as part of the iterative minimization procedure itself.
This is particularly useful in higher-dimensional settings, where grid-based methods and other alternatives may be hampered by the curse of dimensionality.

The performance of pathwise approximate posterior Gaussian processes depends on the approximation accuracy of the prior term.
We presented a number of methods for doing so, each suited to their respective settings.
For stationary kernels, random Fourier feature methods are particularly attractive due to their good behavior with respect to dimension, but are by no means the only choice.
We hope these developments prompt further study of other possible choices for approximate priors.

Pathwise approximations are also of interest as a potential organizing principle for designing Gaussian process software packages.
In particular, one can consider implementing a Gaussian process with separate methods for evaluating the process and re-sampling its random weights.
Depending on the situation, this might be a more convenient application programming interface than working with distributions at user-specified locations.
Exploring these alternatives is a promising avenue for further work.

Similarly, pathwise approximations make a strong case that efficiently sampling from a Gaussian process prior is a key software primitive that a Gaussian process package should support as part of its kernel implementations. 
Understanding how to organize different approximations, which may possess different properties and may also be computed in different ways, is another promising avenue for future work.

Gaussian processes have been applied in many settings, ranging from areas such as spatial statistics where they are mostly used by humans, to areas such as Bayesian optimization and model-based reinforcement learning where they are mostly used by computers.
We hope that the contributions presented here improve their ease of use, broaden their applicability, and enable new applications not yet considered.
We now proceed to study a different route for expanding the set of settings Gaussian processes can be applied in.

\chapter{Non-Euclidean Matérn Gaussian Processes}
\label{ch:noneuclidean}

\lettrine{S}{tationary kernels} on Euclidean spaces are one of the most widely-used Gaussian process model classes.
These kernels are attractive because they work effectively and it is generally easy to understand the kind of prior information they introduce into a problem.
This makes them a valuable tool for practitioners to use in the manner required for the task at hand.

Our goal throughout has been to expand the settings in which Gaussian process models and decision systems built atop them can be used.
We have so far focused on doing so by making existing models easier to work with, but now pursue a different approach: namely, we focus on expanding the scope of models one can consider working with.
We again emphasize constructiveness and making abstract ideas accessible to practitioners.

We focus on the setting of Riemannian geometry, which describes a widely-occurring class of geometric shapes and spaces.
We thus study Gaussian processes whose domains are manifolds, rather than Euclidean spaces.
Working with manifolds often involves thinking carefully about discretization: we therefore also study purely discrete settings involving meshes and weighted undirected graphs, which are of inherent interest in their own right.

Our guiding theme is to ask: how can we make Gaussian processes on Riemannian manifolds be just as effective a model class as their Euclidean counterparts?
We proceed to introduce, develop, and explore this topic.

\section{Riemannian Matérn Gaussian processes}

We begin with a key question: how should one generalize the most widely-used class of Gaussian process models to the Riemannian manifold setting?
There are multiple potential definitions one can introduce, some of which turn out to be much better-behaved mathematically than others.
In order to pursue these questions, we begin by introducing and reviewing the setting of differential geometry.

\subsection{Review of differential geometry}
\begin{figure}
\tikzset{external/export next=false}
\begin{tikzpicture}
\node at (0,0) {\includegraphics[scale=0.25]{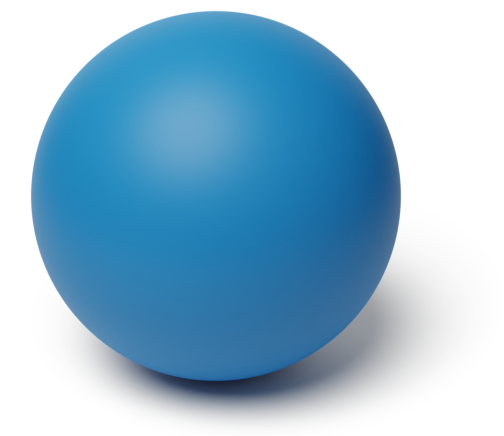}};
\node at (5.5,-0.5) {\includegraphics[scale=0.25]{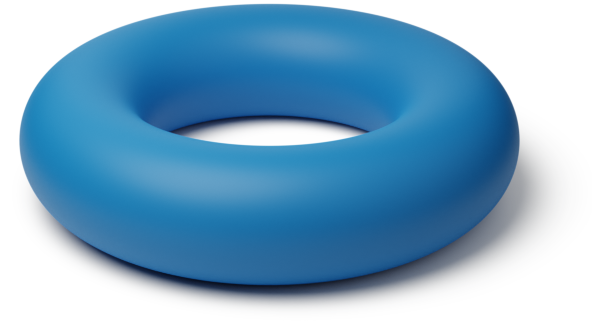}};
\end{tikzpicture}
\caption[Common manifolds]{Two common manifolds: the sphere $\mathbb{S}^2$ and torus $\mathbb{T}^2$.}
\label{fig:common-manifolds}
\end{figure}

We now briefly review the notions needed to define and understand manifolds.\footnote{The Russian word for \emph{manifold} is \emph{mnogoobrazie} (mn\rotatebox[origin=c]{180}{e}g\rotatebox[origin=c]{180}{aa}br{\textquotesingle}az\textsuperscript{j}\textsc{i}j\rotatebox[origin=c]{180}{e}). The word \emph{mnogo} means \emph{many}, and the word \emph{obraz} roughly means \emph{form}, \emph{view}, or \emph{image}. 
Thus, a \emph{manifold} is a \emph{space with many views into it}---a name one could easily call well-chosen and evocative.}
These are sets equipped with additional structure encoding their geometric shape.
Many geometric shapes, in turn, can be realized as manifolds: two examples are shown in \Cref{fig:common-manifolds}.
We use the works of \textcite{lee10,lee12,lee18} as our primary references on this topic.
For a machine-learning-oriented text, see \textcite{bronstein21}.

A \emphmarginnote{topological space} is a set $X$ together with a \emph{topology}, which is a collection of subsets $\c{O}_X \subseteq 2^X$ called the \emph{open sets}---these include the empty set, the space itself, and are closed under pairwise intersections and arbitrary unions.
A given set can admit many topologies.
Topological spaces admit notions of \emph{locality}, \emph{convergence}, \emph{continuity}, \emph{compactness}, \emph{paracompactness}, \emph{denseness}, and many others.
The \emph{Hausdorff} property implies that limits are unique.

\label{ntn:manifold}
\parmarginnote{Topological manifold}
A paracompact Hausdorff topological space $(X,\c{O}_X)$ is called a $d$-dimensional \emph{topological manifold} if it is locally homeomorphic to $\R^d$ equipped with the standard topology.
For such a manifold, there is a set $\c{A}_X$ called the \emph{atlas} whose elements are the local homeomorphisms, called \emph{charts}.
An atlas is \emph{maximal} if it is the largest possible such set: for any given atlas, a maximal atlas containing it it is unique.

A \emphmarginnote{smooth differentiable manifold} is a triple  $(X,\c{O}_X,\c{A}_X)$ where $X$ is a topological manifold, and $\c{A}_X$ is a $C^\infty$-atlas.
A $C^\infty$-atlas is an atlas in which, for any charts $x,y \in \c{A}_X$ with overlapping domain, the \emph{chart transition map} $y\after x^{-1}$ is an infinitely differentiable function.
Homeomorphisms compatible with such charts are called \emph{diffeomorphisms}.
A smooth manifold is called \emph{oriented} if the Jacobian determinant of all chart transition maps in its atlas is positive.
Such atlases admit maximal analogs.

\label{ntn:smooth-fns}
\parmarginnote{Smooth function}
The notion of smoothness extends to real-valued functions on manifolds, and functions between manifolds.
For a smooth manifold $X$, we say that $f \in C^\infty(X;\R)$ if it is infinitely differentiable in any chart.
Similarly, we say that a map $f : X \-> Y$ between a pair of smooth manifolds is smooth if it is infinitely differentiable in any pair of charts.
Smooth maps between manifolds are automatically continuous.
To ease notation, we often omit real ranges from function spaces that occur in the sequel.

A \emphmarginnote{manifold with boundary} is one of a wide class of spaces generalizing the notion of a manifold while preserving most geometric characteristics that make manifolds useful and interesting in the first place.
Such a space is defined analogously to an ordinary manifold, except that it is also possible for it to be locally homeomorphic to a Euclidean half-plane.
The ideas we consider admit natural generalizations to such settings, but we do not pursue these, and in particular only consider manifolds without boundary.

\label{ntn:tangent-bdl}
Every smooth manifold $X$ gives rise to its \emphmarginnote{tangent bundle} $(TX,\proj_X,X)$, which contains a smooth manifold $TX$ and a surjective projection map $\proj_X : TX \-> X$.
The former is defined as $TX = \U_{x\in X} \{(x,v) : v \in T_x X\}$, where $T_x X$ is the tangent space, defined as a vector space of equivalence classes of directional derivatives of smooth curves through $x$.
This set can be equipped with the structure of a smooth manifold, arising canonically from the smooth structure of $X$.

\label{ntn:vector-pushforward}
Given a smooth map $f: X \-> Y$ that transforms a smooth manifold $X$ into another smooth manifold $Y$, the \emphmarginnote{pushforward map} $f_* : TX \-> TY$ describes how each tangent space $T_x X$ transforms into $T_{f(x)} Y$ by the action of $f$.
Since $f$ is smooth, each tangent space is transformed linearly: in this sense, $f_*$ encodes the local behavior of $f$ around each point on its domain.

\label{ntn:smooth-sections}
A map $f : X \-> TX$ satisfying $\proj_X \after f = \id_X$ is called a \emphmarginnote{vector field}.
Note that this is \emph{not the same} as a map $X \-> \R^d$ and often cannot usefully be expressed in this way: we expand on this in the sequel.
We say that $f$ is a \emph{cross-section}, or simply a \emphmarginnote{section}, of the tangent bundle.
If $f$, when viewed as a map between manifolds, is smooth, we write $f \in \Gamma(TX)$.
Functions can multiply against vector fields by scaling them pointwise.

\label{ntn:cotangent-bdl}
Analogously, we can define the \emphmarginnote{cotangent bundle} $T^* X = \U_{x\in X} \{(x,\phi) : \phi \in T_x X^*\}$ where, compared to the tangent bundle, we have replaced the vector space $T_x X$ with its topological dual $T_x X^*$.
A section of this bundle is called a \emphmarginnote{covector field}.
We can similarly define a number of other bundles, each equipped with a projection map onto the base space.
Introducing sections of such bundles gives rise to the notion of a \emphmarginnote{tensor field} and other generalizations.
All such maps are called smooth if they are smooth maps between manifolds.

\parmarginnote{Vector bundle}
\emph{Vector bundles} are a particularly important class of bundles.
Given the projection map $\proj_X$ onto the base space, the \emph{fiber} over $x$ is defined as the preimage $\proj_X^{-1} \{x\}$.
In a vector bundle, all fibers admit the structure of a vector space.
This means that a space of sections of a vector bundle can itself be given the structure of a vector space. 
Many vector spaces which generalize the usual function spaces, such as spaces of vector fields equipped with additional structure, are constructed using suitable vector bundles.

\parmarginnote{Interior product}
Vector fields can be inserted into covector fields by pairing vectors in each tangent space: this defines the \emph{interior product} $\mathbin{\lrcorner}$, which extends to tensor fields as long as the pairing being considered makes sense.
We say that a totally antisymmetric $(0,k)$-tensor field is a $k$-form, and that a smooth function is a $0$-form.
The \emphmarginnote{exterior derivative}, denoted by $\d$, maps $k$-forms into $(k+1)$-forms.
Interior products preserve antisymmetry.

\parmarginnote{Volume}
Differential forms are often used to formalize and encode geometric structure in a way that is amenable to analysis.
For example, a nowhere-vanishing $d$-form is called a \emph{volume form}---such a form induces a notion of a \emph{volume density}, which induces a notion of \emphmarginnote{integration} of smooth functions, and in turn, by the Riesz--Markov--Kakutani representation theorem, a Radon measure on the manifold, which we call the \emph{volume measure}.
On a general smooth manifold, the choice of a volume form is heavily non-unique.

\label{ntn:metric-tensor}
A \emphmarginnote{Riemannian manifold} is a quadruple $(X,\c{O}_X,\c{A}_X,g)$, usually written $(X,g)$, where $X$ is a smooth manifold and $g$ is the \emph{metric tensor}, which is a smooth symmetric positive definite $(0,2)$-tensor field.
The metric tensor can be thought of as an algebraic object encoding the manifold's quantitative shape, and canonically gives rise to notions such as \emph{volume}, \emph{integration}, and \emph{geodesics}.
In particular, we denote the volume form, volume measure, and volume density induced by the metric tensor as $\vol_g$.

\parmarginnote{Nash Embedding}
Riemannian manifolds are, by definition, topological spaces with additional structure.
By Nash's Embedding Theorem, every Riemannian manifold can be isometrically embedded within an $\c{O}(d^2)$-dimensional Euclidean space.
This identifies Riemannian manifolds as geometric shapes located within Euclidean spaces.
This perspective is often avoided for both technical and conceptual reasons: if the spacetime of the universe is a manifold, can an ambient Euclidean space actually have physical meaning?

\subsection{The Laplace--Beltrami operator}

Riemannian manifolds admit many different objects, such as connection one-forms, Ricci tensors, and other constructions which encode and mathematically describe their geometric properties.
These are generally built using the metric along with derivatives and other operations.
The \emph{Laplace--Beltrami operator} is one such object, and forms a basic building block for defining differential equations on manifolds.

\label{ntn:laplace-beltrami}
\begin{definition}[Laplace--Beltrami operator]
Let $(X,g)$ be a Riemannian manifold, assumed oriented without loss of generality.
Define the \emph{divergence} of a vector field to be the unique map 
\[
\f{div}_g : \Gamma(TX) &\-> C^\infty(X)
&
\d(v \mathbin{\lrcorner} \vol_g)  &=  \f{div}_g v\.\vol_g
\quad
\forall v \in \Gamma(TX)
\]
where $\.$ is pointwise multiplication of differential forms by smooth functions, and the \emph{gradient} of a scalar function to be the unique map 
\[
\f{grad}_g : C^\infty(X) &\-> \Gamma(TX)
&
g(\f{grad}_g f,v) &= (\d f) \mathbin{\lrcorner} v
\quad
\forall v \in \Gamma(TX)
.
\]
Define the \emph{Laplace--Beltrami operator} to be 
\[
\lap_g : C^\infty(X) &\-> C^\infty(X)
&
\lap_g f &= \f{div}_g \f{grad}_g f
.
\]
\end{definition}

The Laplace--Beltrami operator, then, maps functions into the divergence of their gradient at every point.
There are multiple equivalent ways of defining the Laplace--Beltrami operator: an alternative is to define it as the trace of the Riemannian Hessian, and another is to define it in coordinates.
The latter expression illustrates that a Laplace--Beltrami operator maps functions into their locally averaged analogs, in a sense.
For us, the relatively rich spectral properties possessed by this operator will be of key interest.

\begin{result}
Let $(X,g)$ be a compact Riemannian manifold.
Then the operator $-\lap_g : C^\infty(X) \-> C^\infty(X)$ is positive semi-definite, and extends uniquely to a self-adjoint unbounded positive operator $-\lap_g: D(\lap_g) \-> L^2(X)$, where $D(\lap_g) \subseteq L^2(X)$.
\end{result}

\begin{proof}
\textcite[Theorem 2.4]{strichartz83}.
\end{proof}

\begin{figure}
\begin{subfigure}{0.19\textwidth}
\includegraphics{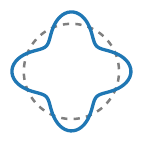}
\end{subfigure}
\begin{subfigure}{0.19\textwidth}
\includegraphics{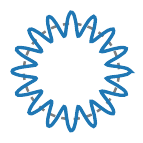}
\end{subfigure}
\begin{subfigure}{0.59\textwidth}
\tikzset{external/export next=false}
\begin{tikzpicture}
\node at (0,0) {\includegraphics[scale=0.25]{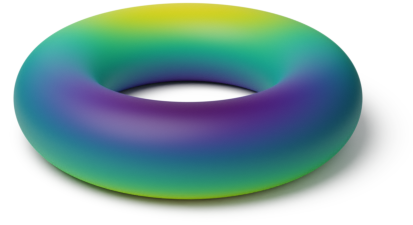}};
\node at (3.5,0) {\includegraphics[scale=0.25]{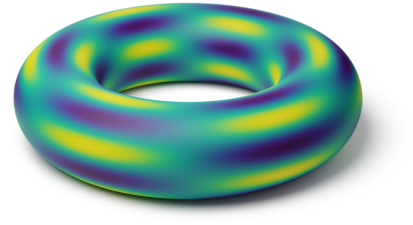}};
\end{tikzpicture}
\end{subfigure}
\caption[Laplace--Beltrami eigenfunctions: circle and torus]{Eigenfunctions of the Laplace--Beltrami operator on a circle and torus.}
\label{fig:eig-s1-t2}
\end{figure}

Viewed in this way, the range of the Laplace--Beltrami operator is a Hilbert space, enabling us to use ideas from spectral theory to better understand its behavior.
In particular, one can show that, owing to compactness, the spectrum of $-\lap_g$ is discrete---this gives a particularly simple view of $-\lap_g$ through its eigenvalues and eigenfunctions.

\label{ntn:laplace-beltrami-eigenpairs}
\begin{result}[Sturm--Liouville decomposition]
Let $(X,g)$ be a compact Riemannian manifold.
Then there exists an orthonormal basis $f_n$, $n\in\Z_+$, of $L^2(X)$ such that $-\lap_g f_n = \lambda_n f_n$ with $0 = \lambda_0 \leq \lambda_1 \leq .. \leq \lambda_n$ and $\lambda_n\-> \infty$ as $n\->\infty$.
Moreover, $-\lap_g$ admits the representation
\[
-\lap_g f = \sum_{n=0}^\infty \lambda_n \innerprod{f}{f_n} f_n
\]
which converges unconditionally in $L^2(X)$ for all $f \in D(\lap_g)$.
\end{result}

\begin{proof}
\textcite[139]{chavel84}, or \textcite[Theorem 44]{canzani13}.
\end{proof}

\begin{figure}
\tikzset{external/export next=false}
\begin{tikzpicture}
\node at (0,0) {\includegraphics[scale=0.25]{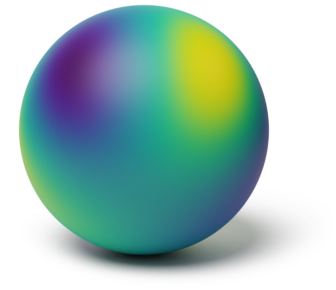}};
\node at (2.625,0) {\includegraphics[scale=0.25]{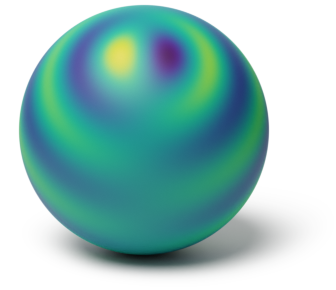}};
\node at (5.75,0) {\includegraphics[scale=0.25]{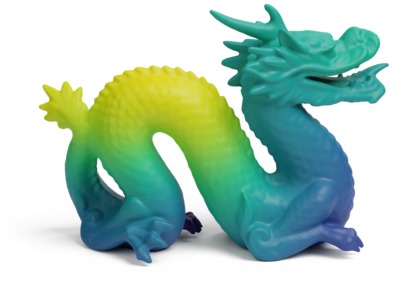}};
\node at (9.125,0) {\includegraphics[scale=0.25]{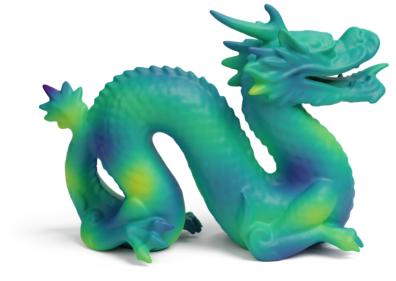}};
\end{tikzpicture}
\caption[Laplace--Beltrami eigenfunctions: sphere and dragon manifold]{Eigenfunctions of the Laplace--Beltrami operator on a sphere and dragon manifold.}
\label{fig:eig-s2-dr}
\end{figure}

This is a powerful result: the eigenfunctions $f_n$, shown in \Cref{fig:eig-s1-t2,fig:eig-s2-dr}, can be viewed as analogs of the Fourier basis adapted to the manifold's geometry.
By expanding a function within the basis of eigenfunctions, we obtain an infinite sequence of basis coefficients---just like representing a periodic function in $L^2([-\pi,\pi];\R)$ by an infinite sum of complex exponentials.
The Sturm--Liouville decomposition gives rise to a notion of \emph{functional calculus}, which is key for our purposes, and which we now introduce.

\label{ntn:functional-calculus}
\begin{definition}[Functional calculus]
Let $\Phi : [0,\infty) \-> \R$. 
Define the (possibly unbounded) operator $\Phi(-\lap_g) : D(\Phi(-\lap_g)) \-> L^2(X)$ by
\[
\Phi(-\lap_g) f = \sum_{n=0}^\infty \Phi(\lambda_n) \innerprod{f}{f_n} f_n
\]
where $D(\Phi(-\lap_g)) = \{f \in L^2(X) : \sum_{n=0}^\infty \abs{\Phi(\lambda_n)}^2 \abs{\innerprod{f}{f_n}}^2 f_n < \infty\}$.
\end{definition}

Functional calculus lets us extend the idea of applying functions from numbers to operators.
This is done by applying the function of interest to the eigenvalues of the operator.
We will use this to construct Gaussian processes as solutions of stochastic partial differential equations.
First, however, we take a step back and examine why one would want to take this rather roundabout approach in the first place as opposed to a more direct alternative.

\subsection{A no-go theorem for kernels on manifolds}

Here, we begin exploring the idea of defining Gaussian processes whose domains are Riemannian manifolds.
Recall that to define such a Gaussian process, we need to define its kernel, which is a positive semi-definite function $k : X \x X \-> \R$.

Before attempting to do so in generality, consider first how to extend the Euclidean squared exponential kernel.
The simplest idea one can consider is to replace the Euclidean distance $\norm{x-x'}$ with the Riemannian \emph{geodesic distance} $d_g(x,x')$, defined as the length of the shortest path between $x$ and $x'$.
To ensure $d_g$ behaves like the Euclidean distance, assume that $X$ is \emph{complete} with respect to $d_g$, meaning that all sequences which eventually become close in $d_g$ also converge with respect to $X$.
Consider
\[
\sigma^2 \exp\del{-\frac{d_g(x,x')^2}{2\kappa^2}}
\]
as a candidate kernel.
We must then ask: is this expression necessarily well-defined? 
In particular, is it positive semi-definite for all $\kappa$?

In the Euclidean setting, we can prove positive-semi-definiteness by first proving it for linear kernels, then showing sums, products, and limits of kernels are positive semi-definite, thereby constructing the kernel piece-by-piece.
The argument clearly cannot extend to the manifold setting, where there are no linear kernels.
It turns out that no argument can, because in the manifold setting the corresponding claim isn't true.

\begin{result}
Let $(X,g)$ be a complete Riemannian manifold.
If the geodesic squared exponential kernel is positive semi-definite for all $\kappa > 0$, then $X$ is isometric to a Euclidean space.
\end{result}

\begin{proof}
\textcite[Theorem 2]{feragen15}.
\end{proof}

It turns out that even more can be said.
In a metric space $(X,d)$, the \emph{length} of a path  $\gamma : [0,L] \-> X$ is defined as the least upper bound on the total distance between finite sets of successive points along the curve.
A path is called \emph{geodesic} between $x$ and $x'$ if $\gamma(0) = x$, $\gamma(L) = x'$, and $d(\gamma(t),\gamma(t')) = |t - t'|$ for all $t,t'\in[0,L]$.
A metric space is called a \emph{geodesic space} if every pair of points is connected by a geodesic.

\begin{result}
Let $(X,d)$ be a geodesic space.
If the geodesic squared exponential kernel is positive semi-definite for all $\kappa > 0$, then $X$ is flat in the sense of Alexandrov.
\end{result}

\begin{proof}
\textcite[Theorem 2]{feragen15}.
\end{proof}

See \textcite[Chapter 26]{villani08} for a definition of flatness in the above sense, and a discussion on its interpretation and relationship to various notions of curvature.
Complete Riemannian manifolds are geodesic spaces, but the former result is sharper than the latter: in particular, the torus equipped with the product metric is flat, but is not isometric to a Euclidean space.

These results are an absolute disaster for the geodesic squared exponential kernel, and give good reason to completely abandon this approach.
The fundamental issue is that there are few useful tools for proving positive-semi-definiteness of geodesic kernels, and, in light of the above results, it isn't obvious which functions are going to be positive semi-definite in the first place and which are not.
We therefore explore the alternative, differential-equation-based approach mentioned in the preceding section.

\subsection{Stochastic partial differential equations}

We now develop an appropriate formalism for defining Gaussian processes on Riemannian manifolds.
Rather than building such processes by defining kernels, loosely speaking, we construct them directly as affine maps of white noise processes, which we think of as infinite-dimensional standard Gaussians.
This yields positive semi-definite kernels implicitly defined as covariances of said processes.

We begin by discussing key notions of abstract Gaussian processes defined on general vector spaces.
Specifically, we work with Gaussian processes in the sense of duality whose test functionals are determined by a Hilbert space.

\begin{definition}[Generalized Gaussian~field]
A \emph{centered generalized Gaussian field} $f$ over a Hilbert space $H$ is a stochastic process $f : \Omega \x H \-> \R$ satisfying two key properties.
\1 $\E(f(\.,h)) = 0$ for all $h \in H$.
\2 There exists a bounded linear self-adjoint non-negative operator $\c{K}$ on $H$, called the \emph{covariance operator}, such that 
\[
\E(f(\.,h)f(\.,h')) = \innerprod{\c{K}h}{h'}
\]
for all $h,h' \in H$.
\0 
\end{definition}

We take the right-hand-sides of our stochastic partial differential equations to be such stochastic processes. 
Before continuing, we prove that if $H$ is a reproducing kernel Hilbert space, then generalized Gaussian fields can be reinterpreted as Gaussian processes in the classical sense.

\begin{proposition}
Let $f : \Omega \x H \-> \R$ be a centered generalized Gaussian field defined over a reproducing kernel Hilbert space $H$ with identity covariance operator $\id : H \-> H$.
Then letting $\f{ev}_x \in H^*$ be a pointwise evaluation functional and $\Psi : H \-> H^*$ be the bijective linear isometry given by the Riesz Representation Theorem, the stochastic process $f : \Omega \x X \-> \R$ defined by $f(\omega, x) = f(\omega,\Psi^{-1} \f{ev}_x)$ is a Gaussian process whose covariance is given by the reproducing kernel of $H$.
\end{proposition}

\begin{proof}
It is clear that the resulting map is a centered Gaussian process, so it suffices to compute its covariance.
Let $\f{ev}_x \in H^*$ and $\f{ev}_{x'} \in H^*$ be pointwise evaluation functionals.
Then
\[
\Cov(f(\.,x),f(\.,x')) &= \Cov(f(\.,\Psi^{-1} \f{ev}_x),f(\.,\Psi^{-1} \f{ev}_{x'})) 
\\
&= \innerprod{\Psi^{-1}\f{ev}_x}{\Psi^{-1}\f{ev}_{x'}}_H 
\\
&= \innerprod{\f{ev}_x}{\f{ev}_{x'}}_{H^*}
\\
&= k(x,x')
\]
using the reproducing property, and the claim follows.
\end{proof}

Note that the resulting Gaussian process in general will \emph{not} have sample paths in $H$ almost surely, not even up to a choice of version.
Sample paths will instead generally lie in a less regular function space---this subtlety provides much of the motivation behind introducing generalized Gaussian fields in the first place, rather than working purely in terms of sample paths.
In the other direction, if $k$ is a kernel, define the \emph{covariance operator} by 
\[
\c{K} : \phi \|> \int_X \phi(x)k(x,\.) \d x
.
\]
One can see that a centered Gaussian process $f : \Omega \x X \-> \R$ induces a centered generalized Gaussian field $f : \Omega \x H \-> \R$ with covariance operator $\c{K}$, provided that $H$ is chosen appropriately.
For instance, if $f$ is regular enough that its samples lie in $L^2(X)$ almost surely, one can take $H = L^2(X)$ and $\c{K}$ as above.
This clarifies how this general notion relates to Gaussian processes in the standard sense.
To continue, we introduce the notion of a \emph{solution} of a stochastic partial differential equation.

\begin{definition}[Stochastic partial differential equation]
Let $H$ be a Hilbert space, let $\c{L} : F \-> H$ be a bounded linear operator, and let $\c{W}$ be a centered generalized Gaussian field on $H$.
Then the zero-mean generalized Gaussian field $f$ over $F$ is a solution of the abstract stochastic partial differential equation 
\[
\c{L} f = \c{W}    
\]
if, letting $\c{L}^*$ be the adjoint of $\c{L}$, for every $h\in H$ we have that 
\[
f(\omega,\c{L}^* h) = \c{W}(\omega,h)
\]
holds almost surely.
\end{definition}

For our purposes, it also suffices to replace the almost sure equality with equality in distribution. 
Even if one considers this weaker notion, the same kind of results and calculations follow in our setting, and work equally well in both cases.
We therefore do not dwell on this distinction.

The idea behind this definition is to take advantage of Gaussianity and use it to avoid even attempting to construct a pathwise solution theory on random variables.
Instead, the operator $\c{L}$ is thought of as a map between the associated Hilbert spaces: if one or both are function spaces admitting reproducing kernels, the generalized Gaussian fields respectively yield Gaussian processes in the classical sense.
In the given setting, this solution concept ends up being useful, thanks to the following general result.

\begin{result}[Solution of a stochastic partial differential equation]
If $\c{L}$ is invertible, then
\[
f(\omega,h) = \c{W}(\omega, \c{L}^{-*} h)
\]
is the unique solution to $\c{L}f = \c{W}$.
\end{result}

\begin{proof}
\textcite[Theorem 4.2.2.]{lototsky17}.
\end{proof}

It is remarkable that such a minimalist solution theory, which relies very fundamentally on the fact that Gaussian processes are uniquely determined by their associated reproducing kernel Hilbert spaces or generalizations thereof, gives a description concrete enough for our purposes.
Indeed, this result generally determines but does not reveal what function space $f$ lies in as a random variable, along with its regularity properties such as continuity---however, for Bayesian learning, we don't actually need these.

For this construction to yield a Gaussian process in the classical sense, we should ensure the space $f$ lies in is a reproducing kernel Hilbert space, since this lets us build the Gaussian processes of interest from pointwise evaluation functionals as described previously.
The main challenge, then, is finding appropriate spaces and operators in order to apply the preceding ideas.

\subsection{The Riemannian Matérn kernel}

We now use the above results to define Riemannian Gaussian processes directly, and, following this, calculate their kernels in order to obtain workable numerical expressions.
To proceed, we need to make concrete choices for which Hilbert spaces and operators to use.
To do so, we study the class of equations considered by \textcite{whittle54,whittle63,lindgren11}, which we now review.

Suppose temporarily that $X = \R^d$ is Euclidean.
In that setting, \textcite{whittle54,whittle63} has shown that, if we suppose that $f$ is, in a purely formal sense, a solution to the stochastic partial differential equation
\[
\del{\frac{2\nu}{\kappa^2} - \lap}^{\frac{\nu}{2} + \frac{d}{4}} f = \c{W}
\]
then its covariance kernel must be the \emphmarginnote{Matérn kernel}
\[
k_{\nu}(x,x') = \sigma^2 \frac{2^{1-\nu}}{\Gamma(\nu)} \del{\sqrt{2\nu} \frac{\norm{x-x'}}{\kappa}}^\nu K_\nu \del{\sqrt{2\nu} \frac{\norm{x-x'}}{\kappa}}
\]
where $\Gamma$ is the gamma function and $K_\nu$ is the modified Bessel function of the second kind \cite{gradshteyn14}.
This kernel is very well-studied and among the most widely-used Euclidean kernels in practice.
Building on this idea, \textcite{lindgren11} proposed a formalism based on Galerkin finite element analysis for giving precise meaning to such calculations, including treatment of boundary conditions and other considerations.

\begin{figure}
\begin{subfigure}{0.49\textwidth}
\includegraphics{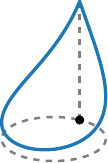}
\end{subfigure}
\begin{subfigure}{0.49\textwidth}
\includegraphics[scale=0.25]{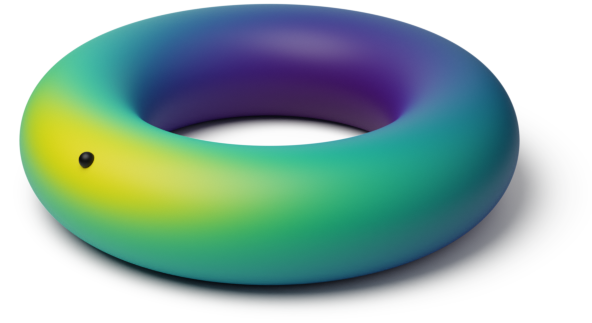}
\end{subfigure}
\caption[Matérn kernel: circle and torus]{The Matérn-1/2 kernel $k_{1/2}(x,\.)$ defined on a circle and torus, where the point $x$ is marked by a black dot. The kernel on the circle is computed using the Poisson summation formula, which relates Laplacian eigenvalues and eigenfunctions on the circle with those on the real line, and therefore also relates kernels on the circle with kernels on the real line.}
\label{fig:ker-s1-t2}
\end{figure}

\begin{figure}
\begin{subfigure}{0.49\textwidth}
\includegraphics[scale=0.25]{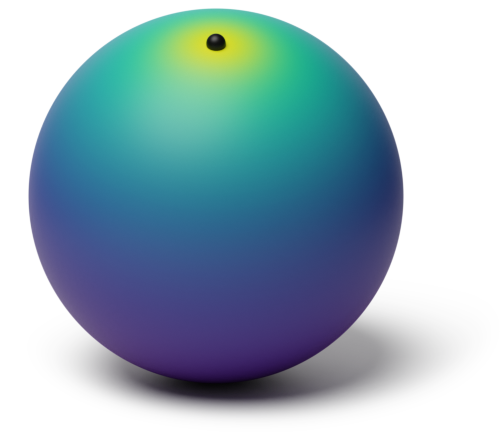}
\end{subfigure}
\begin{subfigure}{0.49\textwidth}
\includegraphics[scale=0.25]{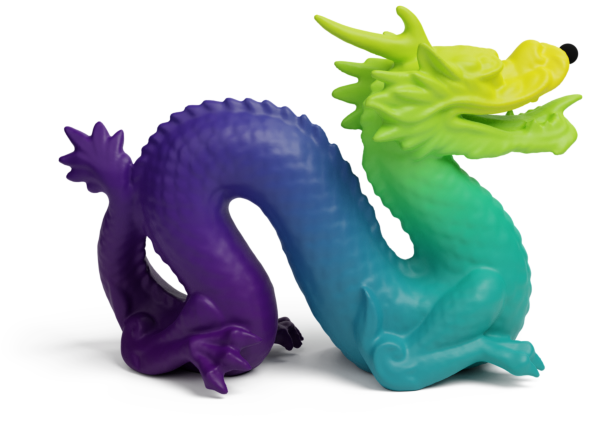}
\end{subfigure}
\caption[Matérn kernel: sphere and dragon manifold]{The Matérn-1/2 kernel $k_{1/2}(x,\.)$ defined on a sphere and dragon manifold, where the point $x$ is marked by a black dot.}
\label{fig:ker-s2-dr}
\end{figure}

In particular, \textcite{lindgren11} define the \emphmarginnote{Riemannian Matérn kernel} to be the covariance kernel of the solutions of analogs of the above stochastic partial differential equation, where $\R^d$ is replaced with a Riemannian manifold $X$, and the Euclidean Laplacian $\lap$ is replaced by the Laplace--Beltrami operator $\lap_g$.
The resulting kernel, shown in \Cref{fig:ker-s1-t2,fig:ker-s2-dr}, is well-defined, but implicit: \textcite{lindgren11} provide techniques for training the resulting processes by solving the stochastic partial differential equation numerically.

We develop an alternative, more constructive formalism.
This is done by improving on the above prior results in two ways: (1) we bypass finite element analysis by instead working with the previously-introduced theory of Gaussian stochastic partial differential equations described by \textcite{lototsky17}, and (2) we deduce numerical expressions for calculating the kernels of the resulting processes, enabling them to be trained using standard methods.
To start, we define the right-hand-side of our equations.

\label{ntn:riemannian-white-noise}
\begin{definition}[Riemannian white noise]
Define the \emph{white noise process} $\c{W}_g : \Omega \x L^2(X) \-> \R$ to be a centered generalized Gaussian field with identity covariance operator $\id : L^2(X) \-> L^2(X)$, where we recall that the inner product on $L^2(X)$ is defined by integration against the Riemannian volume measure.
By applying Kolmogorov's Extension Theorem to a family of finite-dimensional marginals indexed by $L^2(X)$, we conclude such a process exists and is well-defined.
\end{definition}

This stochastic process \emph{cannot} be viewed as a scalar-valued random function.
Though Kolmogorov's Extension Theorem does imply there is a random variable $\c{W}_g : \Omega \-> \R^{L^2(X)}$ whose finite-dimensional marginals coincide with $\c{W}_g : \Omega \x L^2(X) \-> \R$, this is next-to-useless because $\R^{L^2(X)}$ is highly irregular and we can say little about which subspace $\c{W}_g$ concentrates on without additional considerations that we may conveniently avoid.

Next, we define the left-hand-side of the stochastic partial differential equations under study, including the Hilbert spaces and operators $\c{L}$ of interest using functional calculus.
For this, a result on Riemannian reproducing kernel Hilbert spaces will be of key interest.

\begin{result}[Riemannian Sobolev and diffusion spaces]
\label{res:riemannian-sobolev}
Define the Riemannian Sobolev space $H^s(X)$ by
\[
H^s(X) = \cbr{f \in D'(X) : f = (1 - \lap_g)^{-s/2} h : h \in L^2(X)}
\]
and the Riemannian diffusion space $\c{H}^s(X)$ by
\[
\c{H}^s(X) = \cbr{f \in D'(X) : f = e^{\frac{s}{2}\lap_g} h : h \in L^2(X)}
\]
where the operators are defined using functional calculus.
Then $H^s(X)$ with $s > \frac{d}{4}$ and $\c{H}^s(X)$ with $s > 0$ are reproducing kernel Hilbert spaces.
\end{result}

\begin{proof}
\textcite[Theorem 3 and Theorem 6]{devito20}.
\end{proof}

This gives the key technical pillar upon which our calculations rest.
It both provides appropriate Hilbert spaces to use within the solution theory, and, by virtue of admitting reproducing kernels, guarantees that they are spaces of actual functions $f : X \-> \R$ on the Riemannian manifold.
This enables us to plug pointwise evaluation functionals into the obtained generalized Gaussian field, yielding Riemannian Gaussian processes in the classical sense---the objects we sought to construct in the first place.

\parmarginnote{Riemannian squared exponential kernel}
Note that the operators $e^{\frac{s}{2}\lap_g}$ can be viewed as limits of appropriately rescaled versions of the operators $(1 - \lap_g)^{-s/2}$. 
Given that the Euclidean Matérn kernel converges to the Euclidean squared exponential kernel as $\nu\->\infty$, we therefore view stochastic partial differential equations induced by $e^{\frac{s}{2}\lap_g}$ as giving rise to the \emph{Riemannian squared exponential kernel}---we make this perspective precise shortly.

\parmarginnote{Generalized spectral measure}
Our strategy will be to represent the kernels of interest as infinite sums of Laplace--Beltrami eigenfunctions.
Here, compactness of $X$ is key: this ensures that the total number of eigenfunctions is countable, so that summing eigenfunctions makes sense.
The coefficients in the sum can be viewed as a kind of \emph{generalized spectral measure}, which, in our case, owing to compactness, is supported on the non-negative integers rather than on $\R^d$.
This shows one way in which geometry of spaces is reflected in properties of kernels.

To carry the necessary calculations out and compute the kernels of the Gaussian processes defined by our stochastic partial differential equations, we need to relate the Sobolev and diffusion spaces of \textcite{devito20} with the equations studied by \textcite{whittle54,whittle63,lindgren11}.
This gives series expansions of the kernels in terms of Laplace--Beltrami eigenpairs.
Convergence rates of these series depend on the eigenvalue growth rate, which is quantified by Weyl's law \cite{zelditch17}.
We now state and prove the main result.

\begin{theorem}[Riemannian Matérn and squared exponential kernels]
Let $X$ be a compact Riemannian manifold.
For $\nu > 0$ and $\kappa > 0$, define the stochastic partial differential equations 
\[
\del{\frac{2\nu}{\kappa^2} - \lap_g}^{\frac{\nu}{2} + \frac{d}{4}} f &= \c{W}_g
&
e^{-\frac{\kappa^2}{4}\lap_g} f &= \c{W}_g
\]
where, respectively, we have
\[
\del{\frac{2\nu}{\kappa^2} - \lap_g}^{\frac{\nu}{2} + \frac{d}{4}} : H^{\nu + \frac{d}{2}}(X) \-> L^2(X)
\\
e^{-\frac{\kappa^2}{4}\lap_g} : \c{H}^{\frac{\kappa^2}{2}}(X) \-> L^2(X)
.
\]
Then, letting $(\lambda_n,f_n)$ be the eigenvalues and eigenfunctions of the Laplace--Beltrami operator, in both cases the unique solutions $f$ are Gaussian processes with absolutely convergent respective covariance kernels
\[
k(x, x') &= \sum_{n=0}^\infty \del{\frac{2\nu}{\kappa^2} + \lambda_n}^{-\nu-\frac{d}{2}} f_n(x)f_n(x')
\\
k(x, x') &= \sum_{n=0}^\infty e^{-\frac{\kappa^2}{2} \lambda_n} f_n(x)f_n(x')
\]
which we call the \emph{Riemannian Matérn} and \emph{Riemannian squared exponential} kernels.
\end{theorem}

\begin{proof}
Note first that the operators corresponding to the Matérn kernel coincide with those used by \textcite{devito20} in defining the Sobolev spaces of interest if we have a fixed length scale given by $\kappa = \sqrt{2\nu}$.
Similarly, the operators corresponding to the squared exponential kernel always coincide.
In this setting, the reproducing kernels are given by \textcite[Proposition 2]{devito20} as
\[
k(x, x') &= \sum_{n=0}^\infty \del{\frac{2\nu}{\kappa^2} + \lambda_n}^{-\nu-\frac{d}{2}} f_n(x)f_n(x')
\\
k(x, x') &= \sum_{n=0}^\infty e^{-\frac{\kappa^2}{2} \lambda_n} f_n(x)f_n(x')
\]
which are shown to be absolutely convergent in that work.
This proves the claim for the squared exponential case.
However, we are interested in general length scales $\kappa > 0$, so this does not suffice for the Matérn case.
To extend this, the idea will be to let $\tilde{g} = \frac{2\nu}{\kappa^2}g$ be a rescaled metric tensor on $X$, and define the equations
\[
\del{\frac{2\nu}{\kappa^2} - \lap_g}^{\frac{\nu}{2} + \frac{d}{4}} f &= \c{W}_g
&
\del{1 - \lap_{\tilde{g}}}^{\frac{\nu}{2} + \frac{d}{4}} \tilde{f} &= \c{W}_{\tilde{g}}
\]
for which we would like to show that $f = \del{\frac{\kappa^2}{2\nu}}^{\frac{\nu}{2} + \frac{d}{2}} \tilde{f}$.
To begin, we first prove these operators are well-defined, by checking that they are bounded and invertible for all positive $\nu$ and $\kappa$.
By \Cref{res:riemannian-sobolev} we know that for every $f \in H^{\nu + \frac{d}{2}}(X)$, there is an $h \in L^2(X)$ such that $f = (1-\lap_g)^{-\frac{\nu}{2}-\frac{d}{4}} h$.
Moreover, both $f$ and $h$ can be expressed in the orthonormal basis given by Laplacian eigenfunctions $f_n$ as 
\[
h(x) &= \sum_{n=0}^\infty \alpha_n f_n
&
f(x) &= \sum_{n=0}^\infty \del{\frac{1}{1+\lambda_n}}^{\frac{\nu}{2}+\frac{d}{4}} \alpha_n f_n
\]
where the expression for $f$ follows by applying the operator $(1-\lap_g)^{-\frac{\nu}{2}-\frac{d}{4}}$ to the eigenfunctions.
Finally, note that since $\lambda_n \geq 0$ we have
\[
\min\del{\frac{2\nu}{\kappa^2}, 1} \leq \frac{\frac{2\nu}{\kappa^2} + \lambda_n}{1+\lambda_n} \leq \max\del{1, \frac{2\nu}{\kappa^2}}
.
\]
Using these identities, write
\[
\norm{\del{\frac{2\nu}{\kappa^2} - \lap_g}^{\frac{\nu}{2} + \frac{d}{4}} f}^2_{L^2(X)} &= \norm{\sum_{n=0}^\infty \del{\frac{\frac{2\nu}{\kappa^2} + \lambda_n}{1+\lambda_n}}^{\frac{\nu}{2}+\frac{d}{4}} \alpha_n f_n}^2_{L^2(X)}
\\
&\leq \sum_{n=0}^\infty \max\del{1, \frac{2\nu}{\kappa^2}}^{\nu + \frac{d}{2}} \alpha_n^2
\\
&= \max\del{1, \frac{2\nu}{\kappa^2}}^{\nu + \frac{d}{2}} \norm{f}^2_{H^{\nu + \frac{d}{2}}(X)}
\]
and similarly 
\[
\norm{\del{\frac{2\nu}{\kappa^2} - \lap_g}^{\frac{\nu}{2} + \frac{d}{4}} f}^2_{L^2(X)} &\geq \sum_{n=0}^\infty \min\del{\frac{2\nu}{\kappa^2},1}^{\nu + \frac{d}{2}} \alpha_n^2
\\
&= \min\del{\frac{2\nu}{\kappa^2},1}^{\nu + \frac{d}{2}} \norm{f}^2_{H^{\nu + \frac{d}{2}}(X)}  
\]
where we have also used $L^2(X)$-orthonormality of $f_n$ as well as
\[
\sum_{n=0}^\infty \alpha_n^2 = \norm{h}^2_{L^2(X)} = \norm{f}^2_{H^{\nu + \frac{d}{2}}(X)}
\]
which follows from orthonormality of $f_n$ along with the definition of $h$ and \Cref{res:riemannian-sobolev}.
This proves boundedness and invertibility.
To complete the argument, we check that the desired identity holds under a change of metric.
The change of metric expressions of interest are given by 
\[
\lap_{\tilde{g}} &= \frac{\kappa^2}{2\nu} \lap_g
&
\tilde{\lambda}_n &= \frac{\kappa^2}{2\nu} \lambda_n
&
\tilde{f}_n &= \del{\frac{2\nu}{\kappa^2}}^{-\frac{d}{4}} f_n
&
\vol_{\tilde{g}} &= \del{\frac{2\nu}{\kappa^2}}^{\frac{d}{2}} \vol_g
\]
thus starting from $(1 - \lap_{\tilde{g}})^{\frac{\nu}{2} + \frac{d}{4}} \tilde{f} = \c{W}_{\tilde{g}}$ we can write
\[
(1 - \lap_{\tilde{g}})^{\frac{\nu}{2} + \frac{d}{4}} = \del[3]{1 - \frac{\kappa^2}{2\nu} \lap_g}^{\frac{\nu}{2} + \frac{d}{4}} = \del[3]{\frac{\kappa^2}{2\nu}}^{\frac{\nu}{2} + \frac{d}{4}}  \del{\frac{2\nu}{\kappa^2} - \lap_g}^{\frac{\nu}{2} + \frac{d}{4}} 
\]
as well as 
\[
\c{W}_{\tilde{g}} = \del{\frac{2\nu}{\kappa^2}}^{\frac{d}{4}} \c{W}_g
\]
which when combined gives $f = \del{\frac{\kappa^2}{2\nu}}^{\frac{\nu}{2} + \frac{d}{2}} \tilde{f}$.
On the other hand, under a change of metric, the series representation of the Matérn kernel is
\[
\tilde{k}(x,x') &= \sum_{n=0}^\infty (1 + \tilde{\lambda}_n)^{-\nu - \frac{d}{2}} \tilde{f}_n(x) \tilde{f}_n(x')
\\
&= \sum_{n=0}^\infty \del[3]{1 + \frac{\kappa^2}{2\nu} \lambda_n}^{-\nu - \frac{d}{2}} \del{\frac{2\nu}{\kappa^2}}^{-\frac{d}{2}} f_n(x) f_n(x')
\\
&= \del[3]{\frac{\kappa^2}{2\nu}}^{-\nu - d} \sum_{n=0}^\infty \del{\frac{2\nu}{\kappa^2} + \lambda_n}^{-\nu - \frac{d}{2}} f_n(x) f_n(x')
\]
where $\tilde{k}$ is the transformed kernel, for which we see that $\del{\frac{\kappa^2}{2\nu}}^{\nu + d} \tilde{k}(x,x') = k(x,x')$.
But this exactly matches the covariance kernel of $f$ according to the given transformation of the Gaussian process, and the claim follows.
\end{proof}

\subsection{Illustrated examples}

Using the developed tools, we have defined Riemannian Gaussian processes of interest as solutions of stochastic partial differential equations, and derived expressions for their kernels which are explicit enough to be amenable to approximation. 
Here, we explore using these processes as priors in Bayesian learning.
Our goal is to understand how complicated geometry affects posterior uncertainty estimates.

We focus on the dragon manifold from the Stanford 3D scanning repository \cite{curless96}, approximated numerically as a mesh.
This mesh was chosen because it has no boundary and is geometrically more complex than other alternatives.
We work with the largest connected component of the mesh, which has a total of $100\,179$ vertices and $201\,010$ triangular faces.

\begin{figure}
\begin{subfigure}{0.49\textwidth}
\includegraphics[scale=0.25]{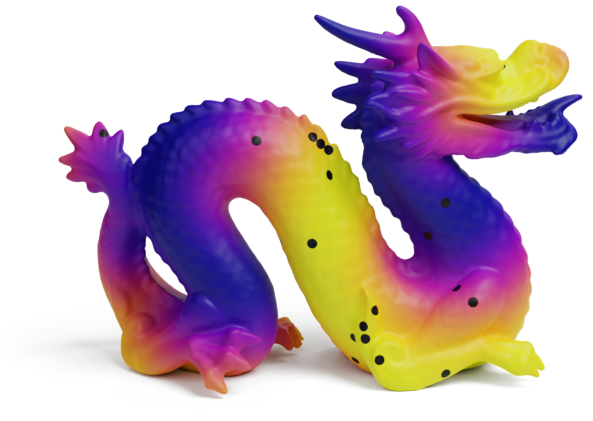}
\caption{Ground truth}
\end{subfigure}
\begin{subfigure}{0.49\textwidth}
\includegraphics[scale=0.25]{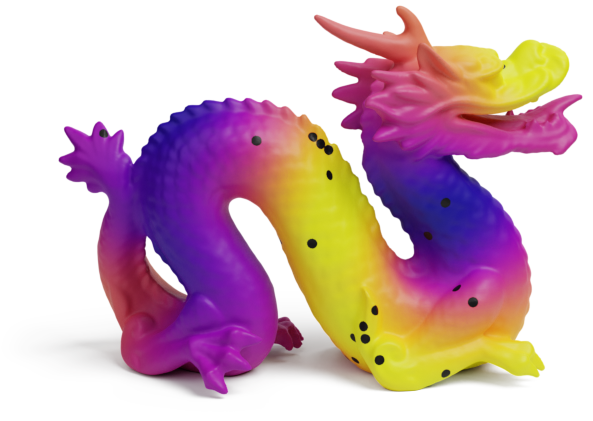}
\caption{Mean}
\end{subfigure}
\begin{subfigure}{0.49\textwidth}
\includegraphics[scale=0.25]{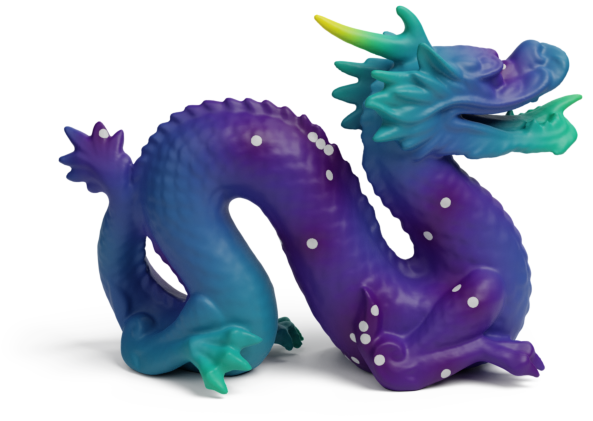}
\caption{Standard deviation}
\end{subfigure}
\begin{subfigure}{0.49\textwidth}
\includegraphics[scale=0.25]{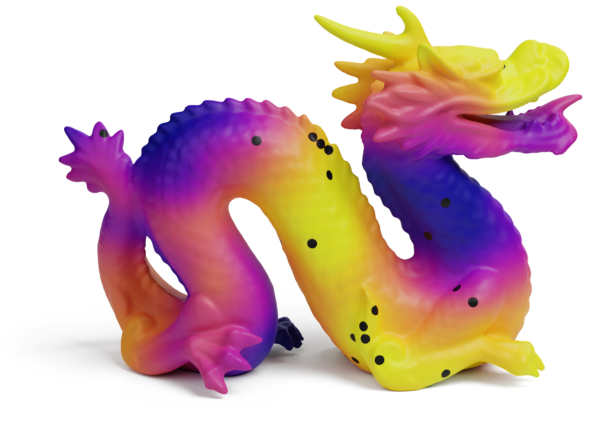}
\caption{One posterior sample}
\end{subfigure}
\caption[Posterior Gaussian process: dragon manifold]{A posterior Matérn Gaussian process on the dragon manifold. We plot true function values, along with the posterior mean and standard deviation. Here, the black dots represent data. We observe that the standard deviation generally increases as we move away from the training locations.}
\label{fig:dr-posterior}
\end{figure}

On the mesh, discrete analogs of all of the differential-geometric notions we require are available, with reasonably well-understood approximation properties \cite{crane17}.
We obtain $500$ discretized Laplace--Beltrami eigenpairs numerically in finite element space using the \emph{Firedrake} partial differential equation package \cite{rathgeber16}.
This, in turn, gives expressions for evaluating the kernel and sampling the prior at any points on the manifold.

To generate training data, we define a ground truth function given by the sine of the geodesic distance from a distinguished point, chosen to be the first vertex on the mesh, which is located at the end of the dragon's snout.
We observe this function at a total of $52$ points chosen from the mesh's vertices.
We train a Matérn prior with smoothness $\nu = 3/2$ on this data using gradient descent, obtaining learned variance and length scale hyperparameters.
Finally, we obtain the posterior samples over the entire mesh using pathwise sampling.

Results are given in \Cref{fig:dr-posterior}.
We observe that the both the mean predictions and uncertainty estimates adapt to the manifold's geometry.
In particular, we see that uncertainty increases when moving away from the data.
We see that the two upper and lower parts of the dragon's snout have different uncertainty estimates, in spite of the fact that they are close in ambient Euclidean distance.
Overall, we conclude that the geometric Matérn models used produce uncertainty estimates which reflect the manifold's geometry.

From the computational tools used, one can see that there is more to say about the developed model class.
In particular, the discrete analogs of the Laplace--Beltrami operator used within the finite element computations give rise to discrete Gaussian processes in their own right.
This leads one to ask: are there interesting analogs of Matérn models in useful classes of discrete spaces?
We now proceed to develop one such analog.

\section{Graph Matérn Gaussian processes}

A general maxim of geometry is that anything one can do with a manifold, one can do with a graph, given some thinking---but, the details and behavior might be a bit different.
In the preceding section, we built Gaussian processes using spectral properties of the Riemannian Laplace--Beltrami operator.
We now ask: can we repeat these constructions on graphs, obtaining Gaussian processes that can be used in very different settings?

\subsection{Review of graph theory}
Graphs are discrete spaces which formalize and describe interconnected networks of points.
For an introduction to the aspects of graph theory of key interest for our purposes, see \textcite{spielman12}.
Here we briefly review the key notions.

\parmarginnote{Nodes, weights, edges}
A weighted undirected graph consists of a finite set of \emph{nodes}, each of which is assigned a positive real number called the \emph{weight}, and \emph{edges} between nodes.
Graphs with unit weights can be visualized by drawing a set of points on a two-dimensional plane or in three-dimensional space, representing the nodes, and drawing lines between the points, representing the edges.

\parmarginnote{Graphs and matrices}
A broad theme in graph theory is that geometric properties of graphs can be encoded as finite-dimensional vectors and matrices by picking an arbitrary ordering of nodes, and associating nodes with rows and columns of a matrix.
Define the \emph{weighted }\emphmarginnote{adjacency matrix} $\m{W}$ by letting the matrix entry corresponding to two nodes be their respective edge weight.
Similarly, define the diagonal \emphmarginnote{degree matrix} $\m{D}$ by $D_{ii} = \sum_j W_{ij}$---for graphs with unit weights, this counts how many neighbors each node has.

We can construct \emphmarginnote{functions on graphs}, which map each node into some space of interest, in a number of ways.
For defining functions which depend only on neighboring nodes, working with matrices induced by the graph is often useful.\marginnote{Permutation invariance and equivariance}
When doing so, we must take care that the functions constructed do not depend on the arbitrary choice of ordering used to associate nodes with matrix rows and columns. Algebraically, this is encoded through \emph{permutation invariance}, \emph{permutation equivariance}, and other similar requirements.

\parmarginnote{Discretizations of manifolds}
Graphs often arise as discretizations of manifolds: there are many ways of making this idea precise.
For example, one can consider an infinite sequence of nested finite subsets of the manifold converging monotonically to a countable dense subset thereof.
This defines a sequence of graphs by taking nodes to be the elements of each sets, and connecting neighboring nodes.
Sufficiently well-behaved such sequences give rise to vector spaces of functions converging monotonically to an appropriate function space on the manifold.

\subsection{The graph Laplacian}

In the Riemannian setting, our strategy for building Gaussian process models essentially amounted to using functional calculus defined using spectral properties of the Laplace--Beltrami operator to construct the Gaussian processes of interest as solutions of stochastic partial differential equations.
This strategy depended on the manifold only through the Laplace--Beltrami operator, and the white noise process, which suggests that it may also work for other classes of spaces.

\label{ntn:graph-laplacian}
Weighted directed graphs in particular admit the notion of a \emphmarginnote{graph Laplacian}, which is a symmetric positive semi-definite matrix defined as 
\[
\m\lap = \m{D} - \m{W}
\]
where $\m{D}$ is the degree matrix and $\m{W}$ is the weighted adjacency matrix.
The graph Laplacian can be interpreted as a linear operator acting the space of all functions $f : G \-> \R$ where $G$ is the set of nodes.
Each such function can be represented as a $|G|$-dimensional vector $\v{f}$ by assigning an ordering to the nodes, as described previously.
From this viewpoint, the graph Laplacian is
\[
(\m\lap\v{f})(x) = \sum_{x'\~x} f(x) - f(x')
\]
where $x$ is a node, and sum is taken over all neighboring nodes $x'$ of $x$.
This expression now resembles its Euclidean and Riemannian analogs in a much more direct manner, and justifies why one would call $\m\lap$ a Laplacian in the first place.
First, though, we clarify the choice of sign in this expression.

\begin{remark}[Sign convention]
Following standard practice in graph theory, we adopt a \emph{different sign convention} for the graph Laplacian compared to its Euclidean and Riemannian analogs.
One should thus view the operators 
\[
&\ubr{\phantom{-}\m\lap}_{\t{no minus sign}}
&
&\ubr{-\lap}_{\t{minus sign}}
\]
as analogues of one-another---in particular, both are positive semi-definite.
\emph{Note the different minus signs!}
This corresponds to adopting the analyst's rather than geometer's convention for studying $\lap$.
\end{remark}

\label{ntn:matrix-functional-calculus}
In the graph case, we can also develop a notion of \emphmarginnote{functional calculus} just as we developed previously.
This time, however, doing so is mathematically more-or-less trivial: for a function $\Phi : \R \-> \R$ and a diagonal matrix $\m\Lambda$, let $\Phi(\m\Lambda)$ be the matrix obtained by applying $\Phi$ to the diagonal.
Then, letting $\m\lap = \m{U}\m\Lambda\m{U}^T$ be the eigenvalue decomposition of $\m\lap$, which by positive semi-definiteness always exists, define
\[
\Phi(\m\lap) = \m{U}\Phi(\m\Lambda)\m{U}^T
.
\]
This gives a notion of functional calculus analogous to the one considered previously in the Riemannian case, only without the need to introduce any mathematical theory beyond elementary eigenvalue factorizations.
In particular, since all matrices are finite, we do not need to consider any kind of convergence.
We now examine Gaussian processes from this perspective.

\subsection{The graph Matérn kernel}

\label{ntn:graph-white-noise}
Our goal now is to construct Gaussian processes $f : G \-> \R$ where $G$ is the set of nodes of a weighted undirected graph.
We'd like to define processes whose covariance reflects the structure of the graph.
To do this, we adapt the notions of Matérn and squared exponential Gaussian processes studied previously to the graph setting.

Recall that in the Euclidean and Riemannian cases, these processes were defined as solutions of stochastic partial differential equations with left-hand-sides defined using functional calculus, and right-hand-sides consisting of white noise processes.
Adapting this definition by replacing $-\lap$ with $\m\lap$ and dropping dimension-dependent terms from exponents gives 
\[
\del{\frac{2 \nu}{\kappa^2} + \m\Delta}^\frac{\nu}{2} \v{f}(\omega) &= \bc{W}(\omega)
&
e^{\frac{\kappa^2}{4} \m\Delta} \v{f}(\omega) &= \bc{W}(\omega)
\]
where $\bc{W} \~[N](\v{0},\m{I})$ are standard Gaussians, and $\v{f} : \Omega \-> \R^{|G|}$ are the random vectors defining the stochastic processes $f : \Omega \x G \-> \R$ defined on the graph's nodes.
Note that the numbers $\frac{2 \nu}{\kappa^2}$ are \emph{not} added element-wise to the graph Laplacian---instead, following the established conventions, they are added to its \emph{eigenvalues}.
By elementary algebra, the \emphmarginnote{graph Matérn Gaussian processes} and \emph{graph squared exponential Gaussian processes} are given by
\[
\v{f} &\~[N]\del{\v{0}, \del{{\textstyle\frac{2 \nu}{\kappa^2}} + \m\Delta}^{-\nu}}
&
\v{f} &\~[N]\del{\v{0}, e^{-\frac{\kappa^2}{2} \m\Delta}}
\]
This defines the graph Matérn and squared Gaussian processes of interest.
We now explore some of their properties.

By virtue of its definition, $\m\lap$ inherits \emphmarginnote{sparsity} properties from graphs.
Thus, for sufficiently small integers $\nu$ and many graphs, the matrices $(\frac{2 \nu}{\kappa^2} + \m\Delta)^{\frac{\nu}{2}}$ will be sparse.
If the precise sparsity pattern is sufficiently well-behaved---for instance in certain planar graphs---these matrices' Cholesky factors will also be sparse, potentially reducing computational costs from cubic to linear or close to it.
Alternatively, Krylov subspace methods for sparse linear systems can be applied, giving another way to leverage sparsity to improve scalability.

\begin{figure}
\begin{subfigure}{0.05\textwidth}
\includegraphics{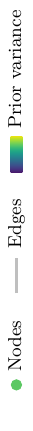}
\end{subfigure}
\begin{subfigure}{0.94\textwidth}
\begin{subfigure}{0.49\textwidth}
\includegraphics{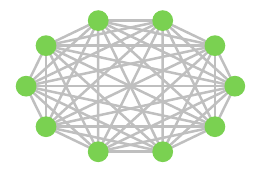}
\caption{Complete graph}
\end{subfigure}
\begin{subfigure}{0.49\textwidth}
\includegraphics{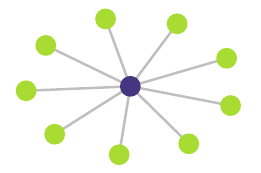}
\caption{Star graph}
\end{subfigure}
\begin{subfigure}{0.49\textwidth}
\includegraphics{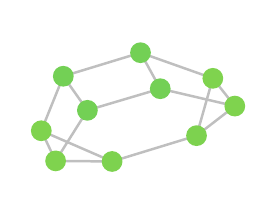}
\caption{Regular random graph}
\end{subfigure}
\begin{subfigure}{0.49\textwidth}
\includegraphics{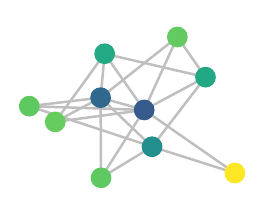}
\caption{Barabási--Albert random graph}
\end{subfigure}
\end{subfigure}
\caption[Graph Matérn kernels]{Here, we show the prior variance of a graph Gaussian process at each node for a number of different graphs. Variances for all graphs are plotted using a common scale. In general, this prior variance is non-uniform and instead reflects the structure of the graph.}
\label{fig:graph-variance}
\end{figure}

Graph Matérn and graph squared exponential kernels possess \emphmarginnote{non-uniform variance}, whose precise form will vary with the graph.
In particular, each node's variance is not simply a function of its degree, and instead depends on the precise geometry in a complex manner.
This can be seen in \Cref{fig:graph-variance}.
Similar phenomena occur in random walk kernels studied by \textcite{urry13}---in that setting, variance is determined by the return time of a certain random walk defined on the graph.

\parmarginnote{Symmetric normalized graph Laplacian}
It's also possible to use the \emph{symmetric normalized graph Laplacian}, which is defined as $\m{D}^{-1/2}\m\lap\m{D}^{-1/2}$, to define analogous kernels to the ones above, by using this matrix in place of the graph Laplacian.
This yields the \emph{symmetric normalized graph Matérn} and \emph{symmetric normalized graph squared exponential} Gaussian processes.
These can be preferable in domains where symmetric normalized Laplacians are customarily used.

\parmarginnote{Connection with random walks}
The graph squared exponential kernel can be connected with \emph{random walks} in a number of ways.
Firstly, it is the Green's function of the graph diffusion equation.
More precisely, if $\v\phi : [0,\infty) \x G \-> \R$ solves the equation 
\[
\v{\dot\phi}_t + \m\lap\v\phi_t &= 0
&
\v\phi_0 &= \v{v}
\]
then $\v\phi(\tau,\.) = e^{-\tau\m\lap}\v{v}$, where our notation again uses the equivalence between vectors and real-valued functions on $G$.
This equation has a strong physical interpretation: it describes heat transfer along the graph---this gives a way to understand what kind of prior information graph squared exponential kernel introduces.
Similarly, if $\m\lap$ is replaced with the symmetric normalized graph Laplacian, then $\v\phi(\tau,\.)$ can be interpreted as the unnormalized probability density of a continuous-time random walk moving along the graph.

Another way to understand the prior information contained in the given kernels is through limits.
Mirroring all other settings considered, graph Matérn kernels converge to graph squared exponential kernels as $\nu\->\infty$.
This is essentially immediate, since their eigenvectors coincide, and eigenvalues converge.
Graph squared exponential kernels also arise as a limit of the \emph{random walk kernel} of \textcite{smola03}, defined using the degree matrix $\m{D}$ as
\[
(\m{I} - (1-\alpha)\m{D}^{-1/2}\m\lap\m{D}^{-1/2})^s
.
\]
This kernel arises from symmetrizing the $s$-step transition matrix of a certain lazy random walk on a graph.
Though it looks similar to the graph Matérn kernel, its structure is very different: $s > 0$ is positive rather than negative, $\alpha\in[0,1)$ is interpreted as the laziness parameter of the underlying random walk, and the Laplacian is subtracted rather than added.
Still, this kernel converges to the graph squared exponential kernel: if we set $\alpha = 1 - \frac{\kappa^2}{2k}$ with $\kappa$ a fixed constant, we have 
\[
\lim_{s\->\infty} (\m{I} - (1-\alpha)\m{D}^{-1/2}\m\lap\m{D}^{-1/2})^s = e^{-\frac{\kappa^2}{2} \m{D}^{-1/2}\m\lap\m{D}^{-1/2}}
.
\]
This provides another view of the connection between the graph squared exponential kernel and random walk models.

\begin{figure}
\tikzset{external/export next=false}
\begin{tikzpicture}
\node at (0,0) {\includegraphics[scale=0.25]{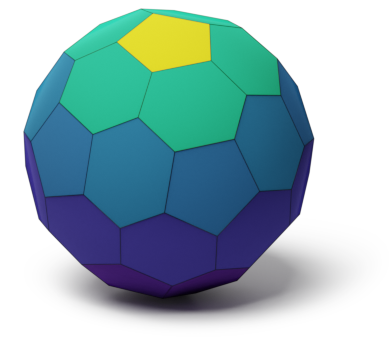}};
\node at (3,0) {\includegraphics[scale=0.25]{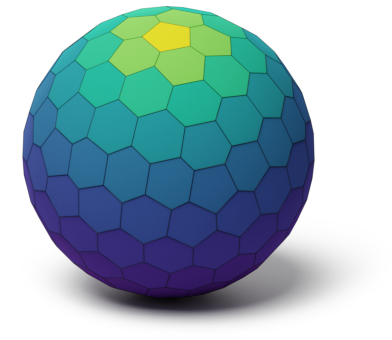}};
\node at (6,0) {\includegraphics[scale=0.25]{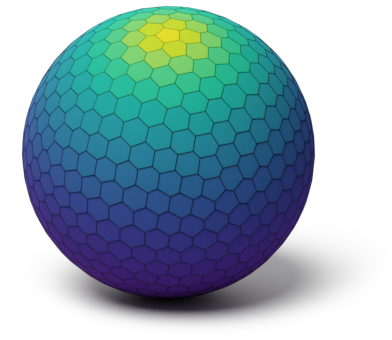}};
\node at (9,0) {\includegraphics[scale=0.25]{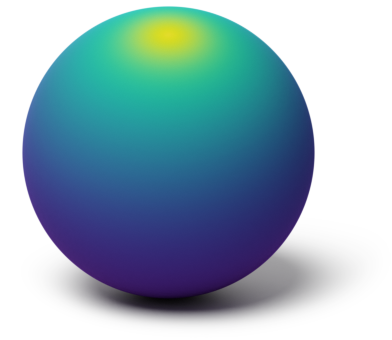}};
\end{tikzpicture}
\caption[Limits of graph Matérn kernels]{Illustration of a sequence of graph Matérn-1/2 kernels, each defined on an icosahedral graph. For every icosahedron, such a graph is defined by letting each face corresponds to a node on the graph, and neighboring faces correspond to edges on the graph. The limiting kernel for this sequence of graphs is a Riemannian Matérn-1/2 kernel on the sphere.}
\label{fig:graph-limit}
\end{figure}

Finally, the introduced graph kernels also converge to \emphmarginnote{Riemannian limits}, provided the graphs are sufficiently geometrically well-behaved and the limits are understood appropriately.
In cases studied by \textcite{belkin07,burago14}, the eigenvalues and eigenvectors of the graph Laplacian converge to those of the Laplace--Beltrami operator.
Using this, \textcite{sanzalonso21} provide a framework for studying limits of graph Matérn kernels.
One example of such a limit is shown in \Cref{fig:graph-limit}.

An alternative way to study Riemannian limits is to embed the graph within a Euclidean space, and study vector spaces of piecewise-linear functions between neighboring nodes.
Sequences of such graphs can arise as \emph{finite element} discretizations of function spaces, which were considered previously in \Cref{ch:pathwise}.
Here, \textcite{lindgren11} show that certain graph Matérn Gaussian processes converge to their Riemannian limits.

In total, these results illustrate that graph Matérn and graph squared exponential kernels are closely-connected with their Riemannian analogs, which justifies both the names they are given, and the choice of defining them using the graph Laplacian in the first place.
In spite of their similarity, however, these models can be used in settings which are very different from the manifold setting that inspired them.
We now illustrate a few possibilities.

\subsection{Illustrated examples}

To illustrate the graph Matérn Gaussian processes, we demonstrate their use in a setting which departs considerably from the Euclidean and Riemannian settings considered before.
Our goal is to show how such models may enable applications that are very different from those in which Gaussian processes have been traditionally used.

\begin{figure}[p!]
\begin{subfigure}{0.49\textwidth}
\includegraphics{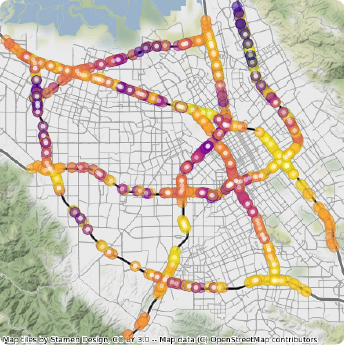}
\caption{Mean}
\end{subfigure}
\begin{subfigure}{0.49\textwidth}
\includegraphics{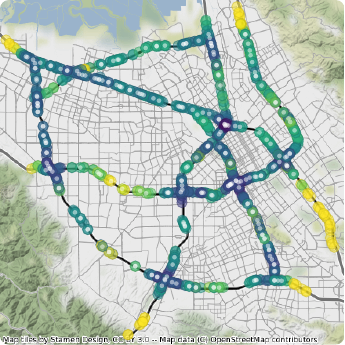}
\caption{Standard deviation}
\end{subfigure}
\caption[Posterior Gaussian process: traffic data, global view]{Posterior means and standard deviations, in miles per hour, for the probabilistic graph interpolation task on the road network. Nodes with white circles indicate sensor locations where data is available. We observe that standard deviation generally increases as we move away from the parts of the state space where there is data. Standard deviation values above $10$ are shown clipped.}
\label{fig:graph-posterior}
\bigskip
\bigskip
\bigskip
\begin{subfigure}{0.49\textwidth}
\includegraphics{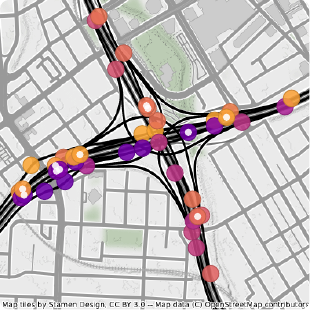}
\caption{Mean}
\end{subfigure}
\begin{subfigure}{0.49\textwidth}
\includegraphics{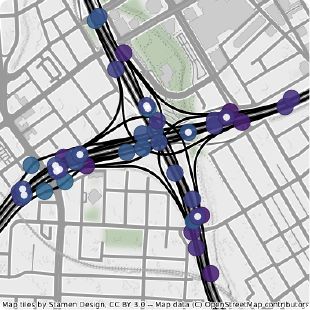}
\caption{Standard deviation}
\end{subfigure}
\caption[Posterior Gaussian process: traffic data, local view]{View of a zoomed-in portion of the posterior means and standard deviations of the probabilistic graph interpolation task on the road network. Note the small-scale variation present within the error bars.}
\label{fig:graph-posterior-zoom}
\end{figure}

To do so, we consider probabilistic interpolation of traffic data along a road network consisting of highways in the city of San Jose, California, obtained from OpenStreetMap \cite{osm17}.
In this graph, nodes are traffic sensors, and edges are roads between sensors, one for each side of the street.
Edges are weighted by inverse distance.
We work with the largest connected component of the graph, which has $1016$ nodes and $1173$ edges.

For training data, we examine traffic flow speed, which is available at $325$ nodes, using $250$ randomly chosen nodes for the training set and the remainder as the test set.
To simplify the problem and aid visualization, we focus on a single time slice consisting of traffic on Monday at 5:30pm, and do not consider space-time behavior.
We compute the kernel approximately using $500$ Laplace--Beltrami eigenpairs, and train the model by optimizing kernel hyperparameters and likelihood error variance.

Results can be seen in \Cref{fig:graph-posterior,fig:graph-posterior-zoom}.
Here, we see that in spite of the heavily non-manifold-like geometry present in this graph, the Gaussian process still reflects the graph's geometric structure.
In particular, the width of the error bars given by the posterior standard deviation increases as we move away from nodes where there is data.
We conclude that, like in the manifold case, the Gaussian process produces uncertainty estimates which reflect the graph's structure given by the connectivity of nodes.

While this example is not particularly realistic, since we have not considered temporal interpolation nor other effects likely to be important in accurate modeling of traffic, it nonetheless illustrates that modeling heavily non-Euclidean data using Gaussian processes is possible.
We hope these illustrations prompt others to imagine new and unexpected use cases for Gaussian processes.
We thus end our detour and return to the manifold setting.

\section{Gaussian vector fields on Riemannian manifolds}

In the preceding sections, we studied Gaussian processes in two different non-Euclidean settings, namely those of graphs and manifolds.
In both cases, the models obtained were scalar-valued.
We now ask: can we use these models to define vector-valued Gaussian processes on manifolds?
The first step, then, is to understand what such a notion ought to actually mean.

\subsection{Vector fields on manifolds}

A \emph{vector field} $f$ on a smooth manifold $X$ is defined as a section of the tangent bundle---we recall that this is a function $f : X \-> TX$ such that $\proj_X \after f = \id_X$.
This requirement is very intuitive: it says that for any point $x$, the tangent vector $v_x = f(x)$ must be attached to $x$.
Vector fields on manifolds exhibit certain mathematical behaviors not present in the Euclidean case---we illustrate one of the most important kinds below.

\begin{result}[Corollary of Poincaré--Hopf Theorem]
There does not exist a nowhere-vanishing continuous section on any even-dimensional sphere.
\end{result}

\begin{proof}
\textcite[Theorem 13.32]{lee10}.
\end{proof}

\begin{figure}
\tikzset{external/export next=false}
\begin{tikzpicture}
\node at (0,0) {\includegraphics[scale=0.25]{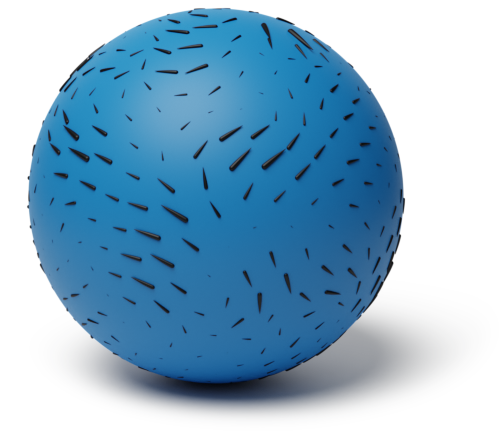}};
\node at (4,0) {\includegraphics[scale=0.25]{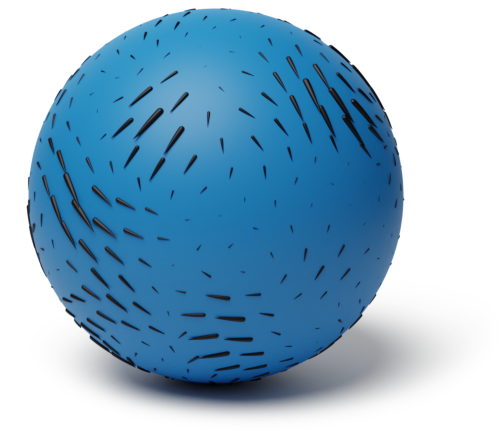}};
\node at (8,0) {\includegraphics[scale=0.25]{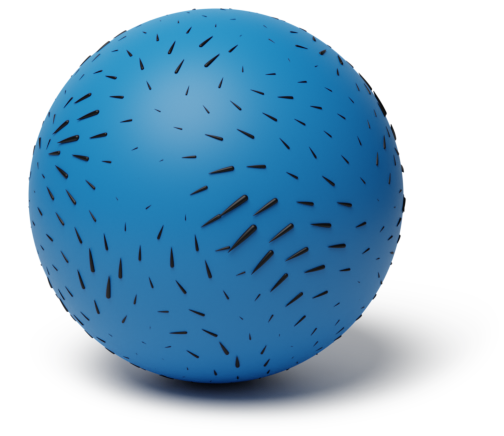}};
\end{tikzpicture}
\caption[Gaussian vector fields]{Three samples from a Gaussian vector field. Each of these samples, by virtue of being a continuous vector field on a two-dimensional sphere, automatically contains at least one location whose value is zero.}
\label{fig:gvf}
\end{figure}

This result is also called the \emph{hairy ball} theorem due to its interpretation: namely, it tell us in particular that if we consider a two-dimensional sphere with hair on its surface, then it is not possible to comb the hair without creating a swirl or other location where the direction of the hair changes abruptly.
An illustrated set of vector fields, described further in what follows, can be seen in \Cref{fig:gvf}.

This result restricts the kind of technical tools available for studying vector fields on manifolds.
In particular, it implies that for a generic smooth manifold, it is not possible to choose a smoothly-varying set of basis vectors in each tangent space simultaneously.
Because of this, it also follows that a smooth section $f : X \-> TX$ \emph{cannot} be reinterpreted a continuous function $f : X \-> \R^d$ in general.
This makes understanding vector fields a more delicate endeavour for manifolds compared to Euclidean spaces.

We are interested in defining vector-valued Gaussian processes on manifolds.
The first question that arises, then, is what should one actually mean when describing a random function $f : \Omega \x X \-> TX$ as \emph{Gaussian}?
The preceding sections suggest two possible approaches.

\1 Adapt the notion of \emph{Gaussian finite-dimensional marginal distributions} to the tangent bundle setting.
\2 Introduce an appropriate infinite-dimensional \emph{vector space of sections} and work with Gaussians in the sense of duality.
\0 

Clearly, the latter approach is more elegant and more general, but it is also also more effortful and potentially less constructive.
Since we are ultimately interested in working with Gaussian vector fields algorithmically, we adopt the former approach, but keep the latter view in mind to ensure soundness of our overall strategy.
Recall that the vector direct sum is denoted by $\oplus$.
We proceed as follows.

\begin{definition}[Gaussian vector field]
Let $X$ be a smooth manifold.
We say that a random section $f : \Omega \x X \-> TX$ is \emph{Gaussian} if for any finite set of locations $x_1, \ldots, x_n \in X$, the random vector $(f(\.,x_1),\ldots,f(\.,x_n)) \in T_{x_1}X \oplus \ldots \oplus T_{x_n}X$ is Gaussian in the sense of duality.
\end{definition}

Thus, we see that in spite of the fact that a tangent bundle is a manifold, and not a vector space, it possesses enough linear structure to admit a sensible notion of finite-dimensional marginals.
Here, we immediately see the value of the duality-based approach to Gaussianity originally developed in \Cref{ch:intro}: by using this, rather than working with bases, we avoid the need for cumbersome change-of-basis consistency checks.
We now examine this notion's view from the vantage point of an embedding into Euclidean space.

\begin{proposition}
Let $i : X \-> \R^p$ be a smooth embedding. Then 
\[
f_i : \Omega \x i(X) &\-> \R^p
&
f_i(\omega,x) &= i_* f(\omega, i^{-1}(x))&
\]
is a vector-valued Gaussian process in the standard sense.
\end{proposition}

\begin{proof}
Since the embedding $i$ is bijective on its image, any set of locations $x_1^{(i)},..,x_n^{(i)} \in i(X)$, will map uniquely onto a set of locations $x_1,..,x_n \in X$.
By definition of vector pushforward maps, $i_*$ maps tangent vectors from $T_{x_1} X \oplus .. \oplus T_{x_n} X$ into $\R^{n\times p}$ linearly.
Since Gaussianity is preserved by linearity, every set of finite-dimensional marginals in the embedded process is Gaussian, and the claim follows.
\end{proof}

This confirms that our definition of a vector-valued Gaussian process is the right one: if we embed the manifold in a Euclidean space, we obtain the correct kind of stochastic process.
\Cref{fig:gvf} illustrates a set of random samples from a Gaussian vector field, constructed using the technique presented in the sequel.

As before, the mean of such a process will simply be a given vector field.
The next step is to determine what is an appropriate notion of a \emph{kernel}.
In the Euclidean case, a \emphmarginnote{matrix-valued kernel} is a symmetric positive semi-definite function $k : X \x X \-> \R^{d \x d}$.
This notion more-or-less completely analogous to the scalar-valued case.
A priori, this notion seems completely coordinate-dependent due to the matrix appearing in the output.
To avoid this, we instead consider the map
\[
((x,\v{v}), (x',\v{v}')) \|> \v{v}^T \m{K}_{xx'} \v{v}'
\]
which describes the action of the kernel matrix on vectors.
To continue, we need an appropriate notion of bilinearity for the given setting.
We formulate this notion in general vector bundles, though we are interested in the tangent bundle $TX$ whose fibers are the vector spaces $T_x X$.

\begin{definition}[Fiberwise bilinear function]
Let $B$ be a vector bundle over $X$ with fibers $V_x$, $x\in X$.
We say that a function $k : B \x B \-> \R$ \emph{fiberwise bilinear} if for all pairs of points $x, x' \in X$ we have that
\[
k(\kappa \alpha_x + \lambda \beta_x, \gamma_{x'}) &= \kappa k(\alpha_x, \gamma_{x'}) + \lambda k(\beta_x, \gamma_{x'})
\\
k(\alpha_x, \mu \gamma_{x'} + \nu \delta_{x'}) &= \mu k(\alpha_x, \gamma_{x'}) + \nu k(\alpha_x, \delta_{x'})
\]
for any $\alpha_x, \beta_x \in V_x$, $\gamma_{x'}, \delta_{x'} \in V_{x'}$ and $\kappa, \lambda, \mu, \nu \in \R$.
\end{definition}

This leads to the following notion of a kernel.

\begin{definition}[Positive semi-definite kernel]
A symmetric fiberwise bilinear function $k : T^* X \x T^* X \-> \R$ is called a \emph{positive semi-definite kernel} if for any set of covectors $\alpha_{x_1}, \ldots, \alpha_{x_n} \in T^*X$, we have 
\[
\sum_{i=1}^n\sum_{j=1}^n k(\alpha_{x_i}, \alpha_{x_j}) \geq 0
.
\]
\end{definition}

At first, this may seem surprising: why should the kernel be a function defined on the \emph{cotangent} bundle, rather than the tangent bundle?
A clue is given by the form of the multivariate Gaussian density, which contains a $\v{x}^T\m{K}_{\v{x}\v{x}}^{-1}\v{x}$ term---note the presence of the inverse.
This suggests that since the inverse kernel matrix acts on vectors, the kernel matrix should act on covectors.
This mirrors the duality-based view described in \Cref{ch:intro}, and is clarified further by considering the kernel of a given Gaussian vector field.

\begin{definition}[Cross-covariance kernel]
The \emph{cross-covariance kernel} of a Gaussian vector field is defined as the map
\[
\alpha_x, \beta_{x'} \|> \Cov(\dualprod{\alpha_x}{f(\.,x)}, \dualprod{\beta_{x'}}{f(\.,x')}).
\]
\end{definition}

We now verify that this is indeed the correct notion of a kernel for the given setting.
To do this, we need a general form of the Kolmogorov Extension Theorem, given by the following result.

\begin{result}[Kolmogorov Extension Theorem]
Let $(X_\alpha,\c{B}_\alpha,\c{O}_\alpha)_{\alpha\in A}$ be a family of measurable spaces, each equipped with a topology.
For each finite $B \subseteq A$, let $\mu_B$ be an inner regular probability measure on $X_B = \bigtimes_{\alpha\in B} X_\alpha$ equipped with the product $\sigma$-algebra $\c{B}_B$ and product topology $\c{O}_B$, obeying
\[
\del{\proj_C}_* \mu_B = \mu_C
\]
whenever $C \subseteq B \subseteq A$ are two nested finite subsets of $A$, where projections $\proj_C: X_B \-> X_C$ are defined by $\proj_C(\cbr{x_\alpha}_{\alpha \in B}) = \cbr{x_\alpha}_{\alpha \in C}$, and $\del{\proj_C}_*$ denotes the measure-theoretic pushforward by $\proj_{C}$.
Then there exists a unique probability measure $\mu_A$ on $\c{B}_A$ with the property that $\del{\proj_B}_* \mu_A = \mu_B$ for all finite $B \subseteq A$.
\end{result}

\begin{proof}
\textcite[Theorem 2.4.3]{tao11}.
\end{proof}

We are now ready to prove our bijective correspondence.

\begin{theorem}
The distribution of every Gaussian vector field is uniquely determined by its mean vector field and cross-covariance kernel.
Moreover, each such pair defines a Gaussian vector field.
\end{theorem}

\begin{proof}
Let $x_1,..,x_n \in X$ and for $\alpha = (\alpha_{x_1},..,\alpha_{x_n})$ and $\beta = (\beta_{x_1},..,\beta_{x_n})$ define 
\[
\mu_{x_1,..,x_n} &= (\mu(x_1),..,\mu(x_n))
&
k_{x_1,..,x_n}(\alpha, \beta) &= \sum_{i=1}^n \sum_{j=1}^n k(\alpha_{x_i}, \beta_{x_j})
.
\]
Let $\pi_{x_1,..,x_n} \~[N](\mu_{x_1,..,x_n}, k_{x_1,..,x_n})$ be defined in the sense of duality.
These are our finite-dimensional marginals: by Gaussianity, it is clear these are uniquely determined by the mean and kernel.
Taking $X$ to be the index set, and $(T_x X)_{x\in X}$ with the standard topology and Borel $\sigma$-algebra to be our measurable spaces, we claim that the family of measures $(\pi_{x_1,..,x_n})_{\{x_1,..,x_n\} \subseteq X}$ satisfies the requirements of Kolmogorov's Extension Theorem.
First, for any $\{x_1,..,x_m\} \subseteq \{x_1,..,x_n\}$, by direct calculation we have 
\[
(\proj_{x_1,..,x_m})_* \pi_{x_1,..,x_n} = \pi_{x_1,..,x_m}
\]
using the canonical projection induced by the direct sum.
Next, note that each $\pi_{x_1,..,x_n}$ is a probability measure on a finite-dimensional real space, it is automatically inner regular.
Thus, it follows that there exists a unique measure on the infinite product space $\bigtimes_{x\in X} T_x X$.
This is a topological measure space: if we equip it with the obvious linear structure, and define the linear operator 
\[
\c{I} : \bigtimes_{x\in X} T_x X &\-> \Gamma_{\f{nns}}(TX)
&
(\c{I}s)(x) &= (x, \proj_x s)
\]
where $\Gamma_{\f{nns}}(TX)$ is the space of not necessarily smooth sections, equipped with the pushforward $\sigma$-algebra, we obtain by pushforward the probability distribution of our desired process.
The claim follows.
\end{proof}

\subsection{Coordinate representations}

Using the preceding recipe, we now understand what needs to be defined in order to construct vector-valued Gaussian processes on manifolds. 
To make this construction concrete, we proceed to develop the calculations needed to implement such processes numerically.
The first step is to understand how to represent vector fields in coordinates---we do so using bases in each tangent space.

\label{ntn:frame}
\begin{definition}[Frame]
Define a \emph{frame} $F$ to be a collection of not necessarily smooth sections $e_i : X \-> TX$ for $i=1,..,d$ such that, for every $x \in X$, the vectors $e_i(x)$ form a basis of $T_x X$.
Given a frame, define the \emph{coframe} $F^*$ to be the collection of sections $e^i : X \-> T^* X$ defined pointwise using the dual basis with respect to $F$.
\end{definition}

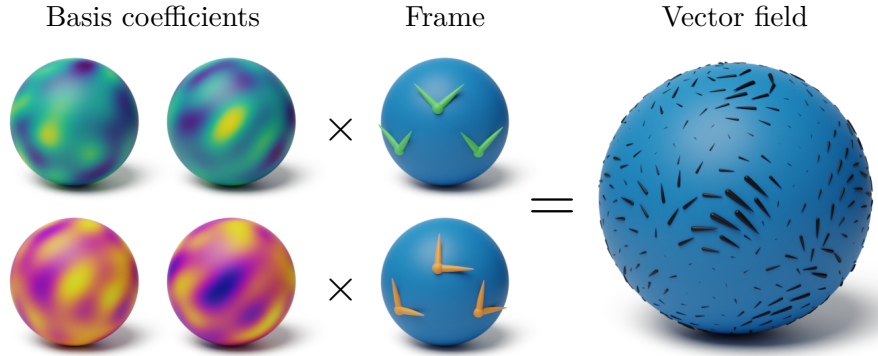
\begin{figure}
\tikzset{external/export next=false}
\begin{tikzpicture}
\node at (0,2.125) {\includegraphics[scale=0.25]{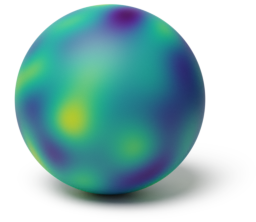}};
\node at (0,0) {\includegraphics[scale=0.25]{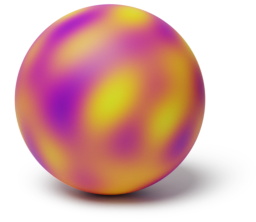}};
\node at (2.0625,2.125) {\includegraphics[scale=0.25]{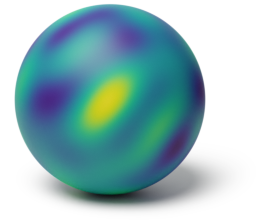}};
\node at (2.0625,0) {\includegraphics[scale=0.25]{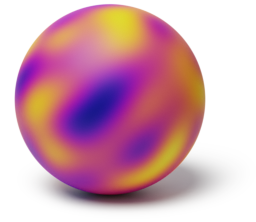}};
\node at (3.34375,2.125) {\LARGE\ensuremath{\x}};
\node at (3.34375,0) {\LARGE\ensuremath{\x}};
\node at (4.875,2.125) {\includegraphics[scale=0.25]{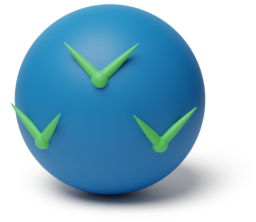}};
\node at (4.875,0) {\includegraphics[scale=0.25]{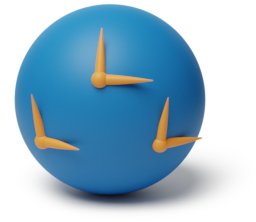}};
\node at (6.125,1.0625) {\Huge\ensuremath{=}};
\node at (8.875,1) {\includegraphics[scale=0.25]{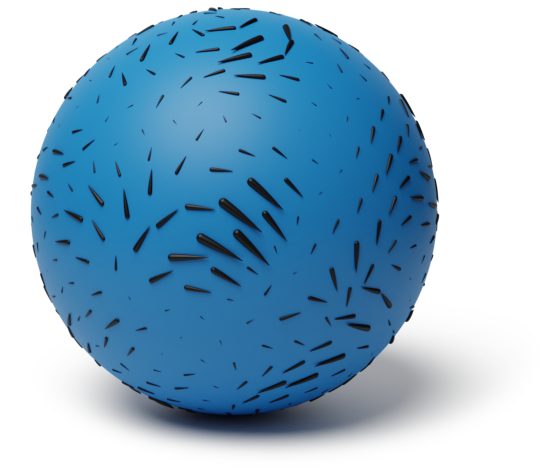}};
\node at (0.875,3.625) {Basis coefficients};
\node at (4.71875,3.625) {Frame};
\node at (8.53125,3.625) {Vector field};
\end{tikzpicture}
\caption[Representing vector fields in different frames]{A vector field can be represented by specifying its values in any frame. Here, we illustrate the same vector field on the sphere represented using two different frames.}
\label{fig:frame}
\end{figure}

The key words in the above definition are \emph{not necessarily smooth}: recall again that for many manifolds, choosing a frame which varies smoothly in space is impossible---a single component of such a frame would yield a non-vanishing vector field, and such a vector field might not exist.
An example of a vector field represented in two different frames can be seen in \Cref{fig:frame}.
We can use this to calculate the coordinate expression of a cross-covariance kernel with respect to a frame.
This is given by a function $\m{K}_F : X \x X \-> \R^{d \x d}$ as
\[
\m{K}_F(x,x') = \begin{bmatrix} \label{eq:K_F}
k(e^1(x), e^1(x')) & \dots  & k(e^1(x), e^d(x')) \\
\vdots           & \ddots & \vdots \\
k(e^d(x), e^1(x')) & \dots  & k(e^d(x), e^d(x'))
\end{bmatrix}
\]
which is a matrix-valued kernel in the usual sense.
This expression is frame-dependent: to see how it transforms upon a change of frame, introduce another frame $\tl{F}$, and consider a function
\[
f(x) &= \sum_{i=1}^d f^i(x) e_i(x) = \sum_{i=1}^d \tilde{f}^i(x) \tilde{e}_i(x)
\]
expressed with respect to $F$ and $\tl{F}$, where $\tilde{f}^i$ and $g^i$ are scalar-valued functions.
We say that $\tl{F}$ is obtained from $F$ by a \emphmarginnote{change of frame} if there is a matrix-valued function $\m{A} : X \-> \R^{d \x d}$ such that 
\[
\v{\tilde{f}}(x) = \m{A}(x) \v{f}(x)
\]
for all $x$.
Since $e_i$ and $\tilde{e}_i$ are bases, such a function always exists.
The range of $\m{A}$ is a subgroup of the general linear group $GL(d,\R)$, and in general depends on what frame-dependent properties one wants to preserve.
For instance, if one wants to preserve orthonormality of bases in each tangent space, one would consider a subgroup of orthonormality-preserving transformations, namely $SO(d)$.
By direct calculation, $\m{K}_F$ can be re-expressed as 
\[
\m{K}_{\tl{F}}(x,x') = \m{A}(x)\m{K}_F(x,x')\m{A}(x)^T
\]
with respect to $\tl{F}$.
This condition is called \emphmarginnote{equivariance under change of frame}.
At this point, it is clear why we did not even attempt to generalize matrix-valued kernels directly by positing candidate formulas: for an arbitrary manifold, there is very little hope of simply guessing an expression that is simultaneously positive semi-definite and equivariant under change of frame.

\subsection{Projected kernels}

Now that we understand how to write kernels of Gaussian vector fields in coordinates, we are ready to consider techniques for constructing such kernels and calculating them numerically. 
With our preparation complete, the construction we present is extremely simple, consisting of two steps.

\1 Embed a set of $p$ scalar-valued Riemannian Gaussian processes from $X$ into $i(X) \subseteq \R^p$, and use them to assemble a vector-valued Gaussian process defined on on $i(X)$.
\2 Project the tangent vectors onto each tangent space, obtaining a tangential vector field in the embedded space, and therefore a vector field on $X$.
\0 

\label{ntn:frame-projection}
We now proceed to make this precise.
Suppose that we have an embedding
\[
i : X &\-> \R^p
&
i_{x,*} : T_x X &\-> \R^p
\]
where the pushforward is defined at all points $x\in X$.
Introducing a frame $F$ allows us to define the projection map, which is a matrix-valued function $\m{P}_F : X \-> \R^{p \x d}$ defined by
\[
\m{P}_F(x) =\begin{bmatrix}
i_{x,*}(e_i(x)) \\
\vdots \\ 
i_{x,*}(e_d(x))
\end{bmatrix} 
\]
which describes how to project arbitrary vectors onto tangent planes.
Note that if $X$ is Riemannian with metric $g$, then under the tangent-cotangent isomorphism, the linear map given by $\m{P}_F$ is a right-inverse to the map given by $\m{P}_F^T$, which describes how a vector field with respect to a frame can be re-expressed with respect to the coordinates in the embedded space.
In particular, $\m{P}_F^T \m{P}_F = \m\Gamma$, where $\Gamma_{ij} = g(e_i(x), e_j(x))$ is the coordinate representation of the metric.
It is now clear how to make our idea precise.

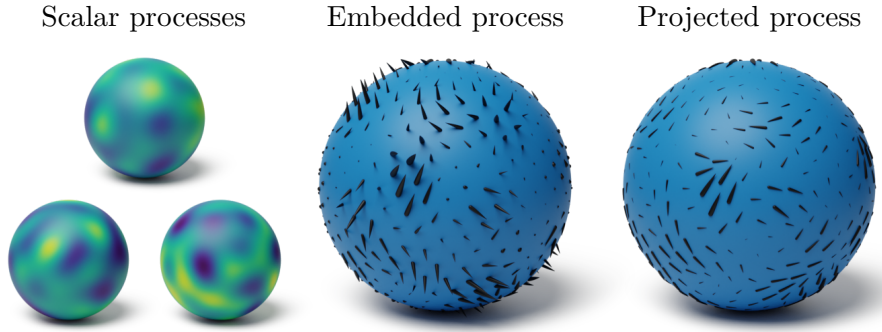
\begin{figure}
\tikzset{external/export next=false}
\begin{tikzpicture}
\node at (1,1) {\includegraphics[scale=0.25]{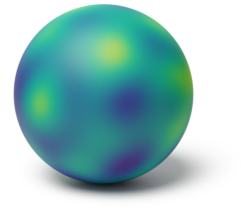}};
\node at (0,-0.875) {\includegraphics[scale=0.25]{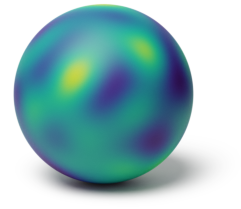}};
\node at (2,-0.875) {\includegraphics[scale=0.25]{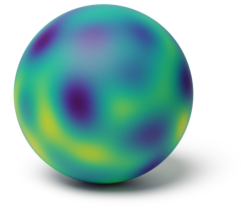}};
\node at (5.125,0) {\includegraphics[scale=0.25]{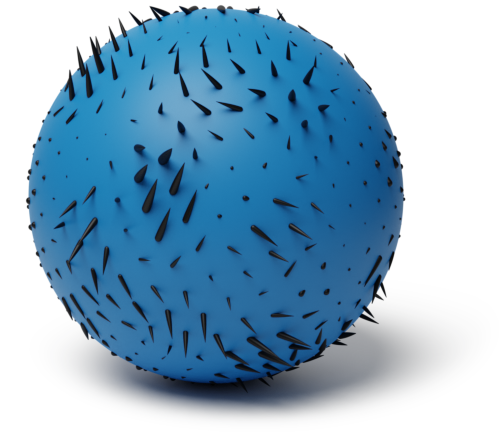}};
\node at (9.125,0) {\includegraphics[scale=0.25]{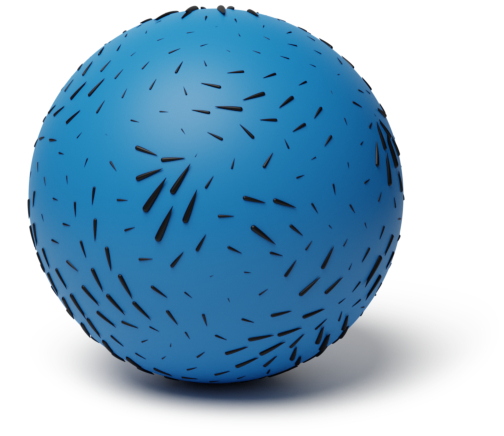}};
\node at (0.84375,2.40625) {Scalar processes};
\node at (4.8125,2.40625) {Embedded process};
\node at (8.8125,2.40625) {Projected process};
\end{tikzpicture}
\caption[Projected kernels]{Here, we show how to construct Gaussian process on the sphere whose cross-covariance kernel is a projected kernel. First, we construct three scalar Gaussian processes on the sphere. Then, embed the sphere in $\R^3$, and the scalar Gaussian processes to form a vector-valued Gaussian process on the embedded sphere. Finally, we project the resulting Gaussian process onto the sphere, yielding the desired tangential vector field.}
\label{fig:proj-ker}
\end{figure}

\begin{definition}[Projected kernel]
Let $\v{f} : \Omega \x X \-> \R^{p\x p}$ be a Gaussian process defined on $i(X) \subseteq \R^p$, and define the Gaussian vector field
\[
f(\omega, x) = \m{P}_F(x) \v{f}(\omega, i(x))
.
\]
We call the cross-covariance kernel of a process constructed this way a \emph{projected kernel}.
\end{definition}

It follows that if $\v\kappa : X \x X \-> \R^{p \x p}$ is the matrix-valued cross-covariance kernel of $\v{f}$, then the cross-covariance kernel of our Gaussian vector field is given by 
\[
\m{K}_F(x,x') = \m{P}_F(x) \v\kappa(x,x') \m{P}_F(x)^T
.    
\]
Note that here, we generally require a Riemannian structure on $X$ in order to define a useful set of scalar-valued processes in order to obtain $\v{f}$ or, equivalently, $\v\kappa$, to begin with.
The simplest approach is to make each component in the embedded space independent.
Particularly in cases where the embedding is isometric, this gives a wide class of easy-to-understand kernels.
An example can be seen in \Cref{fig:proj-ker}.

It's easy to see that all cross-covariance kernels arise this way: using Nash's Theorem, we can embed an arbitrary Gaussian vector field on a Riemannian manifold into a Euclidean space, glue together the resulting scalar components, and project back to obtain the Gaussian vector field we started with.

This completes our technical development. 
Starting from first principles, we defined the notion of a Gaussian vector field, described in what sense such processes possess mean vectors and cross-covariance kernels, and defined a wide-ranging class of kernels which is completely constructive and can be implemented in software.
In particular, our constructions are automatically compatible with standard training methods such as variational inference.

\subsection{Illustrated examples}

We now explore training Gaussian vector fields on Riemannian manifolds on observed data. 
Our goal is to demonstrate that the developed techniques are, in total, sufficiently explicit so as to enable Gaussian vector fields to be trained using standard methods without the need for any additional machinery.

We work with the sphere.
For training data, we consider global wind velocity, chosen because it is freely and readily available.
Specifically, we consider interpolation of wind velocity deviations from historical average, at a fixed height, using wind velocity data directly observed by a satellite.
For the prior, we use a Gaussian vector field model constructed using a projected kernel whose components are independent Matérn-3/2 scalar-valued Gaussian processes on the sphere.

We train the Gaussian process using exactly the same procedure as that used in the preceding sections: by optimizing kernel hyperparameters using gradient descent, and computing the value of the vector field at all locations using pathwise sampling.
All computations are performed with respect to a fixed frame defined by the latitude-longitude coordinate system.

\begin{figure}
\begin{subfigure}{0.49\textwidth}
\includegraphics[scale=0.25]{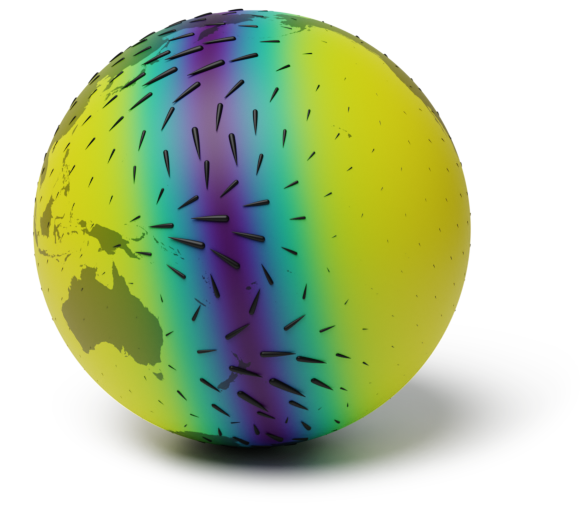}
\caption{Posterior mean}
\end{subfigure}
\begin{subfigure}{0.49\textwidth}
\includegraphics[scale=0.25]{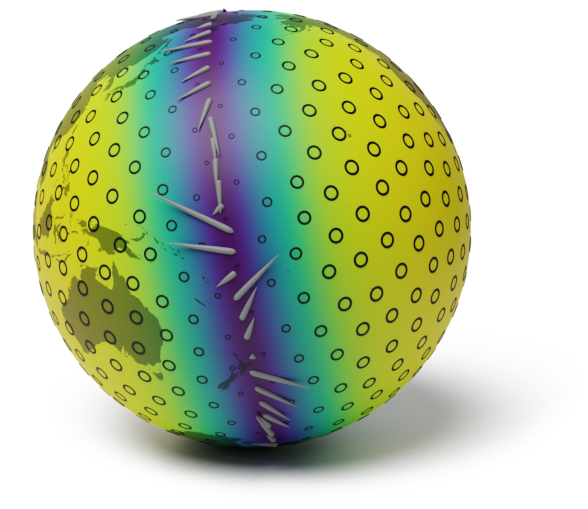}
\caption{Data and posterior covariance}
\end{subfigure}
\caption[Posterior Gaussian vector field: climate data]{Posterior mean and posterior cross-covariance, with the trace of the latter denoted by color, for a projected Gaussian vector field constructed using three independent Matérn-3/2 Gaussian processes on the sphere. Training data is shown in gray. The Gaussian process is seen to oversmooth data in locations with large jumps, such as near the easterly trade wind deviations in the center. We also observe that uncertainty increases as we move away from the training locations.}
\label{fig:posterior-vector-field}
\end{figure}

Results are shown in \Cref{fig:posterior-vector-field}.
Immediately, we see that this dataset exhibits behavior not suitable for the chosen model.
Specifically: in some regions, the data exhibits large jumps, while in other regions, it is very smooth over large distances.
This means the effective length scale governing variability is spatially non-uniform, which requires a more sophisticated model to effectively represent, such as one with a spatially varying length scale.

The chosen Gaussian process model responds to this by overinterpolating the data and reverting to the overall mean in regions with rapidly changing wind velocity.
While this is not ideal from a modeling perspective, it is at least a relatively graceful failure mode, compared to numerical instability or incorrect results in other regions where rapid shifts do not occur.
Effects such as these are closely linked to non-linear behavior in the differential equations governing weather phenomena, and generally require special consideration.

On the other hand, we observe that the trained Gaussian process model is smooth, even though the choice of frame is non-smooth at the top of the sphere---a key motivation behind developing the theory for working with vector fields in a geometrically-sound way in the first place.
Similarly, we see that posterior uncertainty increases as we move away from locations where training data is available.
This shows that the model correctly incorporates geometry and behaves in a manner mirroring its scalar-valued analogs.

One notable characteristic of the obtained posterior vector field is that the covariance of the output vectors at any given point is close to isotropic.
This can be seen from the ellipsoids in \Cref{fig:posterior-vector-field}, which appear circular, though they are clearly elliptical from numerical examination.
This behavior appears to be a feature of the data locations chosen in this particular example, which lie approximately on a great circle, and does not necessarily occur in general.

The model used is also not realistic in other ways, beyond its use of a single global length scale: in particular, it does not incorporate time or any kind of physical information into its design.
Nonetheless, we believe that it serves as a useful demonstration that constructing vector-valued Gaussian processes using the ideas developed is practical.
We hope this motivates further development to apply such models and extensions thereof more broadly, both for the sphere and for other manifolds.

\section{Geometry-aware Bayesian optimization}

We now study the role that geometry plays when working with decision systems built using Gaussian processes.
To do so, we perform Bayesian optimization of standard benchmark functions on Riemannian manifolds using the Riemannian Matérn and squared exponential Gaussian processes with the numerical techniques developed herein.
Following \textcite{jaquier20}, we call this setting \emph{geometry-aware Bayesian optimization}.

\begin{figure}
\begin{subfigure}{0.32\textwidth}
\includegraphics[scale=0.25]{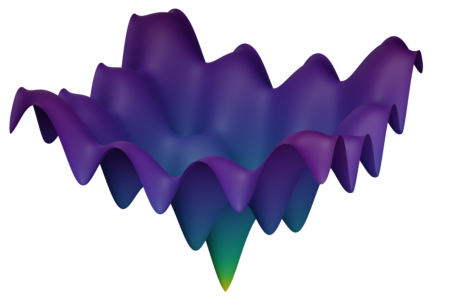}
\caption{Ackley function}
\end{subfigure}
\begin{subfigure}{0.32\textwidth}
\includegraphics[scale=0.25]{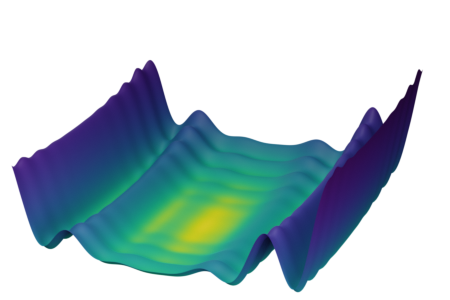}
\caption{Levy function}
\end{subfigure}
\begin{subfigure}{0.32\textwidth}
\includegraphics[scale=0.25]{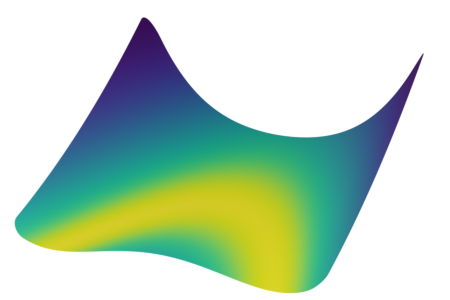}
\caption{Rosenbrock function}
\end{subfigure}
\caption[Global optimization benchmark functions]{Global optimization benchmark functions used in the geometry-aware Bayesian optimization experiments. All three are defined and shown on an ambient Euclidean space of dimension two.}
\label{fig:benchmark-functions}
\end{figure}

We consider three manifolds: the spheres $\bb{S}^3$ and $\bb{S}^5$, and the special orthogonal group $SO(3)$, each embedded within a Euclidean space in the natural manner.
We use three target functions commonly used as benchmarks in global optimization: the \emph{Ackley} \cite{ackley87}, \emph{Levy} \cite{levy82}, and \emph{Rosenbrock} \cite{rosenbrock60} functions.\footnote{The Ackley, Levy, and Rosenbrock benchmark functions, respectively, are given as
{\setlength{\tabcolsep}{0.25ex}\begin{tabular}{r c l}
$f_{\f{A}}(x_1,..,x_d)$ & $=$ & $20-20\exp(-0.2 (\frac{1}d \sum_{i=1}^d x_i^2)^{1/2}) - \exp(\frac{1}d \sum_{i=1}^d \cos(2\pi x_i)) + e$
\\
$f_{\f{L}}(x_1,..,x_d)$ & $=$ & \makebox[10.375cm][s]{$\sin(\pi w_i)^2 + \sum_{i=1}^{d-1} (w_i - 1)^2 (1 + 10\sin(\pi w_i + 1)^2) + (w_d - 1)^2 (1 + \sin(2\pi w_d)^2)$}
\\
$f_{\f{R}}(x_1,..,x_d)$ & $=$ & $\sum_{i=1}^{d-1} 100(x_{i+1} - x_i^2)^2 + (x_i - 1)^2$
\end{tabular}\newline}
where $w_i = 1 + \frac{x_i - 1}{4}$ and $d$ is the dimension of the space on which the functions are defined.}
These are defined in the ambient spaces: we optimize their restrictions onto the manifold of interest, centered to ensure the minima is on the manifold.
\Cref{fig:benchmark-functions} plots these benchmark functions for a Euclidean space.

\begin{figure*}[p!]
\includegraphics{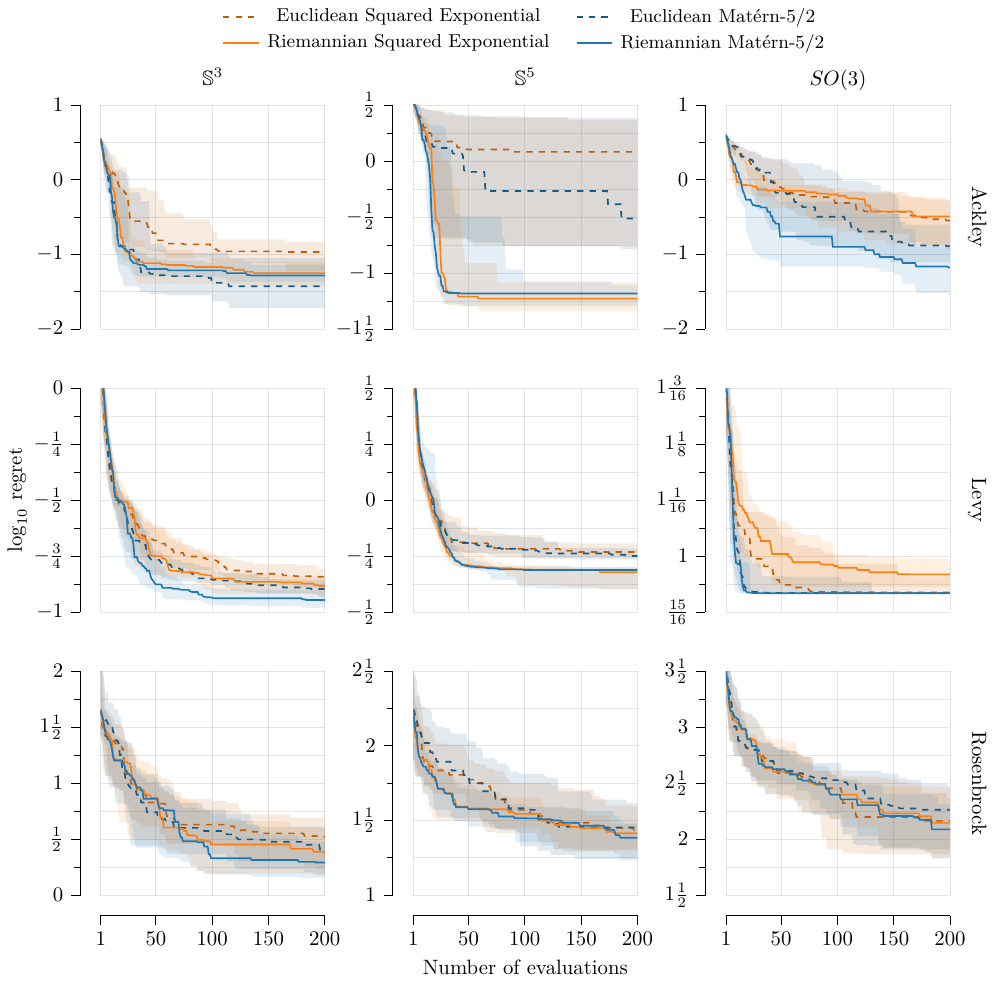}
\caption[Geometry-aware Bayesian optimization benchmark results]{Geometry-aware Bayesian optimization benchmark results.}
\label{fig:gabo}
\end{figure*}

For each manifold, we use two priors: Matérn with smoothness $\nu = 5/2$ and squared exponential, both with a Gaussian likelihood.
The remaining kernel hyperparameters and likelihood error variance are learned from observed data via gradient descent.
We use expected improvement for our acquisition function, computed directly from the kernel.
We run for a total of $n=200$ iterations, and repeat each experiment $30$ times to assess variability.

To establish a performance baseline, we compare against Bayesian optimization in ambient Euclidean space.
Here, the acquisition point is selected by performing constrained optimization on the acquisition function, ensuring that points selected for evaluation are located on that manifold.

Results can be seen in \Cref{fig:gabo}.
We plot median regret curves, along with first and third quartiles.
From this, we see that the performance of geometry-aware Bayesian optimization depends on interplay between the manifold and the overall shape of the function being optimized. 

For the $3$-dimensional sphere, all techniques produce comparable regret curves.
This shows working with geometry-aware models and acquisition optimization, compared to constrained acquisition optimization on Euclidean models defined in ambient spaces, offers less benefit in cases where the ambient space is not too different from the manifold.
The performance of Matérn and squared exponential models, in this case, is similar.

For the $5$-dimensional sphere, the role of geometry becomes more pronounced.
For the Ackley and Levy functions, we see a clear improvement from using geometry-aware methods.
For the Rosenbrock function, which possesses less local variability, performance of all methods remains similar.
For all three target functions, Matérn and squared exponential models perform similarly.

On the $3$-dimensional special orthogonal group, geometry-aware models perform similarly to models defined on the ambient Euclidean spaces for all three benchmark functions.
On the other hand, here, Matérn models outperform squared exponential models. 
Therefore, we see that the effect of kernel smoothness on performance of Bayesian optimization may depend on the geometry of the manifold, as well as the target function.

Overall, the results show that Riemannian models perform comparably to Euclidean ones in some cases, and outperform them in others.
The effect varies according to the interplay between the target function being optimized and the manifold's geometry.
We also see that slightly less smooth Matérn Gaussian processes perform comparably to squared exponential Gaussian processes in some cases, and outperform them in others.

Bayesian optimization is often used in settings where function evaluations are extremely expensive, and therefore even the relatively modest performance gains seen here can be consequential.
In such cases, taking advantage of constraints, symmetries, and other geometric properties is a promising route towards improved performance.
Geometry-aware Bayesian optimization offers a principled framework for doing so.

\section{Conclusion}

In the preceding section, we have developed three key classes of \emph{non-Euclidean Gaussian process} models: the \emph{Riemannian Matérn} and \emph{graph Matérn} classes of scalar-valued Gaussian processes, and the \emph{projected kernel} class of vector-valued Gaussian processes on Riemannian manifolds.
Each of these provides a unifying view of previously-proposed methods, and expands Gaussian processes models' scope of applicability.

The key idea for defining Matérn Gaussian processes on Riemannian manifolds was to not attempt to define kernels using geodesic distances---such constructions, while intuitively appealing, turn out to not lead to an effective mathematical formalism.
Instead, we relied on the idea of \emph{functional calculus} built using spectral theory of the Laplace--Beltrami operator to introduce operators through which we constructed the desired Gaussian processes as transformations of Gaussian white noise processes.

Adopting this viewpoint enabled us to reinterpret models previously-proposed within a cumbersome finite element framework using a more abstract approach built via the theory of stochastic partial differential equations and reproducing kernel Hilbert spaces.
From this approach, we obtained formulas for pointwise evaluation of the \emph{Riemannian Matérn kernel} in terms of eigenvalues and eigenfunctions of the Laplace--Beltrami operator.
The formalism also enabled us to define and compute the \emph{Riemannian squared exponential kernel}.

While our original original formalism was introduced for working with manifolds, it is clear that the constructions are more general.
Inspired by the discrete computations used in implementing the Riemannian Gaussian processes we defined, we explored purely-discrete analogs of these models, obtaining the \emph{graph Matérn kernel} and \emph{graph squared exponential kernel}.
These enabled us to apply Matérn Gaussian processes to model phenomena in purely discrete spaces whose geometry departs significantly from manifold-like settings.

Finally, we returned to the manifold setting, this time studying formalism for defining \emph{Gaussian vector fields}.
This setting is non-trivial from the point of conceptualization: a vector-valued Gaussian is only \emph{locally} a vector-valued random function, and instead should be formulated using the differential-geometric formalism of \emph{random sections}.
We therefore began by clarifying what the appropriate notion of a \emph{Gaussian vector field}  and \emph{cross-covariance kernel} should be, and how to work with this notion numerically.

Having done so, we introduced a wide class of cross-covariance kernels for defining such processes called \emph{projected kernels}.
These kernels can be built from the Riemannian Gaussian process models already defined, together with an embedding of the manifold into an ambient Euclidean space, and produce covariances that reflect the manifold's geometry.
This construction enabled us to implement Bayesian learning of vector-valued data on the sphere.

Having explored these constructions from a purely model-building point of view, we concluded by considering them within a decision-making setting, by studying \emph{geometry-aware Bayesian optimization}.
There, saw that using geometry-aware models also resulted in a modest performance improvement in cases where topology and geometry play a non-trivial role, particularly for higher-dimensional spaces.

It is clear that our ideas apply within a wider scope than what we have explored.
For example, our derivations for the Matérn class apply essentially-unmodified to closely-related settings such as \emph{manifolds with boundary}, provided one introduces boundary conditions which ensure the operators of interest behave the same way as in our setting.
Extensions such as this make excellent candidates for future work.

Our differential-geometric framework also provides a promising way to study \emph{symmetry} within Gaussian processes and Bayesian optimization.
Since symmetry plays a fundamental role in geometry and in physics, understanding how to handle symmetries is a promising avenue for building richer classes of Gaussian process models, and applying techniques such as Bayesian optimization to wider classes of scientific problems.

We hope that techniques such as the ones presented here provide a foundation upon which these and other developments can be built.
We therefore conclude our presentation of contributions, and move to discuss the state and overall picture of Gaussian processes and decision systems that we have studied.

\chapter{Discussion}
\label{ch:discussion}

\lettrine{B}{ayesian learning} with Gaussian process models offers a clear and comprehensive framework for making decisions that assess and propagate uncertainty in order to resolve explore-exploit tradeoffs inherent in decision-making settings.
In this dissertation, we explored two avenues for broadening applicability of Gaussian process models in such settings.

In \Cref{ch:pathwise}, we presented pathwise conditioning methods.
These methods allow a posterior Gaussian process, as an actual random function, to be expressed as a sum of a prior Gaussian process and a dependent update term.
Pathwise conditioning gives rise to Gaussian process approximations for which all stochasticity can be evaluated in advance, which makes it substantially easier for upstream algorithms to interface with the Gaussian process.

In \Cref{ch:noneuclidean}, we presented a collection of techniques for working with Gaussian process models defined on non-Euclidean spaces, including Riemannian manifolds and graphs. 
We derived constructive and practical expressions which enable such models to be trained using standard methods.
For the Riemannian setting, we did so for both scalar-valued and vector-valued processes.

For both pathwise conditioning and non-Euclidean Gaussian process models, the end result of our constructions was the ability to perform Bayesian optimization in a flexible and effective manner tailored to the setting at hand.
This allowed us to benchmark our ideas in one of the simplest, but most important, classes of decision systems.

\section{Future work}

One end goal of Gaussian process research is the design of decision-making systems which simply work in the settings where they apply, without the usual tuning or numerical difficulties, and fail gracefully in settings where they do not apply.
The ideas presented naturally lead to a number of further research directions towards this goal, which we briefly comment on now.

\subsection*{Decision systems beyond optimization}

Perhaps the most promising research direction made possible by pathwise conditioning methods is the design of data-efficient decision systems for tasks beyond optimization.
\textcite{neiswanger21} propose \emph{Bayesian algorithm execution}, a framework which generalizes Bayesian optimization by allowing for the task to be defined by using a general \emph{underlying algorithm} in place of optimization.

For this, \textcite{neiswanger21} introduce \emph{InfoBAX}, an information-theoretic acquisition function constructed directly from the algorithm being considered.
This enables the introduction of \emph{Bayesian Dijkstra's algorithm} for finding shortest paths in a graph in a data-efficient manner, and other novel decision systems.

Pathwise conditioning methods play a key role in computing the InfoBAX acquisition function.
Specifically, one runs the underlying algorithm on posterior random functions, producing a set of \emph{execution paths} of points queried by the algorithm.
InfoBAX is then constructed from the execution trace and output via a set of information-theoretic calculations.
Approximate pathwise conditioning makes it possible to compute the required execution traces in an efficient, accurate, and convenient manner.

Gaussian processes have not yet been widely considered for tasks not conveniently expressed as maximization of rewards.
Bayesian algorithm execution enables such methods to be considered for settings beyond Bayesian optimization and model-based reinforcement learning, which are well-established and whose limitations are largely known.
Developing this framework to the same degree of detail afforded to those cases is therefore a particularly promising research direction.

From a theoretical point of view, one of the first steps to be taken in understanding Bayesian algorithm execution is to understand what is the appropriate notion of \emph{regret} for this setting, among a number of possible variations, and to develop techniques for studying it.
The analysis of closely-related state-of-the-art bandit algorithms, such as \emph{information-directed sampling} \cite{russo14}, offers hints on how to proceed.

In parallel, it seems fruitful to look for applications where more general data-efficient decision systems beyond optimization are potentially useful, and yet where Gaussian process models are viable. 
\emph{Multi-objective Bayesian optimization} \cite{emmerich05,campigotto14,hernandezlobato16,suzuki20}, where the goal is finding a Pareto frontier, rather than obtaining a global minima, gives a concrete algorithm class to consider studying from this point of view.

From the Bayesian algorithm execution framework alone, we see that pathwise conditioning methods enable potential new avenues for constructing decision systems and therefore growing the applicability and scope of Gaussian process methods within machine learning.
We hope that, in the coming years, more of these avenues are explored, enabling Gaussian processes to have larger role in the machine learning toolbox.

\subsection*{Connections to numerical analysis}

Another promising research direction is exploring the connections between Gaussian processes and numerical analysis.
Interest in these connections has grown rapidly in the Gaussian process community within machine learning in recent years, as a way to improve performance and usability of software implementations \cite{matthews17,gardner18,vanderwilk20}. 
Ideas based on pathwise conditioning and stochastic partial differential equations offers new avenues for taking advantage of these connections.

Firstly, one can study use of iterative solvers, such as the conjugate gradient method, in combination with pathwise conditioning and inducing points.
Iterative solvers are promising because they can in principle tackle larger linear systems for which Cholesky factorization simply fails, provided they are set up correctly.
A large set of techniques in this class have recently been proposed for Gaussian process models \cite{dong17,gardner18,pleiss18,meanti20,pleiss20}.

The computational costs of an iterative solver depend on how many iterations are required for convergence, which vary according to the condition number of the linear system being solved \cite{saad03}.
Variational approximations such as the inducing point construction presented in \Cref{ch:pathwise} make it possible to work with linear systems that are substantially better-conditioned than those in the true posterior, due to additional diagonal terms added to the matrix.
Exploring such combinations is a promising route for improving performance.

Pathwise conditioning makes it possible to avoid computation of matrix square roots when working with iterative solvers.
Matrix square roots generally require further considerations than simply solving linear systems \cite{pleiss20} and can be less convenient.
With pathwise conditioning, the only additional quantities needed, beyond solution of the linear system itself, are generally log-determinant terms.
Techniques for computing these terms have been proposed \cite{saibaba17,dong17}, and are key part in making iterative methods effective.

In the partial differential equation and mathematical physics communities, iterative solvers for linear systems constructed using finite element methods achieve state-of-the-art performance for many problems \cite{evans10,lord14}.
Connections with stochastic partial differential equations, such as the ones explored here, offer a fruitful avenue for seeking inspiration for improving Gaussian process performance further.

A key lesson from numerical analysis is that, in order to solve a linear system, the most important step is often the construction of a \emph{preconditioner} to partially solve the problem before an iterative method is used \cite{saad03,e11,xu17}.
There, the degree of sophistication used in constructing preconditioners is often at least as large as that used in constructing iterative solvers themselves.

Multi-grid methods provide a broad framework for constructing preconditioners which achieve state-of-the-art performance for many partial differential equations \cite{e11,xu17}. 
In essence, such methods provide a framework for resolving long-range effects at a coarse resolution, and short-range effects at a fine resolution, improving computational efficiency in the process.
Adapting such ideas to the Gaussian process setting offers possibilities for improvement.

Numerical analysis can also be a source of inspiration for improving variational inference.
Inducing points often struggle on datasets where the overall data size is large, but the amount of data points within any given length scale ball is small.
Multi-scale methods suggest that right kind of approximation for this regime should involve replacing long-range effects with averaged, sparsified versions thereof \cite{e11,xu17}.
Understanding how construct such a variational approximation is therefore a promising avenue for future work.

In total, ideas from numerical analysis provide a fruitful source of ideas for improving Gaussian processes further.
We hope that in the coming years, these ideas come to fruition, and using Gaussian processes in most practical settings become just as easy and seamless as training neural networks on well-understood problem classes has become.

\subsection*{Improved understanding of kernels on geometric spaces}

The constructions presented in \Cref{ch:noneuclidean} for building non-Euclidean Gaussian process models give general formulas applicable for a wide variety of spaces.
Many manifolds encountered in practice, particularly those used in physics and engineering, possess additional mathematical structure, which can be used to both construct new kernels and provide avenues for more efficiently computing the Matérn and squared exponential kernels studied previously.

For example, the eigenfunctions of the Laplace--Beltrami operator on the sphere are the spherical harmonics \cite{chavel84,canzani13}.
The spherical harmonics, in turn, can be re-expressed in terms of the \emph{Gegenbauer polynomials}, which are a set of polynomials on $[-1,1]$ which are orthogonal with respect to a certain weight function \cite{gradshteyn14}.
The Riemannian Matérn and squared exponential kernels can therefore be written in terms of these polynomials, and turns out to only depend on the geodesic distance between two points.

This is a substantial simplification compared to the general case, and one can use it to pre-compute the kernel for various values of the geodesic distance, and interpolate the resulting values, rather than evaluate an infinite series every time the kernel needs to be computed.
Note that the spherical squared exponential kernel, when expressed using this simplification, is not the squared exponential of the geodesic distance: instead, it is a different, more complicated function of the geodesic distance.

Similar simplifications also occur on the special orthogonal group, suggesting that there may be a general theory that describes them.
Finding such a theory is a promising direction for future work.
To do so, the first step would be to understand how symmetries of a manifold affect kernels defined over it via spectral methods similar to the ones we have used.

Simplifications may also occur due to other structures beyond symmetries.
Lie groups are particularly important in many areas, including robotics, computer vision, physics, and many engineering disciplines.
Due to their prominence, it therefore seems fruitful to study potential methods for computing kernels on Lie groups, provided they are equipped with a Riemannian metric compatible with the group structure.

In particular, studying kernels on non-compact Lie groups, and related spaces such as certain matrix manifolds, is a promising direction towards a general theory of Riemannian kernels which does not require compactness.
Here, one would need more general forms of spectral theory than Sturm--Liouville decompositions.
These are mathematically sophisticated, requiring additional machinery such as integration over projection-valued measures \cite{lang12}.
Exploring such generalizations is a promising avenue for further work.

Finally, one can consider extending our constructions to more general spaces than manifolds, such as for example manifolds with boundary, which possess similar spectral properties.
Spaces like this are often important in applications areas \cite{solin18,krainski18,solin19,coveney20}.
Studying heat kernels, which have been considered in settings as general as metric measure spaces \cite{grigoryan03,grigoryan14}, gives hints about the total scope which may be possible.

Summarizing, the constructions we have studied only begin to describe the full scope of geometry-aware kernels such as the Matérn and squared exponential class constructed using Sturm--Liouville theory and described here.
We hope this first step leads to further substantive understanding of this class of methods within machine learning.

\section{Conclusion}

The contributions presented in this thesis expand the set of settings where Gaussian processes can be used.
In particular, pathwise conditioning methods reduce the barriers needed to deploy Gaussian process models once they have been trained, and make it easier to obtain posterior quantities of interest for the setting at hand.
Simultaneously, the obtained kernel expressions for non-Euclidean Gaussian processes enable such models to be trained using off-the-shelf methods in a standard manner.

Overall, these ideas reduce the barriers that need to be overcome in order for practitioners to use Gaussian processes in day-to-day work.
We hope the ideas presented enable more people to consider using Gaussian process models as part of their workflow, and bring decision-making systems such as Bayesian optimization to new areas of science, technology, and engineering.

\printbibliography[heading=bibintoc,title={References}]

\end{document}